# Improving Deep Learning through Automatic Programming

**Master's Thesis in Computer Science**

Dang Ha The Hien

May 14, 2014
Halden, Norway

**Østfold University College**    **www.hiof.no**

# Abstract

Deep learning and deep architectures are emerging as the best machine learning methods so far in many practical applications such as reducing the dimensionality of data, image classification, speech recognition or object segmentation.... In fact, many leading technology companies such as Google, Microsoft or IBM are researching and using deep architectures in their systems to replace other traditional models. Therefore, improving the performance of these models could make a very strong impact in the area of machine learning. However, deep learning is a very fast-growing research domain with many core methodologies and paradigms just discovered over the last few years. This thesis will first serve as a short summary of deep learning, which tries to include all of the most important ideas in this research area. Based on this knowledge, we suggested, and conducted some experiments to investigate the possibility of improving the deep learning based on automatic programming (ADATE). Although our experiments did produce good results, there are still many more possibilities that we could not try due to limited time as well as some limitations of the current ADATE version. I hope that this thesis can promote future work on this topic, especially when the next version of ADATE comes out. This thesis also includes a short analysis of the power of ADATE system, which could be very useful for other researchers who want to know what it is capable of.

**Keywords:** Deep Learning, Automatic Programming, ADATE, Neural Networks, Machine Learning

# Acknowledgments

First and foremost, I would like to thank my thesis supervisor Prof. Roland Olsson for the advice, support, and kindness that he has given to me over the past year. Without him, I definitely could not complete this thesis. Many parts of this thesis were done by both of us. Therefore, I used "we" as the subject in this thesis to indicate that fact. I would also like to thank my fellow students, especially Kristin Larsen, who has allowed me to use her experiment result in my thesis. In addition, I want to thank my family and friends, who were always beside and support me in difficult times. Last but not least, I want to thank my best friend, Nguyet Thanh, who took me out of my loneliness and encouraged me to keep following my dream.

# Prerequisites

Although machine learning as well as neural networks and deep learning is a very specialized field, which covers many different aspects of mathematics, physics, and biology. . . , we believe that a graduated student in computer science could easily understand the main ideas and results in this report. However, to get a deeper understanding, the reader should have basic knowledge about probability, statistics, linear algebra, and calculus. He or she should also be familiar with basic mathematics optimization concepts like Taylor series and steepest gradient descent.

# Contents

# List of Figures

# List of Tables

# Listings

# Chapter 1

# Introduction

Machine learning is a huge research area based on many ideas in mathematics, physics, and biology.... Therefore, in this first part, we only try to give a very brief introduction to machine learning as well as explanations for some related important specialized terms, which hopefully could help one whose major is not machine learning understand the report. Besides, we also introduce artificial neural networks and deep learning - the main field that we are working on in our thesis - their advantages and practical applications as our motivations. Finally, we present our research question, research plan, and outline of the rest of the thesis.

## 1.1 Machine Learning

In the book "Machine Learning" [27], Michell 1997 defined "The field of machine learning is concerned with the question of how to construct computer programs that automatically improve with experience... A computer program is said to learn from experience E with respect to some class of tasks T and performance measure P, if its performance at tasks in T, as measured by P, improves with experience E" and "Machine learning is inherently a multidisciplinary field. It draws on results from artificial intelligence, probability and statistics, computational complexity theory, control theory, information theory, philosophy, psychology, neurobiology, and other fields".

Another more modern and practical definition is used in the book "Data Mining - Practical Machine Learning Tools and Techniques" [51], where the authors state that "Machine learning provides the technical basis of data mining" and "Data mining ... is to build computer programs that sift through databases automatically, seeking regularities or patterns, ..., generalize to make accurate predictions on future data".

In general, a machine learning system is based on three main parts: training data, model, and training (i.e learning) algorithm. Each model has a set of parameters that we can use the training algorithm to train (i.e fit) these parameters to the training data, and apply that learned model on another separated test data to measure its real performance. There are two main types of learning: *supervised* and *unsupervised* learning. In supervised learning, the training data contains both the input and the desired output (e.g. images and their classifications), and the training algorithms try to construct a mapping (through adapting model's parameters) that defines the output patterns in terms of the input patterns. In unsupervised learning, the desired output is unknown, and the training algorithms try to discover the structure in data, or even generate new similar data.

Besides, there are many different types of model, from the simplest ones such as ZeroR (just guess the most likely answer - usually used as the baseline performance) or Linear Regression to the much more complex models such as Support Vector Machine (SVM) or Deep Neural Networks (DNNs). Each model has a different assumption (i.e. inductive bias) about the possible input-output mappings (in supervised learning) or possible data distribution (in unsupervised learning). This is because the problem we are dealing with in machine learning is known to be "ill posed" (refers to chapter 5), which cannot be solved unless some prior information is asssumed. For example, in supervised learning, the mapping function may not even exist, or we do not have enough information from the training examples to reconstruct that mapping, or the unavoidable presence of noise in the training examples can make a perfect fitted model useless in practice [14]. An easy example for inductive bias in linear regression is that we assume that the input and the desired output are linearly dependent, so that we can represent the mapping from input to output through a linear equation. The training algorithm for it therefore just searches for the "best" equation.

In this thesis, we focus on supervised learning algorithm in Artificial Neural Networks (ANNs) and Deep Neural Networks (DNNs), which are very powerful models. However, we also introduce some unsupervised learning algorithms and models, which are commonly used in learning process of DNNs.

## 1.2   Artificial Neural Networks

The artificial neural networks (ANNs) have been inspired in part by the observation that biological learning systems are built of very complex webs of interconnected neurons. Artificial neural networks are built out of a densely interconnected set of simple units, where each unit takes a number of real-valued inputs (possibly the outputs of other units) and produces a single real-valued output (which may become the input to many other units) [27]. While ANNs are loosely motivated by biological neural systems, there are many complexities to biological neural systems that are not modeled by ANNs, and many features of the ANNs are known to be inconsistent with biological systems [27]. For example, we consider here ANNs whose individual units output a single constant value, whereas biological neurons output a complex time series of spikes.

There are many different types of neural networks so far. In this report, the term "Neural Networks" is used to implicitly refer to the "Artificial Feed Forward Neural Networks" if there is no other explicit explanation. A feed forward neural network is an artificial neural network where connections between the units do not form a directed cycle, and that therefore can be visualized as a multiple layers network. In Figure 1.1, we showed a typical feed forward neural network with 2 hidden layers (i.e the layers between input and output layers). The mapping between input and output is calculated by feeding the input into the input nodes, going through the hidden nodes and to the output nodes. The value (i.e activation) of each intermediate node is produced by first weighted summing of all its predecessors then going through an activation function (as described in Figure 1.2, please refer to Chapter 6 for more details). Many different activation functions can be used in ANNs, the simplest one is the $sgn()$ function: return +1 when input is greater than 0 and return -1 otherwise; which is used in *perceptron* networks. The currently popular activation functions are the $tanh()$, the sigmoid function ($f(x) = \frac{1}{1+e^{-x}}$) and the softmax function. These functions are popular partly because of their differentiability, so we can easily compute their gradients, which is crucial in gradient-based training algorithm.

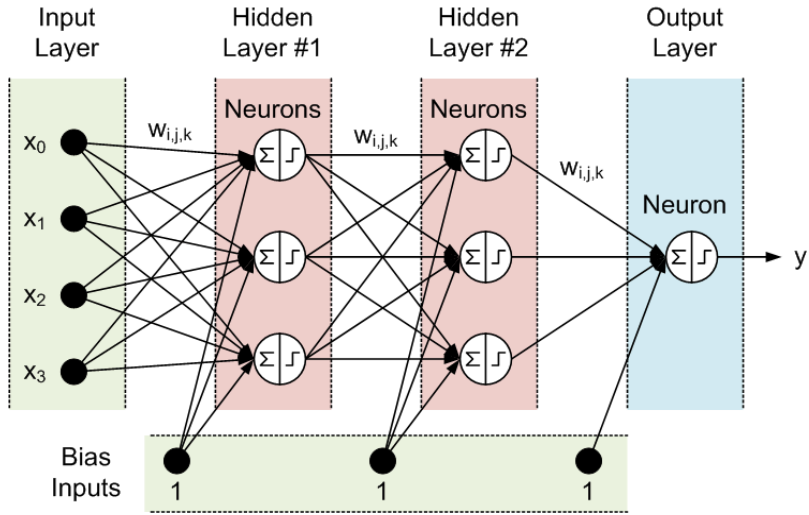

Figure 1.1: A typical feed forward neural network

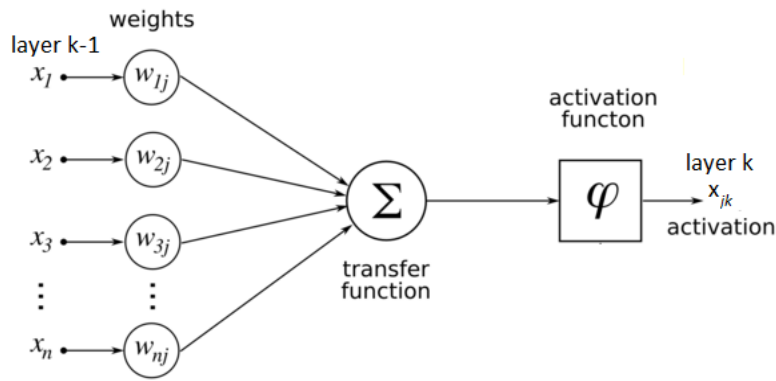

Figure 1.2: Activation of a node $j^{th}$ at layer $k^{th}$ is calculated by $x_{jk} = \varphi(\sum_{i=1}^{n} x_{i(k-1)} w_{ijk})$

The training process in ANNs is to adapt all of the connections' weight to the training data, by minimizing a cost function (i.e. objective function) – which is usually a mean square error (MSE), cross entropy, or negative log likelihood function. There are many different training algorithms for ANNs, including evolution strategy or genetic algorithms. However, gradient-based methods are the most popular. These methods are used to minimize the cost function $f(x)$ by iteratively following a search direction defined by the gradient (or a function of gradient) of the $f$ function at the current point (details of these methods are provided at Chapter 4. In this thesis, we only focus on studying these gradient-based methods as the main optimization approach for deep learning.

## 1.3 Deep Architectures and Deep Learning

Deep architectures are models that are composed of multiple levels of non-linear operations, such as in Deep Neural Networks (i.e. neural networks with more than 3 hidden layers), Deep Belief Nets (DBN), or Deep AutoEncoders. Theoretical results suggest that in order to learn the kind of complicated functions that can represent high-level abstractions (e.g., in vision, language, and other AI-level tasks), we may need deep architectures [1]. However, until 2006, training these deep architectures was believed to be difficult due to the well-known vanishing gradient problem. Hinton et al. 2006 [16] have introduced a new learning procedure to successfully tackle this problem. After that, different deep architecture models as well as new learning algorithms have been proposed and applied successfully in many areas, beating the state-of-the-art in certain applications such as in dimensionality reduction, modeling textures, modeling motion, object segmentation, information retrieval, robotics, natural language processing. . . [1]. In chapter 3, we give an overview of these deep architecture models: why we need deep architectures and the challenges of training these models. We also briefly introduce their popular training algorithms.

In fact, deep architectures are emerging as the best machine learning model so far in many practical applications such as reducing the dimensionality of data [16], image classification [22] or speech recognition [17]. . . we believe that improving just one of these models' architectures or training algorithm, in terms of their performance or computational cost, could make a very strong impact in the machine learning area.

## 1.4   Research Question and Methodology

### 1.4.1   Research question

In this thesis, we intend to study deep architectures (especially deep neural networks) and their learning algorithms, thereby understand the challenges regarding training these models. Based on the knowledge, we could investigate the possibility of improving these models or algorithms using ADATE (Automatic Design of Algorithms Through Evolution) [31]. However, deep learning is known to run extremely slowly (sometimes takes weeks to finish), especially on big dataset. ADATE, on the contrary, needs the program to run fast (usually less than ten seconds), because it has to evaluate at least millions of different program instances to find the best version. Therefore, to use ADATE to improve deep learning, we have to train deep architectures on a very small dataset. The biggest challenge is that we have to make sure that the improved version of that algorithm can still be useful in other dataset. Moreover, deep architectures and their training algorithms are implemented as a very complex program, which definitely could not be completely generated by ADATE. Therefore, we have to choose a small part of that program, which can significantly improve the whole performance if it is improved.

Basically, at the end of this thesis, we need to answer the following research questions:

**RQ 1** *How can the ADATE system improve the performance of deep learning?*
Secondary relevant research questions are:

    **RQ 1.1** *Which part of deep learning is possible to be improved by ADATE?*

    **RQ 1.2** *What dataset is suitable for using in ADATE to improve deep learning?*

    **RQ 1.3** *Could the improved version of deep learning be used in other dataset?*

    **RQ 1.4** *How can we implement correctly deep architectures and their training algorithms in Standard ML (SML)?*

### 1.4.2 Methodology

In general, machine learning is an experimental science, in which many models or algorithms are based on heuristic equations, which are learned from experiments. This thesis is not an exception. To answer these above research questions, we have to experiment with different possible solutions, as much as possible, to find out the best one. First, we have to get a deep understanding of deep architectures and their training algorithms. Then we choose a part that is most likely to be improved using ADATE. Finally, we choose the most suitable dataset and use ADATE to improve the algorithm on that dataset automatically.

## 1.5 Report Outline

Chapter 2, 3, 4, 5 and 6 provide the necessary background about deep learning:

- In chapter 2, we presented answer the question why we need deep learning, in both theoretical and empirical ways.

- In chapter 3, we gave an overview of deep architectures and challenges of training these models. We also introduced briefly about their popular training algorithms.

- In chapter 4, we summarized different popular gradient-based training algorithms for Neural Networks.

- In chapter 5, we presented the overfitting problem, and introduced different regularization techniques as the possible solutions.

- In chapter 6, we summarized recent well-known researches about activation functions used in deep learning.

Chapter 7 provides a short introduction about ADATE system, as well as an analysis of its power. Besides, based on knowledges provided in previous chapters, this chapter also suggests different deep learning algorithms that can probably be improved by ADATE.

Chapter 8 describes our experiments with ADATE in detail, from choosing the target algorithm, buiding tiny dataset, to testing results and analyzing overfitting problem. This chapter aims at answering all the research questions posed above.

Chapter 9 presents our conclusion about this project, as well as suggestions for future works.

# Chapter 2

# The Need for Deep Learning and Deep Architectures

Deep learning is a set of algorithms in machine learning that attempt to learn layered models of inputs (deep architectures). The layers in such models correspond to distinct levels of concepts, where higher-level concepts are defined from lower-level ones, and the same lower-level concepts can help to define many higher-level concepts.

Before the invention of pre-training, which makes deep learning more feasible, most of available learning algorithms correspond to shallow architectures. However, as we all know, the mammal brain is organized in a deep architecture [42]. The brain also appears to process information through multiple stages of transformation and representation. This is particularly clear in the primate visual system, with its sequence of processing stages: detection of edges, primitive shapes, and moving up to gradually more complex visual shapes[42].

Theoretically, some functions cannot be efficiently represented (in terms of the number of tunable elements) by architectures that are too shallow. In fact, functions that can be compactly represented by a depth $k$ architecture might require an exponential number of computational elements to be represented by a depth $k-1$ architecture. Therefore, poor generalization may be expected when using an inefficiently deep architecture for representing some functions [1].

## 2.1   Deep Architectures on Circuit Problem

In [3], Bengio et al. 2007 have summarized the advantages of deep architectures on circuit problems. According to them, complexity theory of circuits strongly suggests that deep architectures can be much more efficient (sometimes exponentially) than shallow architectures, in terms of the number of computational elements required to represent some functions. For example, the parity function with $d$ inputs (XOR problem) requires $O(2^d)$ examples and parameters to be represented by a Gaussian SVM (Bengio et al., 2006), $O(d^2)$ parameters for a one-hidden-layer neural network, $O(d)$ parameters and units for a multi-layer network with $O(\log_2 d)$ layers, and $O(1)$ parameters with a recurrent neural network. More generally, boolean functions (such as the function that computes the multiplication of two numbers from their $d$-bit representation) expressible by $O(\log d)$ layers of combinatorial logic with $O(d)$ elements in each layer may require $O(2^d)$ elements when expressed with only 2 layers (Utgoff & Stracuzzi, 2002; Bengio & LeCun, 2007).

## 2.2   The Challenges in training Deep Neural Networks

Both theoretical and experimental evidence suggests that training deep architectures is much more difficult than training shallow ones. In [1, 3, 7], the authors showed that gradient-based training process on deep supervised multi-layer neural networks (starting from random initialization) usually gets stuck in "apparent local optima or plateaus", which ends up with a poorer performance result compared to the shallow ones'. This phenomenon can be explained by the vanishing gradient, local optima, and pathological curvature problems, which known to be much more serious in deep architectures. These problems prevented us from using deep learning for a very long time until the appearance of pre-training methods. Hinton et al. has completely changed the story when introducing Deep Belief Networks and greedy layer-wise unsupervised pre-training methods in 2006 [18]. After that, many successful deep learning methods have been introduced (Bengio et al., 2007; Vincent et al., 2008; Weston et al., 2008; Lee et al., 2008), but all of them use a common idea with Hinton's one: the DNNs are first pre-trained by an unsupervised pre-training algorithm, and then fine-tuned by other classical supervised learning methods.

In the next chapter, besides deep architectures, many of the most popular models used in pre-training process of deep learning will also be introduced. Because if we could improve one of these pre-training building-block model, we could improve the performance of deep learning in general.

Besides pre-training, some recent researches have shown that learning deep networks can still be done fairly well using other classical learning methods such as Hessian-free Optimization ($2^{nd}$-order method) or even a standard stochastic gradient descent ($1^{st}$-order method)[26, 46]. They argued that while bad local optima do exist in deep-networks, in practice they do not seem to pose a significant threat. Instead, the difficulty is better explained by regions of pathological curvature (e.g. long narrow valley) in the objective function [26]. However, in their experiments, they could only successfully train deep autoencoders using their methods, and did not mention about other types of deep architectures. After conducting some experiments, we concluded that pre-training is still a much better approach than any other methods, in all of deep learning applications. However, there are still some exceptions, where we can train a deep architecture without the need of pre-training, such as using data augmentation [5] or convolutional neural networks (section 5.2.6).

## 2.3   Dominance of Deep Learning

In his presentation about deep learning [29], Andrew Ng. – a leading researcher in the deep learning area, has stated that "This is the first time a single type of model can compete almost all of the previous state-of-the-art results in machine learning".

One of the most important abilities which brings the power to deep learning is that it can automatically extract useful features in almost any type of input domain (e.g. Images/Video, Audio, Text,...). Figure 2.1 shows the importance of feature learning, where using raw input directly make the tasks on it (e.g. labels, classification, image search, ...) become almost impossible. What we need is a better way to present inputs: feature representations. For examples, to detect if the images in figure 2.1 is about motortype, we need to know: is there any wheel? Is there a handlebar?...Image presented in term of these features is much easier to process. However, we do not have wheel detector or handlebar detector. For decades, thousands of experts were trying to hand-design features to capture various statistical properties of the image/audio/text.... Some of these features are demonstrated in Figure 2.2.

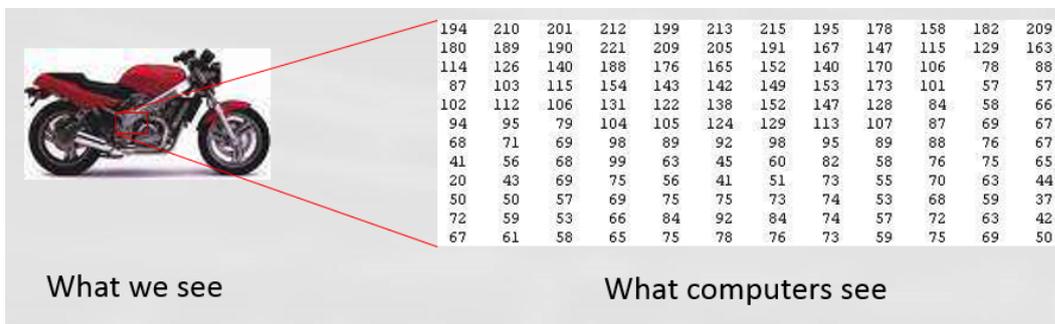

Figure 2.1: Difficulty on object classification task

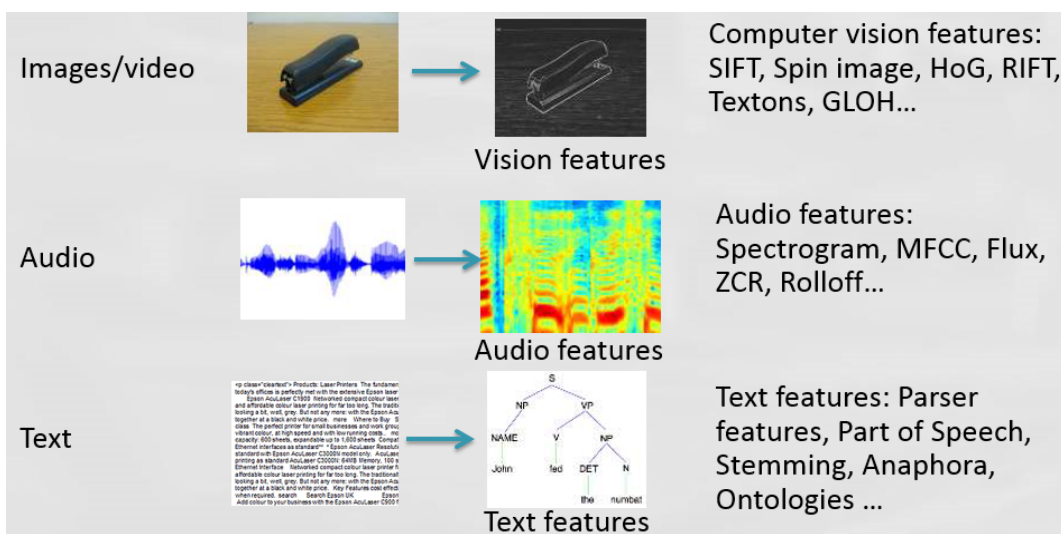

Figure 2.2: Hand-designed features examples

Deep learning, on the other hand, can automatically learn even better features than hand-designed features on many different input domains. This makes deep learning becomes powerful and general-purpose model. Figure 2.3 shows some examples of these different tasks that deep learning outperformed any previous methods. Seeing its potential in real-life applications, many big tech companies have recently hired deep learning leading researchers. In Mar, 2013, Google hired Geoffrey Hinton – the one who enabled deep learning in 2006 by his discovery of unsupervised pre-training – and his team to make AI a Reality (as in the announcement). Facebook also decided to hire prominent NYU professor Yann LeCun – the one who discovered convolutional neural networks – as the new director of their AI lab. Acquisition of Deep Mind – a London-based artificial intelligence – by Google in Jan 2014, which cost more than \$500 million, could show the heat of deep learning.

| Problems | Best Previous accuracy | Deep learning accuracy |
|---|:---:|:---:|
| Hollywood - Activity recognition | 48% | 53% |
| TIMIT - Phoneme Classification | 79.2% | 80.3% |
| CIFAR - Object classification | 80.5% | 82% |
| NORB − Object classification | 94.4% | 95% |
| AVLetters Lip reading | 58.9% | 65.8% |
| Paraphrase detection | 76.1% | 76.4% |

Figure 2.3: Deep learning applications

# Chapter 3

# Deep Architectures and Related Models

In this chapter, we give a brief introduction about different well-known deep architectures. Although we mainly focus on Deep Neural Networks and its related models, it is worth knowing other models such as Boltzmann Machines, Deep Belief Networks, or AutoEncoders, which are used in pre-training process of deep learning. This chapter is split into three general types of models: Deep Neural Networks, Boltzmann Machines-related models, and AutoEncoders models.

## 3.1 Deep Neural Networks

Being a fundamental deep architecture, Deep Neural Network is simply a neural network with many hidden layers. This is a very old idea but could not be used popularly because of difficulties in its training algorithm. After appearance of unsupervised pre-training, deep neural networks have flourished and become one of the most popular machine learning models nowadays. As shown in Figure 3.1, deep neural networks are composed of multi-layer of non-linear operations. Moreover, if learned in a good way, deep neural networks can represent input through multi-level feature representations, where features in higher layer model higher level abstractions in input.

**Layer-wise Unsupervised Pre-trainning**

Greedy layer-wise unsupervised pre-training is the most important strategy in deep learning. It allows us to train very deep neural networks very effectively. This strategy contains two main stages:

- *Pre-training:* We first pre-train one layer of the neural networks at a time (i.e. layer-wise) in a greedy way. To do this, we need a building block model. We pre-train each layer of the deep neural network by training a building block model for each layer and stack them one above another. Different building block models for this purpose can be used. The first appeared one is RBM proposed by Hinton in 2006. Since then, many different building block models have been introduced, such as sparse auto-encoders, denoising auto-encoders, or contractive auto-encoders. Figure 3.2 illustrates this pre-training process using an auto-encoder.

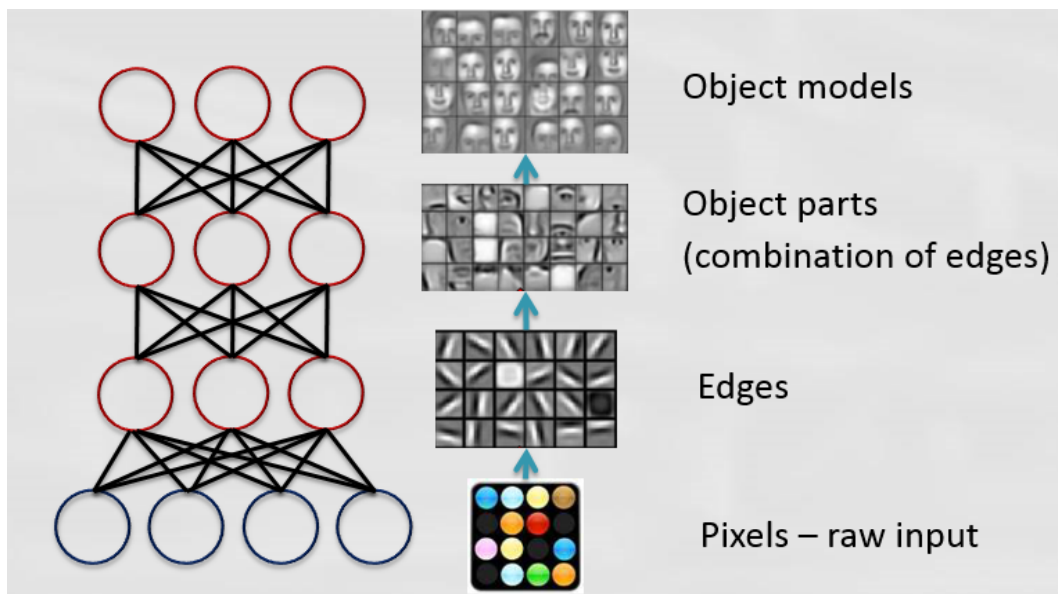

Figure 3.1: Deep neural networks represent input in multi-level feature representations

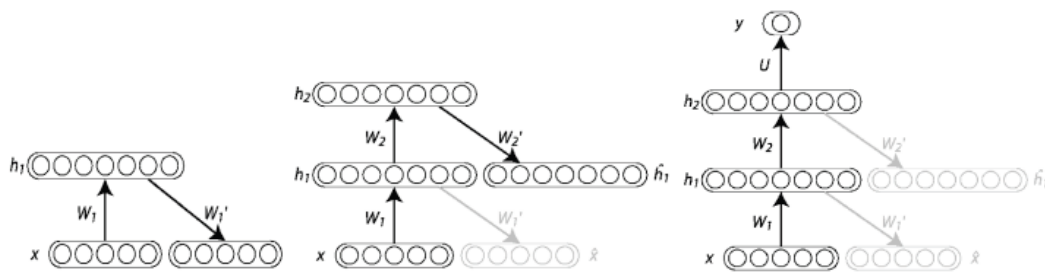

Figure 3.2: Pretraining process using auto-encoder

- *Fine-tuning:* After pre-training, we can use any standard backpropagation methods to fine-tune that pretrained deep neural network. The training task is now much easier than training from a random initialized network.

**Traditional Approaches for Deep learning**

Besides the pretraining strategy, researchers also tried to find a traditional way to do deep learning. Martens 2010 [26] has developed a $2^{nd}$-order optimization method based on the "Hessian-free" approach, and apply it to training deep autoencoder successfully. Ilya et al. 2013 [46] proved that a deep autoencoder could also be trained with 1st-order method like stochastic gradient descent, using well-chosen random initialization schemes called and various forms of momentum-based acceleration [46]. They used a variation of momentum called Nesterov's Accelerated Gradient to improve the convergence rate guarantee. Both Martens and Ilya et al. used "sparse initialization" in their methods. However, these methods are not good at training a general deep neural networks, where the pre-training still a much better strategy.

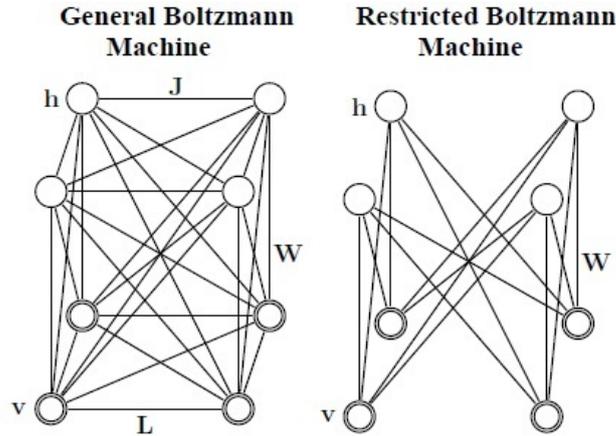

Figure 3.3: *(Taken from [40])* **Left**: A general Boltzmann machine. The top layer represents a vector of stochastic binary "hidden" features and the bottom layer represents a vector of stochastic binary "visible" variables. **Right**: A restricted Boltzmann machine with no hidden to hidden and no visible to visible connections.

## 3.2 Boltzmann Machines - Related Models

The most important model of this type is Restricted Boltzmann Machines (RBM), which is used as a building-block model in the unsupervised pre-training process introduced by Hinton in 2006. After that, other researchers have introduced some new types of autoencoders, which can replace the RBM's role in pre-training process. Actually, the RBM model is very similar to an autoencoder. This similarity will be presented in the next section about autoencoders.

### 3.2.1 Boltzmann Machines

A Boltzmann machine is a type of energy-based models, which maintains an energy function to define a probability distribution of each configuration of the variables of interest. It is a kind of stochastic recurrent neural network invented by Geoffrey Hinton and Terry Sejnowski. As displayed in Figure. 3.3, a Boltzmann machine is a network of symmetrically coupled stochastic binary units. It contains a set of visible units $\boldsymbol{v} \in 0, 1^D$, and a set of hidden units $\boldsymbol{h} \in 0, 1^P$. The energy of the state $\boldsymbol{v}, \boldsymbol{h}$ is defined as:

$$E(\boldsymbol{v}, \boldsymbol{h}; \theta) = -\frac{1}{2}\boldsymbol{v}^T \boldsymbol{L} \boldsymbol{v} - \frac{1}{2}\boldsymbol{h}^T \boldsymbol{J} \boldsymbol{h} - \boldsymbol{v}^T \boldsymbol{W} \boldsymbol{h} \qquad (3.1)$$

where $\theta = \{\boldsymbol{W}, \boldsymbol{L}, \boldsymbol{J}\}$ are the model parameters, which represent visible-to-hidden, visible-to-visible, and hidden-to-hidden symmetric interaction terms. (We have omitted the bias terms for clarity of presentation)

The probability that the model assigns to a visible vector $\boldsymbol{v}$ is:

$$p(\boldsymbol{v}; \theta) = \frac{1}{Z(\theta)} \sum_{\boldsymbol{h}} exp(-E(\boldsymbol{v}, \boldsymbol{h}; \theta))$$
$$Z(\theta) = \sum_{\boldsymbol{v}} \sum_{\boldsymbol{h}} exp(-E(\boldsymbol{v}, \boldsymbol{h}; \theta)) \qquad (3.2)$$

The parameter updates, originally derived by Hinton and Sejnowski (1983), that are needed to perform gradient ascent in the log-likelihood can be obtained from Eq. 3.2:

$$\Delta \boldsymbol{W} = \alpha(E_{P_{data}}[\boldsymbol{v}\boldsymbol{h^T}] - E_{P_{model}}[\boldsymbol{v}\boldsymbol{h}^T])$$
$$\Delta \boldsymbol{L} = \alpha(E_{P_{data}}[\boldsymbol{v}\boldsymbol{v^T}] - E_{P_{model}}[\boldsymbol{v}\boldsymbol{v}^T])$$
$$\Delta \boldsymbol{J} = \alpha(E_{P_{data}}[\boldsymbol{h}\boldsymbol{h^T}] - E_{P_{model}}[\boldsymbol{h}\boldsymbol{h}^T])$$

(3.3)

Exact maximum likelihood learning in this model is intractable because exact computation of both the data-dependent expectations and the model's expectations takes a time that is exponential in the number of hidden units. However, setting both $\boldsymbol{J} = 0$ and $\boldsymbol{L} = 0$ would introduce the well-known restricted Boltzmann machine (RBM) (Smolensky, 1986) (see Fig. 3.3, right panel), which can be learned efficiently using Contrastive Divergence (CD) (Hinton, 2002).

### 3.2.2 Restricted Boltzmann Machines - RBM

A RBM is a Boltzmann machine, which is restricted by omitting intra-layer connections (e.g. hidden-hidden and visible-visible connections). This restriction allows us to obtain exact samples from the conditional distribution $p(\boldsymbol{h}|\boldsymbol{v};\theta)$ and $p(\boldsymbol{v}|\boldsymbol{h};\theta)$, thank to independence between hidden-hidden and visible-visible nodes. Therefore, we can sample configuration of h based on state of v and vice versa.

The parameter updates for RBM can be derived as following (including bias terms):

$$\Delta w_{ij} = \alpha(\langle v_i h_j \rangle_{data} - \langle v_i h_j \rangle_{model})$$

(3.4)

$$\Delta b_i = \alpha(\langle v_i \rangle_{data} - \langle v_i \rangle_{model})$$

(3.5)

$$\Delta b_j = \alpha(\langle h_i \rangle_{data} - \langle h_i \rangle_{model})$$

(3.6)

We can compute $\langle v_i h_j \rangle_{data}$ by clamping the visible units at the data vector $\boldsymbol{v}$ and then compute the expected value of $\boldsymbol{h}$ easily. To compute $\langle v_i h_j \rangle_{model}$, we can first clamp the visible units at data vector $\boldsymbol{v}$, then sampling the hidden units, then sampling the visible units, and repeating this procedure infinitely many times. After infinitely many iterations, the model will forget its starting point and we can sample from its equilibrium distribution. However, it has been shown that this expectation can be approximated well in finite time by running the sampling chain for only a few steps. This method is called Contrastive Divergence (CD).

Using these derivatives, we can maximize the log-likelihood function, which leads to decrease the energy of training cases (increase probability) and increase the energy of other unseen cases.

The main reason RBM are interesting is its capability of pre-training a deep network.

### 3.2.3 Deep Belief Nets - DBN

Deep Belief Nets are probabilistic generative models that are composed of multiple layers of stochastic, latent (hidden) variables. The top two layers have undirected, symmetric

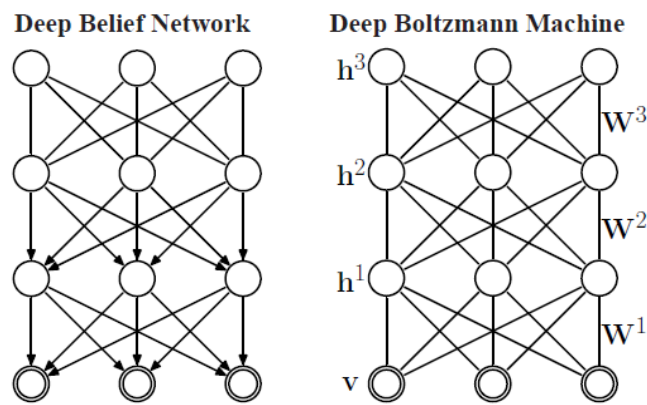

Figure 3.4: *(Taken from [41])* **Left**: Deep Belief Network (DBN), with the top two layers forming an undirected graph and the remaining layers form a belief net with directed, top-down connections **Right**: Deep BoltzmannMachine (DBM), with both visible-to-hidden and hidden-to-hidden connections but with no within-layer connections. All the connections in a DBM are undirected.

connections between them and form an associative memory. The lower layers receive top-down, directed connections from the layer above (see Figure. 3.4)

Basically, DBNs extend the RBM architecture to multiple hidden layers, where the weights in layer $h_l$ are trained by keeping all the weights in the lower layers constant and taking as data the activities of the hidden units at layer $l-1$. Therefore, DBN is a stacked RBMs, which are trained greedily and in sequence. It can be proved that each time we add another layer of features (i.e. RBM) we improve a variational lower bound on the log probability of the training data.

### Fine-tuning for generation

After learning many layers of features, we can fine-tune the features to improve generation using the "wake-sleep" algorithm. First, we do a stochastic bottom-up pass to adjust the top-down weights to be good at reconstructing the feature activities in the layer below, then doing a few iterations of sampling in the top level RBM to adjust the weights in the top-level RBM. Finally, we do a stochastic top-down pass to adjust the bottom-up weights to be good at reconstructing the feature activities in the layer above.

### Fine-tuning for discrimination

After pre-training the DBNs, we can unfold that DBN into a deep neural network, and use standard back propagation to fine-tune the model for better discrimination.

### Applying DBN on digit recognition

Hinton 2006. has designed a DBN shown in figure 3.5 to learn the joint distribution of digit images and digit labels. The model learns to generate combinations of labels and images. To perform recognition, they start with a neutral state of the label units and do an up-pass from the image followed by a few iterations of the top-level associative memory.

For the MNIST data set (including 70,000 digit images), they achieved 1.25% error rate on test set (using generative fine-tuning) and 1.15% (using back propagation fine-tuning)

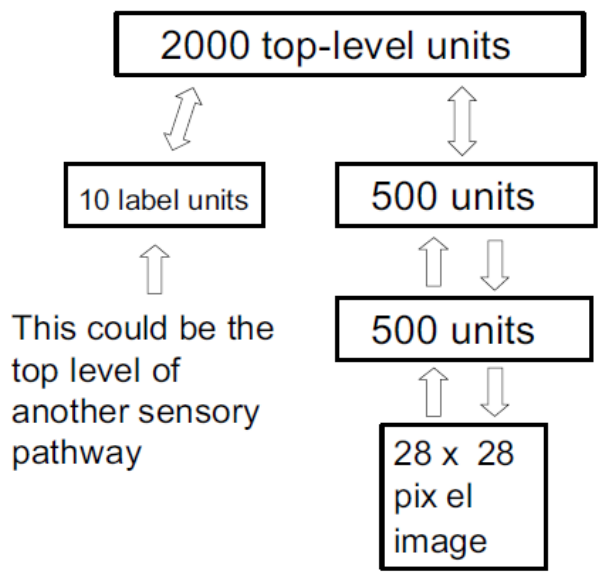

Figure 3.5: *(Taken from [41])* DBN model that was used for MNIST data set

## 3.3   AutoEncoders

A classical autoencoder (i.e. autoassociator) is a special type of feed forward neural network which is trained to encode the input in some representation so that the input can be reconstructed from that representation. In other words, we set the target values to be equal to the inputs. This is an unsupervised learning task because we do not use labels information during the training process. A typical autoencoder is shown in Figure 3.6, which is a feed forward network with two weight-layers (usually constrained to be equal).

### 3.3.1   Basic Autoencoder (AE)

An autoencoder contains two parts:

- *Encoder*: is a function $f$ that maps an input $x \in R^{d_x}$ to hidden representation $h(x) \in R^{d_h}$. It has the form:

$$h = f(x) = s_f(Wx + b_h) \tag{3.7}$$

where $s_f$ is a activation function, typically a logistic sigmoid function. The encoder is parametrized by a $d_h \times d_x$ weight matrix $W$, and a bias vector $b_h \in R^{d_h}$

- *Decoder*: The decoder function $g$ maps hidden representation $h$ back to a reconstruction $y$:

$$y = g(h) = s_g(W'h + b_y) \tag{3.8}$$

where $s_g$ is the decoder's activation function, typically either the identity (yielding linear reconstruction) or a sigmoid. The decoder's parameters are a bias vector $b_y \in R^{d_x}$, and matrix $W'$. The weight matrices of encoder and decoder are usually tied, in which $W' = W^T$.

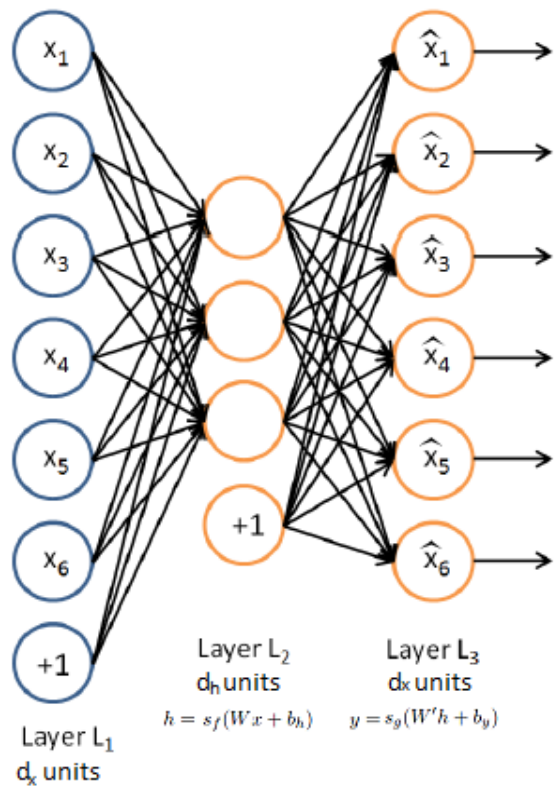

Figure 3.6: A typical autoencoder

Autoencoder training consists in finding parameters $\theta = \{W, b_h, b_y\}$ that minimize the reconstruction error on a training set of examples $D_n$, which corresponds to minimizing the following objective function:

$$J_{AE}(\theta) = \sum_{x \in D_n} L(x, g(f(x))) \tag{3.9}$$

Where $L$ is the reconstruction error. Typical choices include the squared error $L(x, y) = \|x - y\|^2$ used in cases of linear reconstruction [1]; and the cross-entropy loss when $s_g$ is the sigmoid and inputs are in $[0, 1]$: $L(x, y) = -\sum_{i=1}^{d_x} x_i log(y_i) + (1 - x_i) log(1 - y_i)$. We can use any kind of back-propagation optimization methods presented in Chapter 4 to train an autoencoder.

As shown above, the autoencoder tries to learn the function $g(f((x))) \approx x$. In other words, it is trying to learn an approximation to the identity function, so as to output $\hat{x}$ that is similar to $x$. However, by placing constraints on the network, such as by limiting the number of hidden units, we can discover interesting structure about the data. For example, if the number of hidden units (e.g. 50) is smaller than the number of input units (e.g. 100), the network is forced to learn a *compressed* representation of the input. Because it must try to reconstruct the input $x \in R^{100}$ from its hidden activations representation $h^{(2)} \in R^{50}$. If the input were completely random then this compression task would be very difficult. However, if there is structure in the data, for example, if some of the input features are correlated, then this model will be able to discover some of those correlations.

---

[1]This choice gives the hidden representation that very similar to PCA's

If the output units are linear (i.e. linear reconstruction) and the mean squared error criterion is used to train the network, then that simple autoencoder often ends up learning a low-dimensional representation very similar to PCA's.[2]. However, if the hidden units are non-linear, the autoencoder behaves very differencly from PCA, with the ability to capture multi-modal aspects of the input distribution [1].

An autoencoder with many hidden layers is called deep autoencoder. Deep autoencoder is a special type of deep neural network and useful in dimensionality reduction, which will be introduced shortly in the following section.

Besides, many different types of autoencoder have been introduced recently, which are extremely useful in deep learning and feature extraction. In this section, I will present concisely several most interesting kinds of autoencoder, which can be used in pre-training process (replacing the RBM model) or feature extraction. In the next chapter, I will discuss more about greedy layer-wise pre-training process, which is one of the most important invention in the area of deep learning.

### 3.3.2   Similarity between autoencoder and RBMs

The RBMs and basic classical autoencoders are very similar in their functional form, although their interpretation and the procedures used for training them are quite different [48]. More specifically, the deterministic function that maps from input to *hidden representation* is the same for both models. One important difference is that deterministic autoencoders use *real-valued mean* as their hidden representation whereas stochastic RBMs sample a binary hidden representation from that mean. However, after their initial pretraining, the way layers of RBMs are typically used in practice when stacked in a deep neural network is by propagating these real-valued means, which is more in line with the deterministic autoencoder interpretation. The reconstruction error of an autoencoder can also be seen as an approximation of the log-likelihood gradient in an RBM, in a way that is similar to the approximation made by using Contrastive Divergence updates for RBMs [2].

This similarity can explain why initializing a deep network by stacking autoencoders yields almost as good a classification performance as when stacking RBMs [3]. However, many different types of autoencoder have been introduced recently, which can outperform the RBM model in different benchmarks.

### 3.3.3   Sparse autoencoder

As shown above, without any constraint, the autoencoder will try to learn the identity function. To fix this problem, we can use different type of constraint. One common constraint used in basic classical autoencoder is limiting the number of hidden units. We can also add *weight − decay* into the objective function, which favors small weights by

---

[2]Principal component analysis (PCA) is a statistical procedure that uses orthogonal transformation to convert a set of observations of possibly correlated variables into a set of values of linearly uncorrelated variables called principal components. It is a very old procedure ( Karl Pearson, 1901) and usually used in dimensionality reduction - Wiki

adding the term $\lambda \sum_{ij} W_{ij}^2$

$$J_{AE+wd}(\theta) = ( \sum_{x \in D_n} L(x, g(f(x)))) + \lambda \sum_{ij} W_{ij}^2 \qquad (3.10)$$

However, this weight-decay constraint alone does not help much and must be used together with other constraints.

One very useful constraint is sparsity constraint on the hidden units, which can force autoencoders to discover interesting structure in the data, even if the number of hidden units is large [28]

Informally, we will think of a neuron as being "active" (or as "firing") if its output value is close to 1, or as being "inactive" if its output value is close to 0. We would like to constrain the neurons to be inactive most of the time.

To impose this constrain, we first calculate the average activation of hidden units over the training set. Let the activation of hidden unit $j$ when the network is given a specific input x is $h_j(x)$ and a dataset with m training examples, the average activation of hidden unit $j$ can be calculated by:

$$\hat{\rho}_j = \frac{1}{m} \sum_{i=1}^{m} h_j(x^{(i)}) \qquad (3.11)$$

We would like to (approximately) enforce the constraint

$$\hat{\rho}_j = \rho \qquad (3.12)$$

where $\rho$ is a sparsity parameter, typically a small value close to zero (say $\rho = 0.05$). In other words, we would like the average activation of each hidden neuron $j$ to be close to 0.05 (say). To satisfy this constraint, the hidden unit's activations must mostly be near 0.

To achieve this, we will add an extra penalty term to our optimization objective that penalizes $\hat{rho}_j$ deviating significantly from $\rho$. Many choices of the penalty term can give reasonable results. A typical one is the Kullback-Leibler (KL) divergence. KL-divergence is a standard function for measuring how different two different distributions are. This penalty function has the property that $KL(\rho\|\hat{\rho}_j) = 0$ if $\hat{\rho}_j = \rho$, and otherwise, it increases monotonically as $\hat{\rho}_j$ diverges from $\rho$. This sparsity penalty term can be calculated using following equation:

$$\sum_{j=1}^{d_h} KL(\rho\|\hat{\rho}_j) = \sum_{j=1}^{d_h} \rho log \frac{\rho}{\hat{\rho}_j} + (1 - \rho) log \frac{1 - \rho}{1 - \hat{\rho}_j} \qquad (3.13)$$

Our overall cost function is now:

$$J_{sparseAE}(\theta) = J_{AE}(\theta) + \beta \sum_{j=1}^{d_h} KL(\rho\|\hat{\rho}_j) \qquad (3.14)$$

where $J_{AE}(\theta)$ is as defined in basic classical autoencoder, and $\beta$ controls the weight of the sparsity penalty term. To incorporate the KL-divergence term into backpropagation optimization process, there is a simple-to-implement trick involving only a small code change, which is explained clearly in [28]

The sparse autoencoder is very good at learning useful representations/features from different input domains (such as image, audio,...), and therefore it is usually used as feature extractor. By first using sparse autoencoder to extract features from raw input, and then training a typical classifier such as a neural network or SVM on these features, we can get a very good result on different tasks such as MNIST and CIFAR-10. Figure 3.7 shows the filters that learned by sparse autoencoder when trained on MNIST dataset.

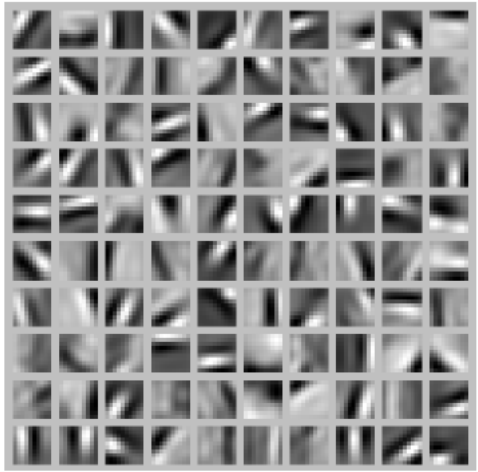

Figure 3.7: *Taken from [28]* Each square in the figure above shows the (norm bounded) input image x that maximally actives one of 100 hidden units. We see that the different hidden units have learned to detect edges at different positions and orientations in the image.

### 3.3.4 Denoising autoencoder

We have seen that the reconstruction criterion alone is unable to guarantee the extraction of useful features as it can lead to the obvious solution "simply copy the input" or similarly uninteresting ones. To overcome that problem, one strategy is to constrain the representation: the traditional bottleneck (i.e. limiting the number of hidden units) and the more recent sparse representations both follow this strategy.

Another different strategy introduced by Vincent et al. 2010 [48] is that we can change the reconstruction criterion for a both more challenging and more interesting objective: cleaning partially corrupted input, or in short *denoising*. Vicent et al. 2010 suggested that "a good representation is one that can be obtained robustly from a corrupted input and that will be useful for recovering the corresponding clean input". They expect that a higher-level representation should be rather stable and robust under corruptions of the input, and performing the denoising task well requires extracting features that capture useful structure in the input distribution.

Based on that strategy, Vincent et al. 2010 introduced a very simple variant of the basic autoencoder, which is called *denoising autoencoder* (DAE). DAE is trained to reconstruct a clean "repaired" input from a *corrupted* version of it. Each time a training example $x$ is presented, a different corrupted version $\hat{x}$ of it is generated according to a stochastic mapping $\hat{x} \sim q_D(\hat{x}|x)$.

Different corruption types can be considered. But the two most commonly used corruptions are additive isotropic Gaussian noise: $\hat{x} = x + \epsilon, \epsilon \sim N(0, \sigma^2 I)$ and binary masking noise, where a fraction $v$ of input components (randomly chosen) have their value set to 0. The degree of corruption ($\sigma$ or $v$) controls the degree of regularization.

The denoising autoencoder is able to learn useful features in different input domains like sparse autoencoder (Figure 3.8). Besides, we can stack many layers of denoising autoencoders to pre-train a deep neural network, similarly to the way we stack RBMs. This pre-training method even yields better results in many different datasets [48].

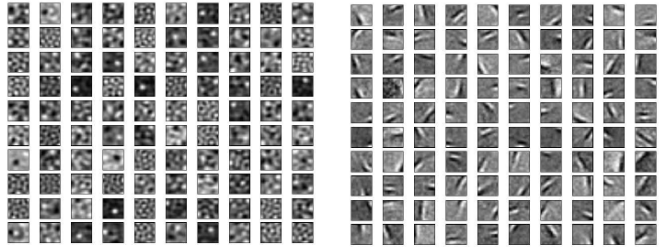

Figure 3.8: *Taken from [48]* Left: regular autoencoder with weight decay. The learned filters cannot capture interesting structure in the image. Right: a denoising autoencoder with additive Gaussian noise ($\sigma = 0.5$) learns Gabor-like local oriented edge detectors. Very similar to the filters learned by sparse autoencoder.

### 3.3.5 Contractive autoencoder

Salah Rifai et al. 2011 [39] proposed a new variant of autoencoder called Contractive autoencoder (CAE). They introduced a new penalty term, which encourages the hidden representation to be robust to small changes of the input *around the training examples*. They hypothesized that whereas the proposed penalty term encourages the learned features to be locally invariant without any preference for particular directions, when it is combined with a reconstruction error or likelihood criterion, we obtain invariance in the directions that make sense in the context of the given training data, i.e., the variations that are presented in the data should also be captured in the learned representation, but the other directions may be contracted in the learned representation [39].

If input $x \in R^{d_x}$ is mapped by encoding function $f$ to hidden representation $h \in R^{d_h}$, then to encourage robustness of the representation $f(x)$ for a training input $x$, they proposed the Frobenius norm of the Jacobian $J_f(x)$. This penalty is the sum of squares of all partial derivatives of the extracted features with respect to input dimensions:

$$\|J_f(x)\|_F^2 = \sum_{ij} (\frac{\partial h_j(x)}{\partial x_i})^2 \tag{3.15}$$

By penalizing $\|J_f(x)\|_F^2$, we want to keep all the first derivatives of the hidden units respect to input units to be small, which induces flatness in mapping function. In other words, we encourage the mapping to the feature space to be *contractive* in the neighborhood of the training data. The *flatness* will imply an *invariance* or *robustness* of the representation for small variations of the input.

The objective function for CAE is now:

$$J_{CAE}(\theta) = \sum_{x \in D_n} (L(x, g(f(x))) + \lambda \|J_f(x)\|_F^2) \tag{3.16}$$

Computing the new penalty and its gradient is similar to and has about the same cost as computing the reconstruction error and its gradient. Please check the [39] for more details.

The CAE has close relationship with weight decay, sparse AE, and DAE.

- *Weight decay*: The Frobenius norm of the Jacobian becomes the L2 weight decay in the case of a *linear encoder* (i.e. when $s_f$ is the identity function). In this case, $J_{CAE} = J_{AE+wd}$.

- *Sparse autoencoders*: sparse autoencoders encourage majority of the hidden representation close to zero. For these features to be close to zero, they must have been computed in the left saturated part of the sigmoid non-linearity, which is almost flat with a tiny first derivative. This yields a corresponding small entry in the Jacobian $J_f(x)$.

- *Denoising autoencoders*: Robustness to input perturbations was also one of the motivations of the denoising autoencoder. However, the CAE and the DAE differ in the way they encourage the robustness.

The CAE can be used as the building block in the layer-wise pre-training strategy, just like the DAE. It gives slightly better results on some datasets and settings. Figure 3.9 show performance comparison of different autoencoders for unsupervised pre-training a 1000-units one hidden layer network on MNIST and CIFAR10-bw (gray scale version of CIFAR10) datasets.

- RBM-binary: Restricted Boltzmann Machine trained by Contrastive Divergence,

- AE: Basic autoencoder,

- AE+wd: Autoencoder with weight-decay regularization.

- DAE-g: Denoising autoencoder with Gaussian noise,

- DAE-b: Denoising autoencoder with binary masking noise.

| | Model | Test error | Average $\|J_f(x)\|_F$ | SAT |
|---|---|---|---|---|
| **MNIST** | CAE | **1.14** | $0.73 \; _{10^{-4}}$ | 86.36% |
| | DAE-g | **1.18** | $0.86 \; _{10^{-4}}$ | 17.77% |
| | RBM-binary | 1.30 | $2.50 \; _{10^{-4}}$ | 78.59% |
| | DAE-b | 1.57 | $7.87 \; _{10^{-4}}$ | 68.19% |
| | AE+wd | 1.68 | $5.00 \; _{10^{-4}}$ | 12.97% |
| | AE | 1.78 | $17.5 \; _{10^{-4}}$ | 49.90% |
| **CIFAR-bw** | CAE | **47.86** | $2.40 \; _{10^{-5}}$ | 85,65% |
| | DAE-b | **49.03** | $4.85 \; _{10^{-5}}$ | 80,66% |
| | DAE-g | 54.81 | $4.94 \; _{10^{-5}}$ | 19,90% |
| | AE+wd | 55.03 | $34.9 \; _{10^{-5}}$ | 23,04% |
| | AE | 55.47 | $44.9 \; _{10^{-5}}$ | 22,57% |

Figure 3.9: *Taken from [39]* Results are sorted in ascending order of classification error on the test set. Best performer and models whose difference with the best performer was not statistically significant are in bold. Notice how the average Jacobian norm (before fine-tuning) appears correlated with the final test error. SAT is the average fraction of saturated units per example.

### 3.3.6 Deep AutoEncoders

A deep autoencoder is just an autoencoder with many hidden layers (Figure 3.10). This model is extremely useful in reducing the dimensionality of data, where we want to transform the high-dimensional data into a low-dimensional code. By reducing dimensionality, other tasks on high-dimensional data will become much easier such as classification, visualization, communication, or storage. The model has been introduced long ago but could only become popular by Hinton's discovery of pre-training strategy. This model can be considered as a nonlinear generation of Principal component analysis (PCA), which is a simple and widely used method for dimensionality reduction task. In [16], Hinton shown how a deep autoencoder could be trained using pre-training strategy Figure 3.10. Of course, we can replace RBM model in his method by another type of building block models.

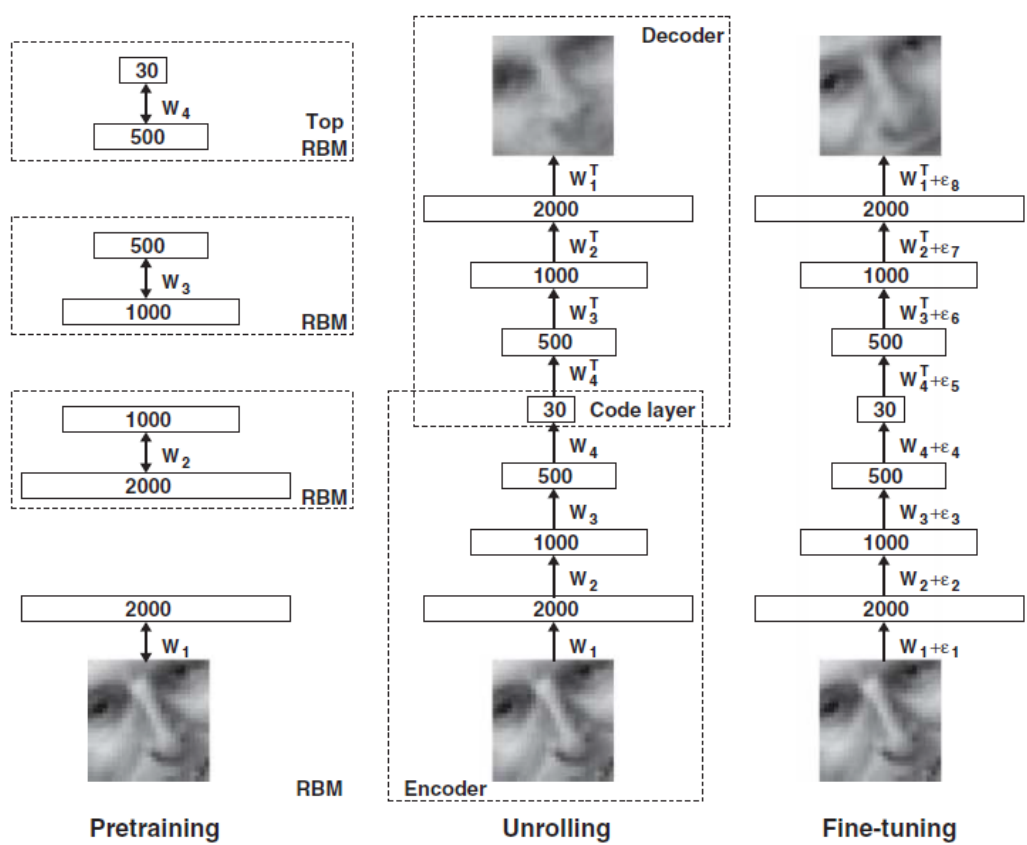

Figure 3.10: *(Taken from [16])* Pretraining consists of learning a stack of RBMs, each having only one layer of feature detectors. The learned feature activations of one RBM are used as the "data" for training the next RBM in the stack. After the pretraining, the RBMs are "unrolled" to create a deep autoencoder, which is then fine-tuned using back propagation of error derivatives

Figure 3.11 shows how a deep autoencoder produces much better compressed codes compared to PCA. This experiment conducted by Hinton et al. [16], in which a very deep autoencoder $784 - 1000 - 500 - 250 - 2$ was used on MNIST dataset. We can see that a two-dimensional autoencoder produced a better visualization of the data than did the first two principal components.

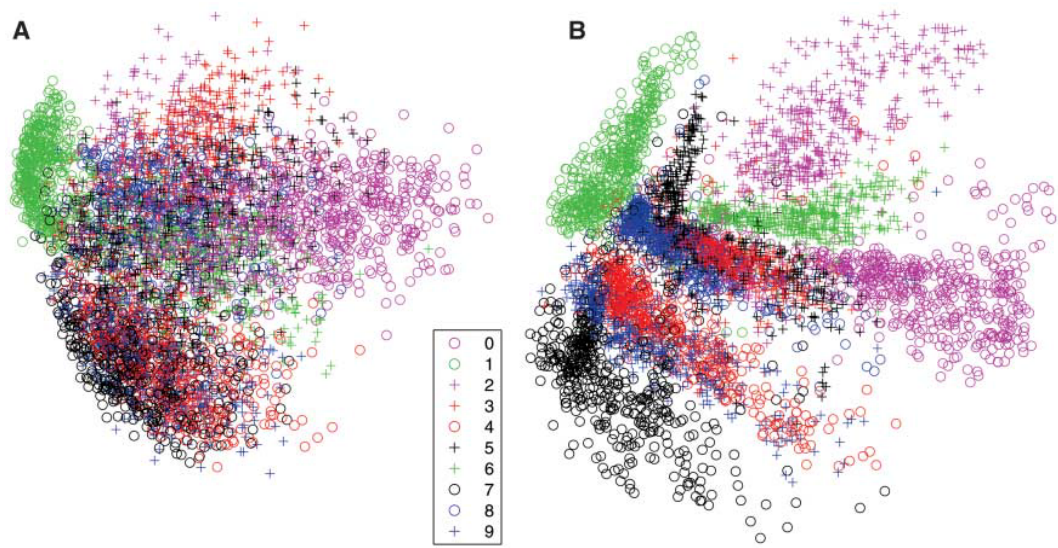

Figure 3.11: *(Taken from [16])* **(A)** The two dimensional codes for 500 digits of each class produced by taking the first two principal components of all 60,000 training images.**(B)** The two dimensional codes found by a $784 - 1000 - 500 - 250 - 2$ autoencoder

# Chapter 4

# Gradient-Based Training Algorithms for Neural Networks

In neural networks, optimization techniques play an important role, especially in supervised learning. They are used to explore the hypothesis space and find the best configuration, which minimizes an objective function (e.g. mean-square error, cross-entropy error...). First-order optimization methods such as Steepest Gradient Descent (SGD) used to be the most popular method due to their simple implementations and computational efficiency, compared to the second-order methods (e.g. Newton's methods). However, first-order methods are easy to be trapped in local minima and exhibit slow convergence, due to the problems described below. To help first-order methods overcome those problems, momentum-related techniques and well-designed initialization methods have been proposed recently [46]. These techniques are successfully experimented on training deep autoencoders and Recurrent Neural Networks (RNN), which was believed to be impossible (even with 2nd-order methods). These improvements could potentially change the "wrong" belief about the ability of first-order methods.

On the other hand, some of the second-other methods are computationally infeasible but show much better convergence characteristics because they take into account the curvature of the error space. The reason of its in-feasibility is the computation of the inverse of Hessian matrix ($O(n^3)$), which could be tackled by computing the inverse matrix directly (as in quasi-Newton's method); approximation of it (as in Gauss-Newton's method); or using an incomplete optimization (as in Hessian-Free method).

This chapter attempts to give a complete overview about all current well-known optimization techniques applied in Neural Network, their believed shortcomings as well as advantages.

## 4.1 First-order Methods

All first-order methods bases on the first-order Taylor series expansion to approximate the objective function $F$ (i.e. cost function) at current state of weight configuration $w$:

$$F(w + \Delta w) \approx F(w) + \nabla F(w)\Delta w$$
$$\nabla F(w) : \text{gradient vector of } F(w)$$

(4.1)

Therefore, if we go from the weight configuration $w$ to the next configuration at the direction of $\Delta w = -\alpha \nabla F(w)$ ($\alpha$ is learning rate), we will get:

$$F(w_{next}) = F(w - \alpha \nabla F(w)) \approx F(w) - \alpha \nabla F(w)^2$$
$$\leq F(w) \text{if } \alpha \text{ small enough}$$

The cost function is decreased as we update the weight. However, the reasoning presented here is approximate and it is true only for small enough learning rates.

### 4.1.1  Steepest Gradient Descent

Steepest Gradient Descent (SGD) is a standard first-order method, which directly uses the proof above. This method assumes that the cost function $F(w)$ decreases fastest if we go in the direction of negative gradient of $F(w)$.

The SGD converges slowly to the optimal solution, and the learning-rate parameter $\alpha$ has a profound influence on its convergence behaviors. If the learning rate is small, the method is more stable but converges slowly. If the learning rate is large, it could follows a zigzagging (oscillatory path) to the optimal solution, or even diverge if it is too large. However, the size of learning rates depends on the shape of the error plane at current point, so we could not easily determine the appropriate learning rates at different point of learning process. In practice, we usually set the learning rates larger at the beginning and decrease it down when reaching the optimal solution. The second-order methods tackle this problem by using the curvature information of the error plane to somehow "adjust" the learning rates.

### 4.1.2  Stochastic and Mini-Batches Gradient Descent

In steepest gradient descent (i.e. batch gradient descent) method, the gradient vector of the cost function $F$ is calculated by summing up the gradients of each training case. In other words, we have to go through the whole training set to make a single update on the weights matrix. Updating this way is inefficient if the training set is large, which is usually happens in practice. The stochastic (i.e. online) gradient descent approximates the gradient vector $\nabla F$ by calculate the gradient of each training case and use it to update the weights matrix right away. This method leads to a zigzagging searching path, because we are not following the steepest gradient of the error plane. The stochastic gradient descent tends to work better than the steepest gradient descent on large datasets where each iteration of gradient descent is very expensive. Besides, if each training case is used only one time, any new training cases are supplied every day, this method is capable of forgetting the old training case.

There is a compromise between the batch and stochastic method, which is often called "mini-batches", where the true gradient is approximated by a sum over a small number of training cases.

### 4.1.3  Momentum and Nesterov's Accelerated Gradient

The SGD method usually get trouble with plateaus or long narrow valleys (i.e. pathological curvature) in the objective function. It is because the gradient is too small (e.g. in plateau) or the curvature is too high (e.g. long narrow valley). The momentum method is used to accelerate the optimization along directions of low but persistent reduction (in plateau), similarly to the way the second-order methods accelerate the optimization along low-curvature directions (but second-order methods also decelerate the optimization along high curvature directions, which is not done by momentum methods) [45]

The momentum method maintains a velocity vector $v_t$ which is updated as follows:

$$v_{t+1} = \mu v_t - \alpha \nabla F(w_t)$$
$$w_{t+1} = w_t + v_{t+1}$$

The momentum decay coefficient $\mu \in [0; 1)$ controls the rate at which old gradients are discarded. Its physical interpretation is the "friction" of the surface of the objective function, and its magnitude has an indirect effect on the magnitude of the velocity [45].

A variant of momentum known as Nesterov's accelerated gradient (Nesterov, 1983) (NAG) has been analyzed with certain schedules of the learning rate and of the momentum decay coefficient $\mu$, and was shown by Nesterov (1983) to exhibit the better convergence rate $O(1/T^2)$ versus the $O(1/T)$ of SGD [46]. NAG formulas:

$$v_{t+1} = \mu v_t - \alpha \nabla F(w_t + \mu v_t)$$
$$w_{t+1} = w_t + v_{t+1}$$

While momentum method compute the gradient update from the current position $w_t$, NAG first performs a partial update to $w_t$, computing $w_t + \mu v_t$, but missing the yet unknown correction. This difference seems to allow NAG to change $v$ in a quicker and more responsive way, letting it behave more stably than momentum in many situations, especially for higher values of $\mu$ [46].

Choosing an appropriate schedule for the momentum $\mu$ and learning rate $\alpha$ is difficult and usually tuned in a heuristic way. As showed in [46], Ilya et. al. 2013 have trained deep autoencoders by using these following formulas:

$$\mu = min(1 - 2^{-1-log_2(\lfloor t/250 \rfloor + 1)}, \mu_{max}) \tag{4.2}$$

where $\mu_{max}$ was chosen from {0.999, 0.995, 0.99, 0.9, 0}. For each point of $\mu_{max}$, the learning rate $\alpha$ is chosen from {0.05, 0.01, 0.005, 0.001, 0.0005, 0.0001}. Besides, they found it beneficial to reduce $\mu$ to 0.9 during the final 1000 parameter updates of the optimization without reducing the learning rate. (see [46] for more information about momentum schedule and its effect on convergence rate and quality of final result)

### 4.1.4 Sparse Initialization

Convex objective functions $F(w)$ are insensitive to the initial parameter setting, since the optimization will always recover the optimal solution, merely taking longer time for worse initializations. But, given that most objective functions $F(w)$ that we want to optimize are non-convex, the initialization has a profound impact on the optimization and on the quality of the solution. Ilya et. al. 2013 shown that appropriate initializations play an even greater role than previously believed for both deep and recurrent neural networks [46]. Unfortunately, it is difficult to design good random initializations for new models, so it is important to experiment with many different initializations [45].

An initialization scheme that was successfully used to train deep and recurrent neural networks in [46] is called sparse initialization. In that initialization scheme, each random unit is connected to 15 randomly chosen units in the previous layer, whose weights are drawn from a unit Gaussian, and the biases are set to zero. The intuitive justification is that the total amount of input to each unit will not depend on the size of the previous layer and hence they will not as easily saturate. Meanwhile, because the inputs to each unit are not all randomly weighted blends of the outputs of many 100s or 1000s of units in the previous layer, they will tend to be qualitatively more "diverse" in their response to inputs.

### 4.1.5   Resilient BackPropagation

Resilient BackProbagation (Rprop) is a first-order local adaptive learning scheme, performing supervised batch learning in multi-layer neural networks [38]. The basic principle of Rprop is to eliminate the harmful influence of the size of the partial derivative on the weight step: small in plateaus and large in ravines. It uses only the sign of the derivative to indicate the direction of the weight update. The size of the weight change is exclusively determined by a weight-specific, so-called "update-value" $\Delta_{ij}$:

$$w_{ij}^{(t)} = \begin{cases} -\Delta_{ij}^{(t)} & \text{if } \frac{\delta E^{(t)}}{\delta w_{ij}} > 0 \\ +\Delta_{ij}^{(t)} & \text{if } \frac{\delta E^{(t)}}{\delta w_{ij}} < 0 \\ 0 & \text{else} \end{cases} \tag{4.3}$$

The second step of Rprop learning is to determine the new update-values $\Delta_{ij}^{(t)}$. This is based on a sign-dependent adaptation process:

$$\Delta_{ij}^{(t)} = \begin{cases} -\eta^+ * \Delta_{ij}^{(t-1)} & \text{if } \frac{\delta E^{(t-1)}}{\delta w_{ij}} * \frac{\delta E^{(t)}}{\delta w_{ij}} > 0 \\ +\eta^- * \Delta_{ij}^{(t)} & \text{if } \frac{\delta E^{(t-1)}}{\delta w_{ij}} * \frac{\delta E^{(t)}}{\delta w_{ij}} < 0 \\ \Delta_{ij}^{(t-1)} & \text{else} \end{cases} \tag{4.4}$$

Where $0 < \eta^- < 1 < \eta^+$. The $\eta^+$ is empirically set to 1.2 and $\eta^-$ to 0.5 [38] The adaptive-rule works as follows: every time the partial derivative of the corresponding weight $w_{ij}$ changes its sign, which indicates that the last update was too big and the algorithm has jumped over a local minimum, the update-value $\Delta_{ij}^{(t)}$ is decreased by the factor $\eta^-$. If the derivative retains its sign. The update value is slightly increased in order to accelerate convergence in shallow regions.

Generally, the Rprop method converges faster than standard gradient descent method. Besides, we do not have to set any meta-parameters (such as learning rates) for Rprop to obtain optimal convergence times.

Another interesting property of Rprop is that the size of the weight-step is only dependent on the sequence of signs, not on the magnitude of the derivative. Therefore, learning is spread equally all over the entire networks: weights near the input layer have the equal chance to grow and learn as weights near the output layer. This property can help to overcome the vanishing gradient problem when training deep neural networks.

## 4.2   Second-order Methods

All second-order methods base on the idea of minimizing the quadratic approximation of the cost function (using the second-order Taylor series expansion). However, each of these methods below are different on how they calculate (or approximate) the inversion of Hessian matrix.

### 4.2.1 Newton's Methods

This is the standard second-order methods. Specifically, using a second-order Taylor series expansion of the cost function $F()$ around the point $w$, we will have:

$$F(w + \Delta w) \approx F(w) + \nabla F(w)\Delta w + \frac{1}{2}\Delta w^T H \Delta w \tag{4.5}$$

With $H$ is the Hessian (i.e. curvature) matrix of $F(w)$.

We attain the extreme of $F$ when its derivative with respect to $\Delta w$ is equal to 0. Therefore, we have to solve the following equation:

$$\nabla F(w) + H\Delta w = 0$$
$$\Leftrightarrow \Delta w = -H^{-1}\nabla F(w)$$

However, finding the inverse of the Hessian in high-dimensional space can be an expensive operation $O(n^3)$. In such cases, instead of directly inverting the Hessian, we can calculate it as a solution of the system of linear equations: $H\Delta w = -\nabla F(w)$. This can be solved by using iterate methods like conjugate gradient (described below, used in Hessian-Free method). However, conjugate gradient requires Hessian matrix be a positive definite matrix, which is not guaranteed during training process.

Besides, we can also approximate the inversion of Hessian matrix directly from changes in the gradient, which is used in quasi-Newton's methods (e.g. DFP, BFGS, L-BFGS, ...).

### 4.2.2 Gauss-Newton's method

The Gauss-Newton algorithm is a method used (and can only be used) to solve non-linear least squares problems and can be seen as a modification of Newton's method. Instead of computing the Hessian matrix, it uses the Jacobian matrix to approximate it. Therefore, this method has the advantage that second derivatives, which can be challenging to compute, are not required.

In least squares problem, given $m$ training examples which their error cost are calculated by $m$ functions $r_{1..m}$ we have to find the minimum of the sum of squares:

$$F(w) = \sum_{i=1}^{m} r_i^2(w) \tag{4.6}$$

Differentiating the above equation with respect to $w_j$ (an element of w), we have the elements of gradient vector of $F()$:

$$\nabla F(w)_j = 2\sum_{i=1}^{m} r_i \frac{\partial r_i}{\partial w_j} \tag{4.7}$$

Differentiating the above gradient element respect to $w_k$, we could get the elements of the Hessian matrix:

$$H_{jk} = 2\sum_{i=1}^{m} (\frac{\partial r_i}{\partial w_j}\frac{\partial r_i}{\partial w_k} + r_i\frac{\partial^2 r_i}{\partial w_j \partial w_k}) \tag{4.8}$$

The Gauss-Newton method is obtained by ignoring the second-order derivative terms (the second term in above expression). So the Hessian is approximated by:

$$H_{jk} = 2\sum_{i=1}^{m} (\frac{\partial r_i}{\partial w_j}\frac{\partial r_i}{\partial w_k}) = 2\sum_{i=1}^{m} J_{ij}J_{ik} \tag{4.9}$$

where $J_{ij} = \frac{\partial r_i}{\partial w_j}$ are entries of the Jacobian $J_r$ matrix. Therfore, Gauss-Newton's methods approximate Hessian by:

$$H \approx 2J_r^T J_r \tag{4.10}$$

This approximation of H is usually called as Gauss-Newton matrix G, and this is used in Hessian-Free method instead of H matrix because it is guaranteed to be positive semi-definite, which avoids the problem of negative curvature.

### 4.2.3    Quasi-Newton's methods

In quasi-Newton methods, the Hessian matrix does not need to be computed. The Hessian is updated by analyzing successive gradient vectors instead. The most common quasi-Newton algorithms are currently the SR1 formula (for symmetric rank one), the BHHH method, the widespread BFGS method (suggested independently by Broyden, Fletcher, Goldfarb, and Shanno, in 1970), and its low-memory extension, L-BFGS.

If we call $B$ is an approximation of Hessian, apply the first-order Taylor series expansion on the gradient of cost function $F()$, we have the following secant equation:

$$\nabla F(w + \Delta w) = \Delta F(w) + B\Delta w \tag{4.11}$$

The various quasi-Newton methods differ in their choice of the solution to the secant equation. Most methods (but with exceptions, such as Broyden's method) seek a symmetric solution $B^T = B$

### 4.2.4    Conjugate gradient method

In mathematics, the conjugate gradient (CG) method is an iterative method to solve the systems of linear equations. In Hessian-Free optimization method, they use conjugate gradient method to solve the linear equation: $H\Delta w = -\nabla F(w)$. Instead of solving this directly, CG method tries to minimize this objective function, which becomes smaller when we come closer to the solution.

$$E(\Delta w) = \frac{1}{2}\Delta w^T H \Delta w - \Delta w^T b \tag{4.12}$$

Therefore, we have to find $\Delta w$ that minimize the function $E()$. Briefly, CG finds the solution by using a sequence of steps, each of which finds the minimum along one direction. Besides, it makes sure that new direction is "conjugate" to the previous directions so you do not mess up the minimization you already did. In fact, the "conjugate" means that as you go in the new direction, you do not change the gradients in the previous directions.

### 4.2.5    Hessian-Free optimization method

HF differs from other Newton's methods only because it is performing an incomplete optimization (via un-converged Conjugate gradient) of approximation of $F()$ [26].

The first main thing about Hessian-Free optimization is that it uses conjugate gradient method to find the $\Delta w$ (direction to go) from the linear equation $H\Delta w = -\nabla F(w)$ instead of computing the inversion of Hessian matrix (directly or indirectly). Besides, in this method, we do not wait for the CG to totally converge.

The second thing is that we use Gauss-Newton matrix G (instead of Hessian) because it is guaranteed to be positive semi-definite, and experimentally showed to be consistently better than H.

The third vital important thing in HF method is its damping technique (also called structural regularization). HF method could find a direction with extremely low curvature and will elect to move very far along it, and possibly well outside of the region where Taylor series expansion is a sensible approximation. Therefore, we add the damping parameter $\lambda$ to control how "conservative" the approximation is, by adding the constant $\lambda \|d\|^2$ to the curvature estimate for each direction $d$. we can also use the "Levenberg-Marquardt" style heuristic for adjusting $\lambda$ over time.

The final important thing is about the sparse initialization scheme described above.

## 4.3   Line Search vs. Trust Region Strategy

In iterative optimization techniques (including all type of first-order and second-order methods), there is two main strategies: line search and trust region. In *line search* approach, one first finds the descent direction (of the objective function $f$) and then computes an appropriate step size. In *trust region* approach, one first determines a step size (the size of the trust region) and then finds the step direction. Generally, the line search approach is usually used to adapt the learning rate in first-order methods, while the trust region approach is used as a damping/regularization technique in second-order methods.

### 4.3.1   Line Search

In line search approach, the step direction is first computed by other methods, such as gradient descent, Newton's method, or Quasi-Newton method.... The step size then can be determined either exactly or inexactly using many different rules. In [43], Shi 2004 has summarized and analyzed seven different line search rules. In this section, we provide the basic ideas of the four most popular rules, which are: minimization rule, Limited minimization rule, approximate minimization rule, and Armijo rule.

Assume that $F$ is our objective function, $w_k$ is weights configuration at step $k^{th}$, $\Delta w_k$ is chosen direction at step $k^{th}$, $\alpha_k$ is step size (or learning rate if using with first-order methods). The $\alpha_k$ can be computed exactly or inexactly using these following rules:

1. *Minimization rule*: In this rule, we try to find the step size $\alpha_k$ that can minimize the objective function $F$ in chosen direction $\Delta w_k$. This rule is implicitly used by Conjugate gradient method.

$$\alpha_k = \operatorname{argmin}_{\alpha > 0} F(w_k + \alpha \Delta w_k) \tag{4.13}$$

2. *Limited minimization rule*: This rule is similar to the first rule, except that we limit the maximum step size by $s_k = -\frac{\nabla F(\Delta w_k)^T \Delta w_k}{\|\Delta w_k\|^2}$. Note that $\nabla F(\Delta w_k)^T \Delta w_k$ is the expected decrement in $F$ if the step is $\Delta w_k$, approximated by first order Taylor series.

$$\alpha_k = \operatorname{argmin}_{\alpha \in [0, s_k]} F(w_k + \alpha \Delta w_k) \tag{4.14}$$

3. *Approximate minimization rule*: In this rule, we try to go as far as possible, as long as the objective function is still getting decreased.In the minimization rule, we go directly to the global minima along the chosen direction of the objective function, while in this rule, we go to the nearest local minima on that direction.

$$\alpha_k = \min(\alpha | \nabla F(w_k + \alpha \Delta w_k)^T \Delta w_k = 0, \alpha > 0) \qquad (4.15)$$

4. *Armijo rule*: This rule ensures that the step length $\alpha_k$ decreases $F$ "sufficiently", by using this inequality:

$$F(w_k) - F(w_k + \alpha_k \Delta w_k) \geq -c_1 \alpha_k \nabla F(w_k)^T \Delta w_k \qquad (4.16)$$

This inequality ensures that the real decrement in $F$ is bigger than a proportion $c_1$ of the expected decrement. $c_1$ is usually chosen to be quite small, say $10^{-4}$. However, because the above inequality is always satisfied when $\alpha$ is small enough. Therefore, to prevent choosing too small step size, we usually add the following *curvature condition* to ensures that the slope has been reduced sufficiently.

$$\nabla F(w_k + \alpha_k \Delta w_k)^T \Delta w_k \geq c_2 \nabla F(w_k)^T \Delta w_k \qquad (4.17)$$

In practice, besides using line search to control the learning rate in first-order methods, the learning rate is also usually set by $\alpha_k = \frac{a}{b+k}$, with $a, b$ are predefined constants.

### 4.3.2  Trust Region

In almost every iterative optimization techniques, the objective function is approximated using a certain model function (e.g. quadratic in second-order methods). However, that approximation is only good (trusted) in a limited region around the sample point - the trust region. The trust region approaches restricts the step size by first compute the trust region, and then find the best direction within that region. The trust region is expanded when the approximation is fit the objective function well, and contracted otherwise. This is also known as the restricted step method. The model fit is usually evaluated by comparing the ratio of expected improvement from the model approximation with the actual improvement observed in the objective function.

This method is usually used with the second-order methods such as Newton's, Gauss-Newton or Hessian-Free optimization. Generally, in second-order methods, we find the descent direction $\Delta w$ by solving the following equation:

$$H\Delta w = \nabla F(w) \qquad (4.18)$$

With H is the Hessian matrix or its approximation. To restrict the step size, the trust region method instead solves the following equation:

$$(H + \lambda I)\Delta w = \nabla F(w) \qquad (4.19)$$

With $I$ is the identity matrix, and $\lambda$ is the damping parameter that controls the trust-region size. Geometrically, that term adds a paraboloid centered at $\Delta w = 0$ to the quadratic form, resulting in a smaller step. If the $\lambda$ is large enough, the Hessian matrix will be ignored, and the step will be taken approximately in the direction of the gradient.

In Levenberg-Marquardt algorithm, Marquardt replace the identity matrix $I$ with the diagonal matrix consisting of the diagonal elements of $H$. This approach can scale each component of the gradient according to the curvature, so that there is larger movement along the directions where the gradient is smaller:

$$(H + \lambda \text{diag}(I))\Delta w = \nabla F(w) \tag{4.20}$$

The damping factor $\lambda$ is adjusted by looking at the ratio $\rho = \frac{\Delta F_{actual}}{\Delta f_{pred}}$. In [26], Martens 2010 has successfully used this damping method (using identity metrix) with Hessian-Free optimization to train deep autoencoders. He used Levenberg-Marquardt style heuristic for adjusting $\lambda$ directly: **if** $\rho < \frac{1}{4} : \lambda \leftarrow \frac{3}{2}\lambda$ **elseif** $\rho > \frac{3}{4} : \lambda \leftarrow \frac{2}{3}\lambda$

## 4.4 Initialization Analysis

In [9], X. Glorot and Y. Bengio 2010 have provided a profound analysis about the influence of the non-linear activations functions, cost function types, and initialization methods, on using standard gradient descent for training deep neural networks. They showed that the logistic sigmoid activation is unsuited for deep networks with random initialization because of its mean value, which can drive especially the top hidden layer into saturation. Besides, they found that the logistic regression or conditional log-likelihood cost function coupled with softmax outputs worked much better (for classification problems) than the quadratic cost. Regarding the initialization, they explained why the standard random initialization could lead to vanishing gradient problem in training deep neural networks.

In the standard random initialization, the biases is initialized to be 0, and the $W_{ij}$ at each layer is sampled from the uniform distribution: $U[-\frac{1}{\sqrt{n}}, \frac{1}{\sqrt{n}}]$, where $n$ is the size of the previous layer and $U[-a, a]$ is the uniform distribution in the interval $(-a, a)$. They proved that standard initialization would lead to variance of the weights with the following property:

$$n_i Var[W^i] = \frac{1}{3} \tag{4.21}$$

Where $n_i$ is the layer size (assuming that all layers have the same size), and $W^i$ is weights of layer $i^{th}$. This will cause the variance of the back-propagated gradient to be dependent on the layer (and decreasing). We prefer the following conditions, which can keep information flowing:

$$\forall i, n_i Var[W^i] = 1$$
$$\forall i, n_{i+1} Var[W^i] = 1$$

To approximately satisfy the objectives of maintaining activation variances and back-propagated gradients variance as one moves up or down the network, X. Glorot and Y. Bengio 2010 have proposed the following *normalized initialization*:

$$W \sim U[-\frac{\sqrt{6}}{\sqrt{n_j + n_{j+1}}}, \frac{\sqrt{6}}{\sqrt{n_j + n_{j+1}}}] \tag{4.22}$$

## 4.5 Cascade correlation algorithm

Cascade-Correlation is a supervised learning algorithm for neural networks. Instead of just adjusting the weights in a network of fixed topology, Cascade-Correlation begins with a minimal network, then automatically trains and adds new hidden units one by one, creating a multi-layer structure [8]. Once a new hidden unit has been added to the network, its input-side weights are frozen. This unit then becomes a permanent feature-detector in the network, available for producing outputs or for creating other, more complex feature detectors. This approach has several advantages: it learns very quickly, the network determines its own size and topology, it retains the structure it has built even if the training set changes, and it requires no back-propagation of error signals through the connections of the network ([8]).

# Chapter 5

# Overfitting and Regularization

## 5.1 Introduction

In general, by training neural networks using a set of examples with input and output pattern (i.e. supervised learning), we are trying to construct a mapping that defines the output pattern in terms of the input patterns. However, the information content of the training examples is ordinarily not sufficient itself to reconstruct that mapping, which leads to the possibility of overfitting [14].

Hadamard (1902) has defined the term "well-posed" to indicate that whether a problem could be solved on a computer using a stable (e.i reproducible and unique result) algorithm or not. If a problem is not "well-posed", it is said to be ill-posed. The problem of reconstructing the mapping $f$ between input and output is said to be well-posed if Hadamard's three conditions are satisfied:

- Existence: the mapping function "f" exists.

- Uniqueness: the mapping function is unique.

- Continuity: slight change in the input only make a limited change in the output.

However, in the context of supervised learning, Hadamard's conditions are violated for the following reasons [14]:

- The mapping function may not exist (a certain output may not exist for any input).

- We typically do not have as much information from training examples as we need to reconstruct an unique mapping.

- The unavoidable presence of noise in training examples could lead to the violation on the continuity criterion.

There is no way to overcome these difficulties unless some prior information about the input-output mapping is available. In fact, when we are constructing a neural network for a specific task (e.g. classification, regression . . . ), we have already made some assumptions (i.e. prior information, inductive bias) about what the solution should be, such as: number of hidden units, type of units, number of layers. . . Regularization is also a set of methods to embed prior information into the learning process. The most common form of prior information involves the assumption that the input-output mapping function is smooth (i.e. similar inputs produce similar outputs) and simple (Occam's Razor).

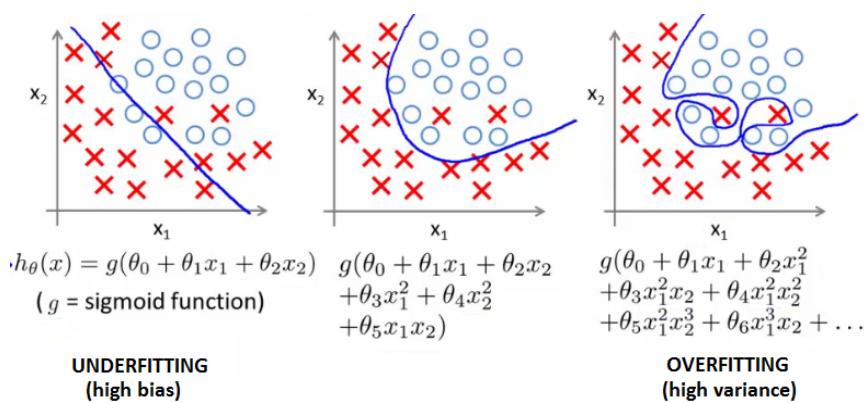

Figure 5.1: Underfitting and Overfitting in classification task, taken from [30]

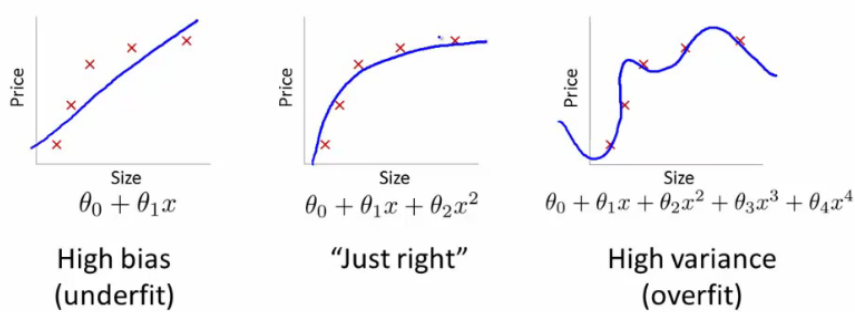

Figure 5.2: Underfitting and Overfitting in regression task, taken from [30]

### 5.1.1   Overfitting - What Is It?

The overfitting problem happens when a model is trained to fit the training data so well that it loses the generality properties (ability to predict the output for an unseen input). The training data contains information about the regularities in the mapping from input to output. However, it also contains sampling error, which comes from the way the particular training cases are chosen (not representative), or the technique we use to collect these training cases. Therefore, when we fit the model to the training data, it cannot tell which regularities are real and which are caused by sampling error. So if the model is very flexible (i.e. powerful), it can model the sampling error really well, and fail to generalize to unseen examples.

In Figure 5.1 and 5.2 [30], you can see how the models classify or fit the training data very well in overfitting cases, but they are probably not the good solutions. In general, we prefer the solution that is "smooth", simple, and able to explain the training data well enough, rather than a complex well-fitted solution. This inductive bias can be explained using the Occam's Razor, which will be described later.

On the other hand, the underfitting problem happens when the model cannot fit the training data. It happens because the model has low representational power compared to complexity of the problem, or because the learning algorithm gets stuck (e.g. in local minima or ravine). This is a serious problem when training deep or recurrent neural networks which are believed to be hard to train, using standard learning methods.

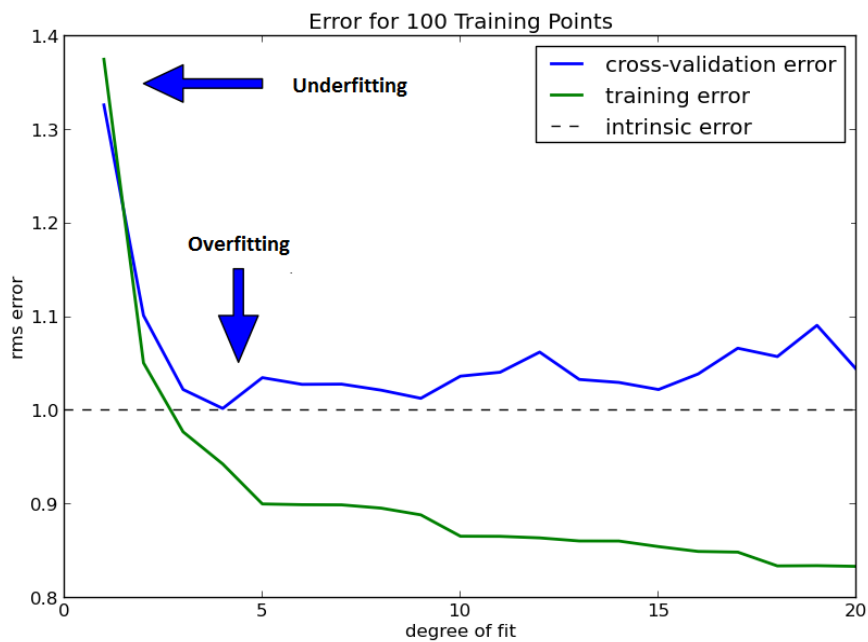

Figure 5.3: Error curves in cross validation methods

### 5.1.2 How to Detect Overfitting?

A common method to detect overfitting is to use validation set [27, 30]. Firstly, we split the whole data set into three parts: training data, validation data, and test data (typically 60%, 20%, and 20% respectively). Then, we use the training data to train the model, use the validation data to tune meta-parameters (i.e. learning rate, momentum...), and use test data to test the real performance of model over unseen examples. To detect overfitting, we could slot the two curves: training error and validation error (error calculated on validation set) as shown in Figure 5.4. Clearly, the overfitting problem happens when the validation error start to increase while the training error is still getting improved.

### 5.1.3 How to Solve Overfitting?

There are many ways to prevent overfitting.

1. *Get more data.* This is the best way to overcome overfitting, because more data probably provides more information about solution and weaken the influence of sampling error. However, collecting data is typically costly (in terms of time and labor). Besides, training with more data would require more computational power, which could be impossible in some cases.

2. *Combine different models.* We can learn many models with different forms, or train the model on different subsets of the training data (i.e. bagging), and average predictions from these models.

3. *Limit the model capacity (i.e. representational power)* to enough to fit the true regularities and not enough to fit the spurious regularities (sampling errors).

We can control the capacity of a neural networks in many ways:

1. *Early stopping:* Start with small weights and stop learning before it overfits.

2. *Architecture:* Limit the number of hidden layers or the number of units per layer. Besides, weights in the network can be shared as in convolutional neural network

3. *Weight-decay:* Penalize large weights using penalties of constraints on their squared values (L2 penalty) or absolute values (L1 penalty).

4. *Noise:* Add noise to the input, weights, or the node activities.

5. *DropOut and DropConnect:* Randomly selected subsets of activations or weights are set to zero within each layer.

### 5.1.4   From the Occam's Razor Point of View

Occam's (or Ockham's) razor is a principle attributed to the 14th century logician and Franciscan friar William of Ockham [20]. Ockham was the village in the English county of Surrey where he was born. The principle states that *"Entities should not be multiplied unnecessarily"*. Many scientists have adopted or reinvented Occam's Razor, as in Leibniz's "identity of observables" and Isaac Newton stated the rule: *"We are to admit no more causes of natural things than such as are both true and sufficient to explain their appearances"*. Stephen Hawking writes in A Brief History of Time: *"We could still imagine that there is a set of laws that determines events completely for some supernatural being, who could observe the present state of the universe without disturbing it. However, such models of the universe are not of much interest to us mortals. It seems better to employ the principle known as Occam's razor and cut out all the features of the theory that cannot be observed"*.

The most useful statement of the principle for scientists is *"when you have two competing theories that make exactly the same predictions, the simpler one is the better"*, or a stronger form which is relevant to machine learning fields: *"If you have two theories that both explain the observed facts, then you should use the simplest until more evidence comes along"*.

In machine learning, Occam's razor can be viewed as an inductive bias during the learning process. This theory explains why we prefer a simple relatively fitted model than a complex well-fitted model over the training data.

### 5.1.5   From the Bayesian point of view

From the Bayesian point of view, many regularization techniques correspond to imposing certain prior distributions on model parameters. For example, the L2 regularization assumes that prior distributions of weights in neural network are zero-mean Gaussian.

Figure 5.4: Posterior distribution

The Bayesian framework assumes that we always have a prior distribution for everything, but this prior may be very vague. When we observe some data, we combine our prior distribution with a likelihood term to get a posterior distribution. The likelihood term takes into account how probable the observed data to be predicted by the model. During the learning process, the likelihood term will fights against the prior and with enough data, the likelihood terms always wins. However, if we do not have enough data, prior distribution will keep the solution reasonable.

An easy example is about tossing coin when we try to predict the probability $p$ of producing head of a coin. Suppose we observe 100 tosses and there are 53 heads, so the maximum likelihood answer (the value of $p$ that makes the observation of 53 heads and 47 tails) will be $p = 0.53$. However, what if we only tossed the coin one and we got 1 head? Clearly, $p = 1$ is not a good answer because we do not have enough information. A better solution for this problem is to imply a prior distribution for the coin, such as uniform or 0.5-mean Gaussian distribution. We then combine the probability of observing a head with that prior distribution to get the posterior distribution. Therefore, choosing the right prior information could help the maximize posterior method overcome the overfitting in maximize likelihood method when lacking of training data.

## 5.2 Regularization Overview

Regularization, in mathematics and statistics and particularly in the fields of machine learning, refers to a process of introducing prior information in order to solve an ill-posed problem or to prevent overfitting. In machine learning, regularization techniques are usually used to constrain the weights of network to be small (L2 method), sparse (L1 method), or shared over different parts (Convolutional neural network). Besides, we can also add noise to weight or node activities. New published methods such as Dropout or Dropconnect could also do the job particularly successfully in many cases.

### 5.2.1 Least-Squared Method as Regularization

When using the Least-Squared cost function to maximize the likelihood between model's predictions and the target values, we are actually doing a simple regularization with the prior information is that the target solution is generated by adding the Gaussian noise to the output of the neural network ([27, 15]).

Suppose that we have $y_c = f(input_c, W)$ is the output of the net and $t_c$ is the target value. Therefore, the probability density of the target value is given by the network's output plus Gaussian noise:

$$p(t_c|y_c) = \frac{1}{\sqrt{2\pi\sigma^2}} e^{-\frac{(t_c - y_c)^2}{2\sigma^2}}$$

$$p(t|y) = \prod_{c \text{ examples}} \frac{1}{\sqrt{2\pi\sigma^2}} e^{-\frac{(t_c - y_c)^2}{2\sigma^2}}$$

$$\ln p(t|y) = \sum_c \ln \frac{1}{\sqrt{2\pi\sigma^2}} - \frac{(t_c - y_c)^2}{2\sigma^2}$$

Therefore, minimizing the squared error is the same as maximizing the log probability under a Gaussian noise.

$$W_{ML} = \text{argmax}_W \sum_c \ln \frac{1}{\sqrt{2\pi\sigma^2}} - \frac{(t_c - y_c)^2}{2\sigma^2}$$

$$= \text{argmin}_W \sum_c \frac{(t_c - y_c)^2}{2\sigma^2}$$

### 5.2.2   L2 regularization

L2 regularization is a standard weight penalty method, which is widely used in least-squared cost function. It adds an extra term to the cost function that penalizes the squared weights. However, there is also a similar regularization term which can be used with cross-entropy cost function (in case the output unit is logistic or softmax node).

$$C = E + \frac{\lambda}{2} \sum_i w_i^2$$

$$C : \text{ Final cost function}$$

$$E : \text{ Squared error - Likelihood term}$$

$$\frac{\lambda}{2} \sum_i w_i^2 : \text{ L2 regularization term}$$

The L2 regularization attempt to keep the weights small unless they have big error derivatives. This method prevents the network from using weights that it does not need which can improve generalization and make a smoother model because it helps to stop the network from fitting the sampling error. For example, if the network has two very similar inputs, it prefers to put half the weight on each rather than all the weight on one.

From the Bayesian point of view, this regularization method is equivalent to assuming a zero-mean Gaussian prior for the network's weights [15].

$$\ln p(W|D) = \ln p(D|W) + \ln p(W) - \ln p(D)$$

$$\implies W_{ML} = \text{argmin}_W \frac{1}{2\sigma_D^2} \sum_c (t_c - y_c)^2 + \frac{1}{2\sigma_W^2} \sum_i w_i^2$$

$$W : \text{ Network weights}$$

$$D : \text{ Training data}$$

$$\ln p(D) : \text{Independent from W}$$

The first term in above equation come from the assumption that the model makes a Gaussian prediction. And the second term assumes a zero-mean Gaussian prior for the weights.

### 5.2.3  L1 regularization

Sometimes, it works better to penalize the absolute values of the weights instead of the squared. This is called L1 regularization.

$$C = E + \frac{\lambda}{2} \sum_i |w_i| \tag{5.1}$$

This method is widely used in sparse modeling, a popular and effective model using in image processing. It pushes many weights in network to become exactly equal to zero, which yields sparse models - easier to interpret. Besides, the L1 also outperform the L2 penalty when irrelevant features are presented in training data, because it can learn to completely ignore them.

However, this L1 regularization term makes the cost function in equation above non-differentiable. Thus, we cannot use the standard optimization method like gradient descent to find the global minimum in the same way that is done in L2 penalty.

Besides, L2 and L1 penalty, sometimes, we can use different weight penalty that allows large weights but pushes small weights to become zero.

### 5.2.4  Weight constraints

Instead of penalize the squared weights separately; we can put a constraint on the maximum squared length of the incoming weight vector of each unit. If an update violates this constraint, we scale down the vector of incoming weights to allowed length. This method has several advantages over weight penalties: It is easier to set a sensible threshold and can prevent hidden units getting stuck near zero as well as weights exploding [15]. This is more effective than a fixed penalty at pushing irrelevant weights towards zero. Besides, using a constraint rather than a penalty prevents weights from growing very large no matter how large the proposed weight-update is. This makes it possible to start with a very large learning rate which decays during learning, thus allowing a far more thorough search of the weight-space than methods that start with small weights and use a small learning rate [19].

This method has been used together with dropout in [19] and gave a very good results on many different applications (see Dropout section).

### 5.2.5  Adding noise

In fact, adding Gaussian noise to the inputs is equivalent to using L2 regularization. We have the variance of the noise is amplified by the squared weight before going into the next layer as showed in Figure 5.5. This makes an additive contribution to the squared error, so minimizing the squared error tends to minimize the squared weights when the inputs are noisy.

However, adding Gaussian noise to the weights of a multilayer non-linear neural network is not exactly equivalent to using an L2 penalty and could be better especially in recurrent networks. Alex Graves showed that recurrent networks significantly better if noise is added to the weights [12].

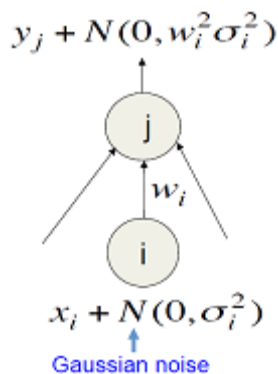

Figure 5.5: Adding Gaussian noise to inputs, taken from [15]

Another way is to add noise to the node activities [15]. Suppose that we have a multilayer neural network composed of logistic units and trained by back propagation. We can make the logistic units binary and stochastic (sampled from its activities) on the forward pass, but do the backward pass as in original back propagation method. This way produce the worse result on training set, and require considerably more time to train, but it does significantly better on test set.

### 5.2.6    Convolutional neural network

When applying fully-connected multilayer neural network on learning complex high-dimensional non-linear mapping such as image recognition or speech recognition, we traditionally have to use hand-designed feature extractor to gathers relevant information from the input and eliminates irrelevant variabilities. Feeding "raw" inputs directly into the network and let it learn feature extractor automatically is more interesting but causes many problems [24]. Firstly, because typical images or spoken words contain at least several hundred variables. Therefore, a first fully-connected layer with, say a few 100 units, would already contain several 10,000 weights, which leads to overfitting if the training data is scarce. Secondly, unstructured nets have no built-in invariance with respect to translations, or local distortions of the inputs (e.g. size, slant, or position variations). In principle, a fully-connected network of sufficient size could learn to produce outputs that are invariant with respect to such variations. However, learning such a task would probably result in multiple units with identical weight patterns positioned at various locations in the input. Besides, learning these weight configurations requires a very large number of training instances to cover the space of possible variations [24]. Thirdly, a fully-connected architectures entirely ignore the topology of the input (i.e. the input variables can be presented in any fixed order without affecting the outcome of the training).

The CNNs overcome these above problems by designing a network architecture that contains a certain amount of a prior knowledge about the problem [6]. CNNs combine three architectural ideas to ensure some degree of shift and distortion invariance: local receptive fields, shared weights (or weight replication) and spatial or temporal subsampling [24], which give these following advantages:

1. *Local receptive fields*: neurons can extract elementary visual features such as oriented edges, end-points, corners. This comes from the Hubel and Wiesel's discovery of locally-sensitive, orientation-selective neurons in the cat's visual system.

2. *Shared weights*: the elementary feature detectors that are useful on one part of the image are likely to be useful across the entire image. This knowledge can be applied by forcing a set of units, whose receptive fields are located at different places on the image, to have identical weight vectors. The outputs of such a set of neurons constitute a *feature map*. In order words, units in a feature map are constrained to perform the same operation on different parts of the image. A convolutional layer is usually composed of several feature maps (with different weight vectors), so that multiple features can be extracted at each location.

3. *Subsampling*: once a feature has been detected, its exact location become less important, as long as its approximate position relative to other features is preserved. Therefore, each convolutional layer is followed by an additional layer that performs a local averaging, and a subsampling, reducing the resolution of the feature map, and reducing the sensitivity of the output to shifts and distortions. Successive layers of convolutions and subsampling are typically alternated, resulting in a "bi-pyramid": at each layer, the number of feature maps is increased as the spatial resolution is decreased.

Besides, the weight sharing technique has the interesting side effect of reducing the number of free parameters, thereby reducing the "capacity" of the machine and improving its generalization ability.

A very successful CNN architecture used in MNIST problem has been introduced by Y. LeCun et al. 1990, which gives one of the best performances so far (0.4% error rate) if combined with elastic deformations, and early-stopping (1.0% error rate if using CNN alone) (Figure 5.6). However, CNN architecture is usually designed in a heuristic way. Recently, many new methods have been introduced to effectively train deep neural networks, such as using generative pre-training, Hessian-free optimization, or even a standard gradient descent with well-designed initialization and momentum. Therefore, we now can train a fully-connected deep neural networks with millions of parameters to extract useful features from the training images automatically. A DNN has been used for MNIST problem with raw input and provided 1.25% error rates [16].

Recently, E. Hinton, Alex and Ilya Sutskever 2012 [23] have provided an amazing result on using Deep CNN on classifying image. They trained a large, deep CNN to classify the 1.2 million high-resolution images in the ImageNet LSVRC-2010 contest into the 1000 different classes. On the test data, they achieved top-1 and top-5 error rates of 37.5% and 17.0% which is considerably better than the previous state-of-the-art. They also used "dropout" method (described below) to reduce overfitting and that was proved to be very effective.

## 5.2.7 Dropout

Hinton et al. 2012 have proposed a very effective regularization method called Dropout. It can reduce overfitting by preventing complex co-adaptions (a feature detector is only helpful in the context of several other specific feature detectors) on the training data [19]. On each presentation of each training case, each hidden unit is randomly omitted from the network with a certain probability, say 0.5, so a hidden unit cannot rely on other hidden units being present. At test time, they use the "mean network" that contains all of the hidden units but with their outgoing weights halved to compensate for the fact that twice as many of them are active.

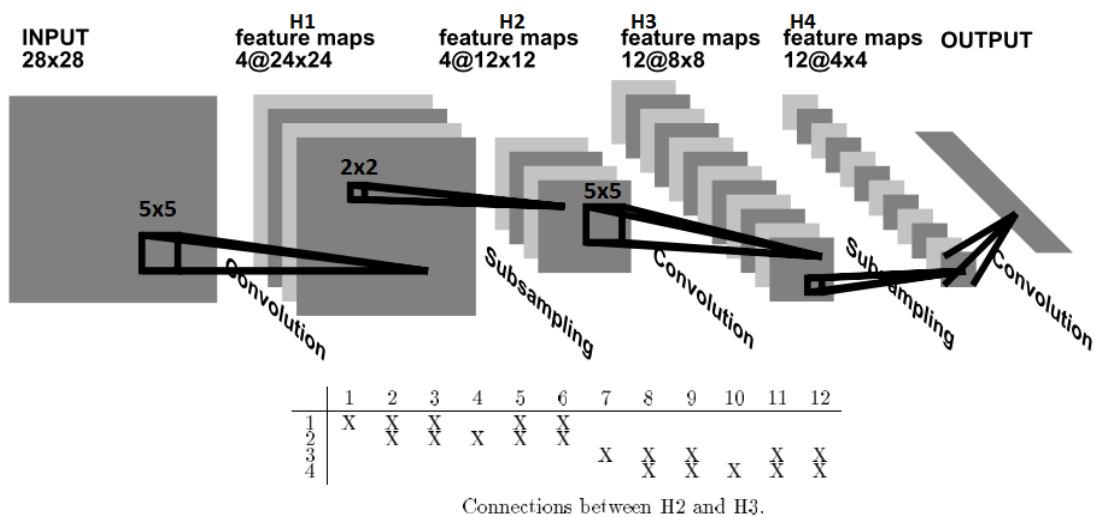

Figure 5.6: CNN architecture used in MNIST problem [24]

Another way to view the dropout procedure is as a very efficient way of performing model averaging with neural networks. We could think of it as training different network for each presentation of each training case but all of these networks share the same weights for the hidden units that are present.

Using together with weight constraint, Hinton et al. 2012 have tested this method on many different dataset. On MNIST dataset, the best published result for a standard feed forward neural network (without adding prior knowledge, preprocessing or generative pre-training) is 1.6%, which could be reduced to 1.3% using 50% dropout and weight constraints, and to 1.1% by also dropping out a random 20% of the pixels. On TIMIT, a widely used benchmark for recognition of clean speech with a small vocabulary, using 50% dropout with deep, pre-trained, feedforward neural networks reduces the recognition error rate from 22.7% to 19.7%. That is a record for methods that do not use any information about speaker identity. Lately, E. Hinton, Alex and Ilya Sutskever 2012 [23] have used a deep convolutional neural networks with dropout method to produce the best result on Imagenet dataset (classify high-resolution images into 1000 different classes).

### 5.2.8   Dropconnect

Lately, a generation of Dropout called Dropconnect has been introduced by Li Wan et al., 2013 [50]. This method sets a randomly selected subset of weights within the network to zero. Each unit thus receives input from a random subset of units in previous layer, instead of dropping out random units as in dropout method. In MNIST dataset, this method gives a slightly better result in some cases. However, it converges more slowly than Dropout. The same thing happens to other dataset such as CIFAR-10, SVHN and NORB. In general, the Dropconnect method can slightly outperform the Dropout method, but it needs more time to converge.

# Chapter 6

# Activation Functions

What makes the deep neural networks become very powerful and universal model is the activation function. A deep neural network contain multiple layers of linear transformation can be represented by a simple one-layer neural network. The nonlinear activation function is what gives neural networks their nonlinear capabilities [25]. The activation function is generally chosen to be monotonic. There are many different types of activation functions that have been proposed. Sigmoid is one of the most common form, which is a monotonically increasing function that asymptotes at some finite value as $\pm\infty$ is approached. In this chapter, we will present the motivation behind the sigmoid function, some of its variants, and introduce some new activation forms coming from recent researches.

## 6.1 Logistic Sigmoid function

The logistic sigmoid function is given by $f(x) = \frac{1}{1+\exp(-x)}$ (Figure 6.1). One important motivation for this form of function is the output of the logistic sigmoid function can be interpreted as posterior probabilities [27]. For example, we consider building a discriminant function for a two-class problem using logistic regression, which has form:

$$y = f(W^T x + b) \tag{6.1}$$

we want to predict the two-class label y from the input x, given that the class-conditional densities are given by Gaussian distributions with equal covariance matrices $\Sigma_1 = \Sigma_2 = \Sigma$, so that:

$$P(x|C_k) = \frac{1}{(2\pi)^{d/2}|\Sigma|^{1/2}} \exp\left\{-\frac{1}{2}(x - \mu_k)^T \Sigma^{-1}(x - \mu_k)\right\} \tag{6.2}$$

Using Bayes' theorem, the posterior probability of membership of class $C_1$ is given by:

$$P(C_1|x) = \frac{p(x|C_1)P(C_1)}{p(x|C_1)P(C_1) + p(x|C_2)P(C_2)} \tag{6.3}$$

Let

$$a = \ln \frac{p(x|C_1)P(C_1)}{p(x|C_2)P(C_2)} \tag{6.4}$$

we will see that the posterior probability can be expressed as the logistic sigmoid function of $a$:

$$P(C_1|x) = \frac{1}{1 + \exp(-a)} \tag{6.5}$$

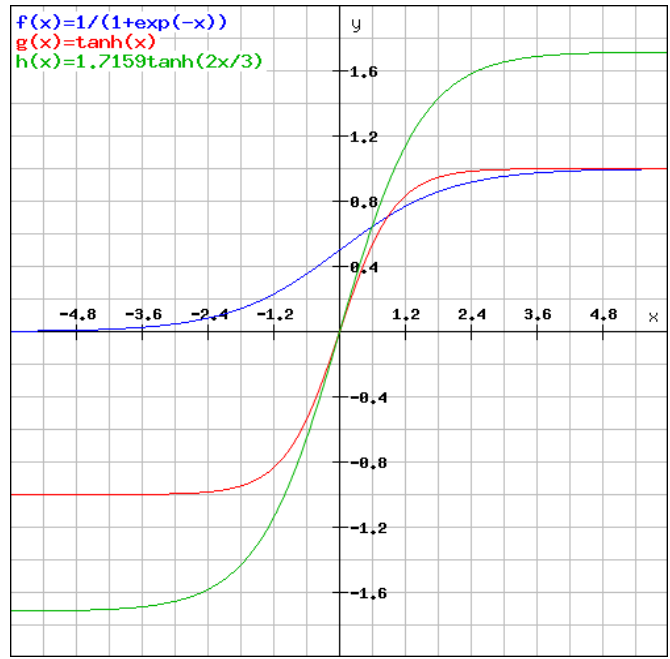

Figure 6.1: The sigmoid, tanh, and scaled tanh functions

If now we substitute the expressions for the class-conditional densities from 6.2 into 6.4, we obtain:

$$a = W^T x + b \tag{6.6}$$

where

$$W = \Sigma^{-1}(\mu_1 - \mu_2)$$

$$b = -\frac{1}{2}\mu_1^T \Sigma^{-1} \mu_1 + \frac{1}{2}\mu_2^T \Sigma^{-1} \mu_2 + \ln \frac{P(C_1)}{P(C_2)}$$

Therefore, we can see that the logistic sigmoid activation function allows the outputs of the discriminant function in 6.1 to be interpreted as posterior probabilities.

## 6.2   Hyperbolic tangent and its scaled version

The Hyperbolic tangent or tanh function is a rescaling of the logistic sigmoid, such that its outputs range from $-1$ to $1$ instead of $0$ to $1$ as in logistic sigmoid. Let $s(x)$ is the logistic sigmoid function, we can represent the $\tanh(x)$ function as a linear transformed version of $s(x)$:

$$\tanh(x) = \frac{e^x - e^{-x}}{e^x + e^{-x}} = 2s(2x) - 1 \tag{6.7}$$

In neural network, the tanh function is more popular because of its symmetry about the origin (i.e. zero-mean). In other words, the tanh are more likely to produce outputs (which are inputs to the next layer) that are on average close to zero. Moreover, the logistic function has been shown to slow down the learning process because of its none-zero mean that induces important singular values in the Hessian [25].

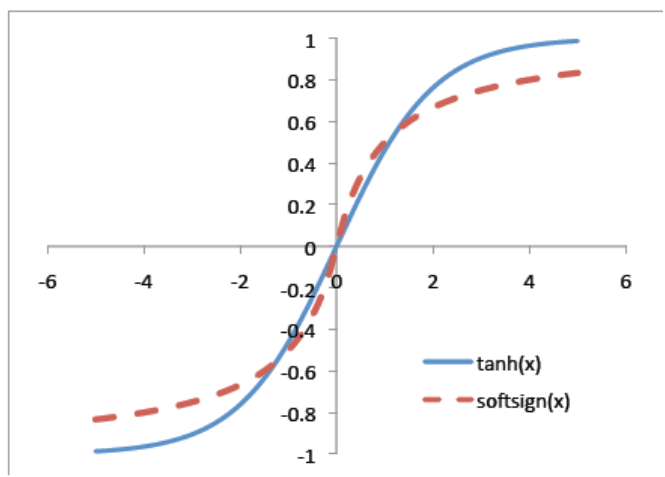

Figure 6.2: *Taken from [4], tanh* versus the *softsign*, which converges polynomially instead of exponentially towards its asymptotes

Glorot and Bengio 2010 [9] did a deep investigation on the effect of using activation functions on deep neural network. They showed that when using sigmoid activation on 5-layers network, the last hidden layer quickly saturates at 0 (slowing down all learning), but then slowly desaturates around epoch 100. The hyperbolic tangent networks do not suffer from that kind of saturation behavior. However, with random weight initialization, the saturation phenomenon occur sequentially starting with first layer and propagating up in the network.

Lecun et al. 1998 also recommended a scaled version of tanh: $f(x) = 1.7159 \tanh(\frac{2}{3}x)$. The constants in this function are chosen so that, when used with normalized inputs (e.g. zero mean, uncorrelated, and unit covariance), the variance of the outputs will also be close to 1. In particular, this function has the properties that (a) $f(\pm 1) = \pm 1$, (b) the second derivative is a maximum at $x = 1$, and (c) the effective gain is close to 1.

## 6.3 Softsign function

Bergstra et al. 2009 [4] has proposed a new activation function called softsign: $f(x) = \frac{x}{1+|x|}$. The softsign is similar to the hyperbolic tangent (its range is $-1$ to 1) but its tails are quadratic polynomials rather than exponentials, i.e., it approaches its asymptotes much slower [9]. In Glorot and Bengio 2010 experiment, they showed that saturation does not occur one layer of the other in deep softsign networks like for the hyperbolic tangent networks. It is faster at the beginning and then slow, and all layers move together towards larger weights [9]. Without pre-training, the softsign networks perform better than the tanh or logistic sigmoid networks on different datasets such as MNIST, Shapeset, CIFAR10, ...

## 6.4 Rectifier and Softplus function

Many differences exist between neural network models used by machine learning researchers and those used by computational neuroscientists. Glorot et al. 2011 [10] wanted to bridge (in part) a machine learning / neuroscience gap in terms of activation function and sparsity. They have successfully applied the rectifier function suggested in neuroscience into machine learning neural network models, which could perform better than tanh or sigmoid networks in some particular settings.

There are two main neuroscience observations that inspired their works:

- Studies on brain energy expense suggest that neurons encode information in a sparse and distributed way (Attwell and Laughlin, 2001), estimating the percentage of neurons active at the same time to be between 1 and 4%. However, without additional regularization, such as an $L_1$ penalty, ordinary feed forward neural nets do not have this property.

- A common biological model of neuron, the leaky integrate-and-fire (or LIF) (Dayan and Abott, 2001) is very different from the logistic sigmoid or tanh function used in machine learning.

Moreover, they were also inspired particularly by the sparse representations learned by sparse auto-encoder. However, they argued that when using the sparsity penalty to induce the sparse representations, the neurons end up taking small but non-zero activation. They want to build truly sparse representations, which gives rise to real zeros of activations.

Combining all of these ideas, they ended up using the rectifier neurons, which introduced in the neuroscience literature by Bush and Senowski 1995: $f(x) = \max(0, x)$. This activation function shows the following advantages:

- The rectifier activation function allows a network to obtain sparse representations easily. For example, after uniform initialization of the weights, around 50% of hidden units continuous output values are real zeros, and this fraction can be easily increase with sparsity-inducing regularization.

- The only non-linearlity in the network comes from the path selection associated with individual neurons being active or not (illustrated in Figure 6.3). For a given input, only a subset of neurons is active. Once this subset of neurons is selected, the output is a linear function of the input. We can see the model as an exponential number of linear models that share parameters. Because of this linearity, gradients flow well on active paths of neurons (there is no gradient vanishing effect due to activation non-linearities of sigmoid or tanh units).

However, there is one potential problem, that the hard saturation at 0 may hurt optimization by blocking gradient back-propagation. The authors have evaluated this by investigating the soft-plus activation: $f(x) = \log(1 + e^x)$, as smooth version of the rectifier. However, experimental results suggested that hard zeros could actually help supervised training. Another problem could arise due to the unbounded behavior of the activations. Therefore, the authors used the $L_1$ weight decay on the activation values, which also promotes additional sparsity.

Experimental results in [10] showed that rectifier performs better than other traditional activation functions on supervised training (1.43% on MNIST compared to 1.57% of tanh network). However, with unsupervised pre-training, the difference is not significant. Glorot et al. 2011 [10] also presented a short summary of sparse representation advantages, which is worth mentioning here.

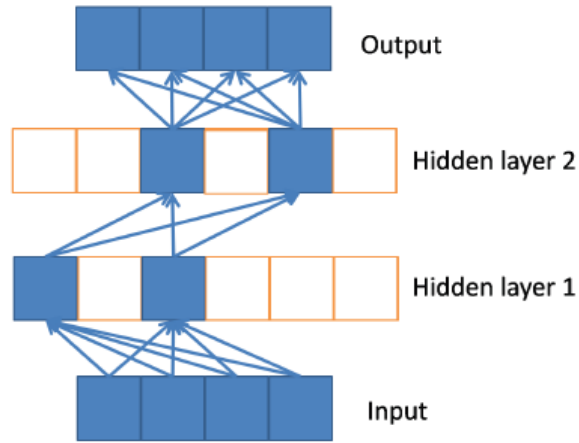

Figure 6.3: *Taken from [10]*, Sparse propagation of activations and gradients in network of rectifier units. The input selects a subset of active neurons and computation is linear in this subset

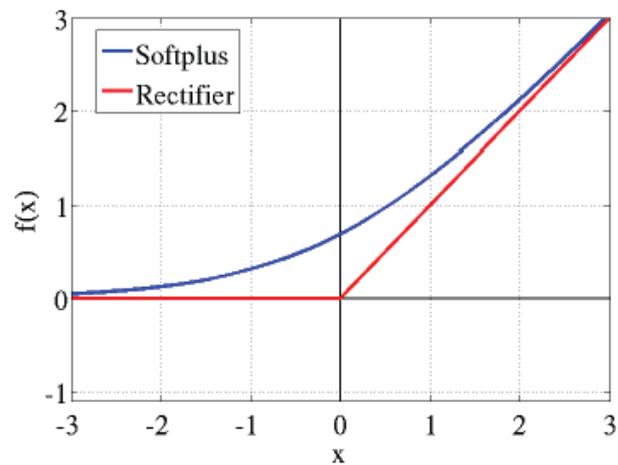

Figure 6.4: *Taken from [10]*, Rectifier and softplus activation functions. The second one is a smooth version of the first

- *Information disentangling*: A dense representation is highly entangled because almost any change in the input modifies most of the entries in the representation vector. Instead, if a representation is both sparse and robust to small input changes, the set of non-zero features is almost always roughly conserved by small changes of the input.

- *Efficient variable-size representation*: Different inputs may contain different amounts of information and would be more conveniently represented using a variable-size data-structure, which is common in computer representations of information. Varying the number of active neurons allows a model to control the effective dimensionality of the representation for a given input and the required precision.

- *Linear separability*: Sparse representations are more likely to be linearly separable simply because the information is represented in a high-dimensional space.

## 6.5   Maxout Function

Goodfellow et al. [11] proposed maxout activation function, which yields state of the art performance on MNIST, CIFAR10 and CIFAR100 datasets when applied on convolutional neural networks with dropout (0.45%, 11.58%, and 38.57% respectively). This is a very simple model, which designed to both facilitates optimization by dropout and improves the accuracy of dropout's fast approximate model averaging technique.

As introduced in chapter 5 about regularization techniques, Dropout (Hinton et al., 2012) provides an inexpensive and simple means of both training a large ensemble of models that share parameters and approximately averaging together these models' predictions. While dropout is known to work well in practice, it has not previously been demonstrated to actually perform model averaging for deep architectures [11]. Therefore, the authors tried to design a model that enhances dropout's abilities as a model averaging technique.

The maxout network is simply a feed-forward neural network or deep convolutional neural network, which uses a new type of activation function: the *maxout* unit. Given an input $x \in R^d$ ($x$ may be a training input or a hidden layer's state), a maxout hidden layer implements the function:

$$h_i(x) = \max_{j \in [1,k]} z_{ij} \tag{6.8}$$

where $z_{ij} = x^T W_{...ij} + b_{ij}$ and $W \in R^{d \times m \times k}$

Note that in maxout network, each layer contains $k$ different weight matrix instead of only one weight matrix in feed forward network. This idea fits well in convolutional network where we have multiple feature maps on each layer. In a convolutional network, a maxout feature map can be constructed by taking the maximum across $k$ affine feature maps (i.e., pool across channels, in addition spatial locations). When training with dropout, we perform the element wise multiplication with the dropout mask immediately prior to the multiplication by the weights in all cases - we do not drop inputs to the max operator. A single maxout unit can be interpreted as making a piecewise linear approximation to an arbitrary convex function. Therefore, maxout networks learn not just the relationship between hidden units, but also the activation function of each hidden unit. The maxout abandons many of the mainstays of traditional activation function design. The representation it produces is not sparse at all. Moreover, maxout is locally linear almost everywhere, while many popular activation functions have significant curvature. However, it is very robust, easy to train with dropout, and achieves excellent performance.

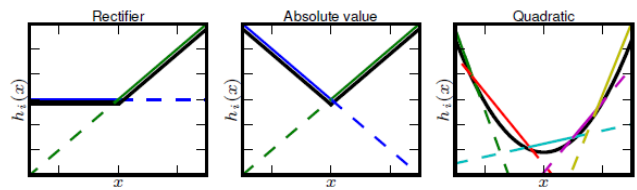

Figure 6.5: *Taken from [11]*, Graphical depiction of how the maxout activation function can implement the rectified linear, absolute value rectifier, and approximate the quadratic activation function. This diagram is 2D and only shows how maxout behaves with a 1D input, but in multiple dimensions a maxout unit can approximate arbitrary convex functions.

# Chapter 7

# Introduction to ADATE

In this chapter, we only introduced the ADATE system very briefly, which could help you understand the main ideas behind it. For more details about how its internal algorithms work, please refer to [33, 34]. Besides, writing specification file for ADATE is also an art, which needs both knowledge and experience. The paper [32] could be a very useful source for anyone wanting to write a good specification. A complete User Manual is also available, which provides many practical information about how to use ADATE [47].

## 7.1 A Short Introduction to ADATE

ADATE [35] has been developed by Prof. Roland Olsson, to automatically generate purely functional programs. For example, it has been employed to improve state-of-the-art SAT solvers [36].

To evolve a solution to a problem, the system needs a *specification file* that defines data types and auxiliary functions, a number of training and validation input examples, and an *evaluation function* that is used to grade and select potential solutions during evolution. Additionally, the specification file may contain an initial program from which evolution will start. Of course, it is possible to start the evolution from any given program, for example to search for improvements for the best known program for a given problem.

The programs are constructed using a limited number of so-called *atomic program transformation*. The most important ones are as follows.

- **R** (**R**eplacement) - A part of an existing program is replaced by a newly synthesized expression. Due to the extremely high number of expressions that can be synthesized, **R** transformations are combinatorially expensive.

- **REQ** (**R**eplacement preserving **Eq**uality) - An **R** transformation that does not make the program worse according to the given evaluation function. REQ transformations are quite useful due to their ability to explore plateaus in the search landscape.

- **ABSTR** (**Abstr**action) - Like **REQ** transformations, these neutral transformations exist to aid the system in exploring plateaus. In contrast to the general **REQ** transformation, **ABSTR** transformations have the very specific task of introducing new functions in the program by factoring out a piece of code and replacing it with a function call. This gives the system the important ability of inventing needed help functions on the fly, something which has proven to be an extremely useful feature.

Atomic program transformations are composed to *compound program transformations* using a number of different heuristics to avoid common cases of infinite recursion, unnecessary transformations etc. For example, after an **ABSTR** transformation, the newly introduced function should be used in some way by a following **R** or a **REQ** transformation. More details on the atomic program transformations and the heuristics employed to combine them can be found in [35].

Each time a new program is created by a compound transformation, it is considered for insertion into the so-called kingdom [49]. As in all evolutionary systems, individuals with good evaluation values are preferred, but in ADATE, the syntactic complexities of the individuals also play an important role. According to what is commonly known as Occam's Razor, simple theories should usually be preferred over more complex ones, as the simpler theories tend to be more general. This principle is utilized by ADATE to reduce the amount of overfitting, in that small programs are preferred, and if a large program is to be allowed in the kingdom, it has to be better than all programs smaller than it [35]. In other words, a new program will only be allowed to be inserted into the kingdom if all other programs in the kingdom are either larger or worse than it. Each time a program is added to the kingdom, all programs in the kingdom both larger and worse than than the new one are removed.

Having described the most important components of the ADATE system, we conclude this section by giving a brief overview of the overall evolution occuring in a run of the system.

1. Initiate the kingdom with the single program given as the start program in the specification file (either an empty program or some program that is to be optimized). In addition to the actual programs, the system also maintains an integer value $C_P$ for each program $P$, called the *cost limit* of the program. For new programs this value is set to the initial value 1000.

2. Select the program $P$ with the lowest $C_P$ value from the kingdom.

3. Apply $C_P$ compound program transformations to the selected program, yielding $C_P$ new programs.

4. Try to insert each of the created programs into the kingdom in accordance with the size-evaluation ordering described above.

5. Double the value of $C_P$, and repeat from step 2.

The above loop is repeated until the user terminates the system. The ADATE system has no built-in termination criteria and it is up to the user to monitor the evolving programs and halt the system whenever he considers the evolved results good enough.

## 7.2   A Short Analysis of the Power of ADATE System

ADATE is a general-purpose automatic programming system, which can be used in different ways. Besides being well known for its ability of meta-learning or "learn how to learn", ADATE can also be applied as a traditional machine learning model on classification or regression problems. Moreover, it has showed its superior fitting power in many cases, compared to other traditional approaches.

In this section, I will do a short analysis of the power of ADATE system in different use cases, based mainly on the structure of its synthesized programs. In other words, I will basically do a black box analysis where I ignore the ADATE's internal algorithm and only focus on what kind of programs it would probably synthesize in different use cases. For more information about how it does the "magic", please refer to [33, 34].

### 7.2.1 ADATE on Classification Problems

Classification is the problem of predicting the value of a categorical variable based on other explanatory variables. An algorithm that implements classification is called a classifier. In this section, I will further split this problem in two different types: when explanatory variables are continuous or categorical. Because the ADATE tends to synthesize programs in different structure when applied to different type.

**Continuous Explanatory Variables**

An example for this type is an edge detection problem, which is done by Kristin Larsen[1] as her master thesis. In this problem, the classifier has to predict if the middle pixel is on an edge of a 5x5 pixels image. The explanatory variables are intensity of 24 pixels surrounding the middle one (T1 to T24). The intensity value range is an integer number in [0..255]. After trained for a prolonged time (more than two weeks), the best ADATE's synthesized program (tested on separate validation set) is showed below [2]:

```
fun f (
    T1, T2, T3, T4, T5,
    T6, T7, T8, T9,T10,
    T11, T12, T13, T14, T15,
    T16, T17, T18, T19, T20,
    T21, T22, T23, T24 ) =
let
    fun g x =
        realLess( x, 11 )
in
    case
        case g( T12 ) of
            false =>
                g(
                    case g( T8 ) of
                        false =>
                            case g( T18 ) of
                                false => T7
                              | true =>
                                    realSubtract(
                                        T18,
                                        case g( T17 ) of false => T13
                                                       | true => T4
                                    )
                      | true =>
                            case g( T1 ) of
                                false =>
                                    case g( T10 ) of
                                        false => T8
```

---

[1]Kristin is also a master student at Hiof. Her thesis is not finished yet.
[2]This code has been modified from the original one to make it clean

```
                          | true => realSubtract( T9, T12 )
31              | true => T19
              )
33      | true => true        of
        false => g( T16 )
35    | true => true
    end
```

Code 7.1: Clean version of the best synthesized program for edge detection problem

We can easily recognize that the above program has a similar structure to a decision tree. However, as experimented by Kristin Larsen, the random forest - an ensemble learning method for classification by constructing a multitude of decision trees - was outperformed by that program. Figure 7.1 illustrates the equivalent decision tree of that program. How could the code 7.1 outperform the decision tree, random forest, or even deep neural network classifiers as shown in the Kristin's experiment? I hypothesize that that is come from these following abilities of ADATE's synthesized classifier:

- *Feature extraction*: The ADATE can discover feature extractors when they are needed. In figure 7.1, you can see that ADATE has discovered three new features: $T9 - T12$, $T18 - T4$ and $T18 - T13$ as highlighted in green boxes. In this case, ADATE only learned linear feature extractors (i.e. linear filters). However, it is capable of learning much more complex feature extractors. In the next example about four-way item-to-item navigation algorithm, you will see this more clearly. This ability is extremely useful when the explanatory variables are highly correlated. There is a well-known problem (may be considered as a paradox) in statistics and machine learning literature that using all available explanatory variables to build a predictive model can hurt its performance very bad. This problem occurs when some of these explanatory variables are not independent and highly correlated. One simple and famous example of this is the collinearity problem in multivariate linear regression model, when two or more explanatory variables have linear relationship.

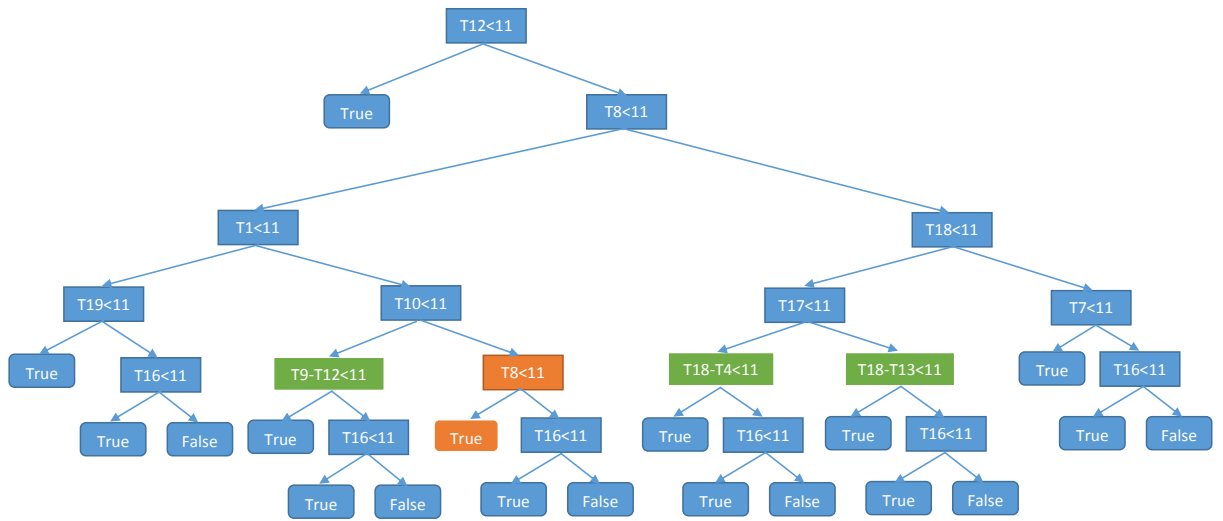

Figure 7.1: The equivalent decision tree of the ADATE solution for edge detection problem; Left branches are the True cases

We can overcome (or minimize) this correlation problem in many ways, such as eliminating redundant variables manually or automatically, normalizing the input, or doing feature extraction to extract useful feature from high-dimensional and highly correlated inputs. The last solution has been shown to be the best one, especially in computer vision where we have very high-dimensional inputs (images) with highly correlated components (pixels). This explains why deep learning and feature extraction have dominated in many computer vision tasks such as digits recognitions or object classification. Please refer to chapter 3 and chapter chap:autoencoder about deep architectures and autoencoders for more information about this.

- *Divide and Conquer*: The ADATE can split the input space into many sub-spaces, and build different feature extractors in different sub-spaces. This is a very powerful ability, which may explain why ADATE can easily synthesize classifier that fit very well to many different problems. However, this also makes ADATE's synthesized programs easy to be overfitted. To demonstrate this power, take example of an classification problem where the posterior distribution of the target variable $y$ conditioned on the input $x$: $P(y|x)$ is different in different sub-spaces of input, we will need to fit different model to each of these sub-spaces to make the best classifier. Even if the $P(y|x)$ is distributed consistently in the input domain, but if it is a very complex function, dividing the input domain into many sub-spaces and trying to approximate that complex function in each sub-spaces using much more simple functions is also a very good strategy. This is a similar strategy employed by the Maxout network where it uses piecewise linear functions to approximate the activation function for each neural node[3]. Moreover, in classification problems where the relationships between explanatory variables are different in different input's sub-spaces, we also need to have different feature extractors in different input domains.

  Generally, the ADATE actually builds an ensemble of many different models in different input's sub-spaces. This divide and conquer strategy not only brings the fitting power to the ADATE system, but also introduces overfitting problem. Splitting the input space too much will increase the chance of discovering accidental relationships (between explanatory variables and target variable as well as among explanatory variables) of ADATE system, especially when we have low-granularity training data (i.e. not enough data). Of course, the Occam's razor employed in ADATE searching strategy, and a good validation set could be very helpful in this case. Nevertheless, empirical results showed that in general, using ADATE to build a classifier for a high dimensionality continuous input domain is more likely to create an overfitted solution, compared to other use cases of ADATE. You will see in the next section that, on regression problem, ADATE usually does not divide the input space in its solutions, which make it harder to be overfitted.

- *Good Searching Strategy*: Evolution strategy used in ADATE, as far as I know, is one of the best optimization technique in machine learning, which does not suffer from the local optima and other similar problems in greedy optimization methods such as ID3 in decision tree or gradient-based methods in neural networks. Indeed, ID3 algorithm is a very greedy approach where it choose the best node at each time, while the gradient-based optimization methods can only discover the optima that are close enough to the initial point, i.e., it can only search for a local parameter space. The evolution strategy, on the other hand, can covers much larger parameter space. However, this optimization technique runs much slower than the others, as a trade-off for its performance.

---

[3]Please refer to chapter about activation function

- *Compact Representation*: ADATE can represent a complex decision tree in a very compact way by using helper functions or nested case instructions, which could allow it to learn a very complex tree represented in only a few lines of code. This makes ADATE powerful but easy to overfit, even if the Occam's razor employed as a regularization in ADATE. A short program is preferred in ADATE searching strategy, but even a short program could represent a very complex model.

Another good example for the feature extraction ability of ADATE system is its solution for the four-way item-to-item navigation problem, where we have to build a model to predict which item that the user want to select, providing current selected item, list of available items and the navigation direction [21]. Figure 7.2 illustrates the problem clearly, while the best ADATE's solution is shown in Code 7.2.

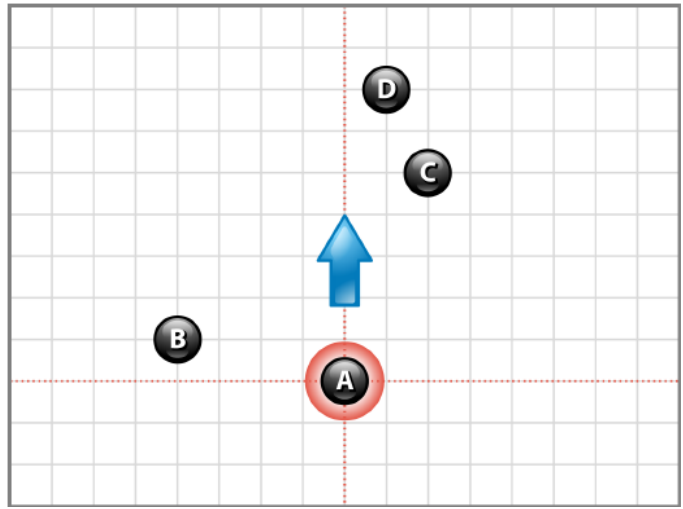

Figure 7.2: *Taken from [21]*, mobile navigation example: Item A is selected, and the user has pressed the up-button. The model has to predict which item the user intends to select

```
function f(
  list_of_points,
  direction_of_movement,
  current_pos
) : point
begin
  best_p = bad_initial_value
  for each point p in list_of_points do
    vector v = line from current_pos to p
    if angle between v and
      direction_of_movement <
      45 degrees
```

```
14        and
          distance( current_pos , p ) <
          distance( current_pos , best_p )
16      then
          best_p = p
18      return best_p
    end
```

Code 7.2: *Taken from [21]* Pseudo code for the best ADATE's solution

You can see that the ADATE has discovered two new features for each point *p*: the angle between the vector *v* (vector from current selected item to the point *p*) and the movement direction; and the difference between the distances between selected point and the current point and the best point so far (the function for calculating the Euclidean distance is given as primitive function). If these features are used when building a decision tree (instead of the point's coordinates) to decide whether we should select the current point as the best point so far, or ignore it, it probably becomes a very easy learning problem. However, it is very hard to build new useful features for decision-making manually. Therefore, ADATE's ability of discovering new useful features is extremely valuable.

**Categorical Explanatory Variables**

In this type of problem, all of the explanatory variables are categorical. A simple toy example for this is the CAR problem, where you have to predict the risk of being stolen of a car, based on its following attributes: the car model, is it has an security alarm?, is it parked in urban or rural area?, is it parked on street or in garage?. The ADATE solution for this type of problem will look exactly like a decision tree. One of its solution is shown in Code 7.3. However, thanks to its better optimization strategy, the ADATE will probably come up with a better decision tree, compared to a decision tree learned by a traditional approach like ID3.

```
1 fun f( X0model , X1alarm , X2area , X3parking ) =
    case X0model of
3     modelopel => (
        case X3parking of
5         parkingstreet => rischigh
          | parkinggarage => risclow
7         )
      | modelvolvo => risclow
```

Code 7.3: One ADATE's solution for CAR problem

## 7.2.2 ADATE on Regression Problems

Regression is the problem of predicting value of a continuous variable based on other explanatory variables. In this problem, ADATE usually synthesizes solutions that can be generalized better than in classification problem. To understand this phenomenon, I have investigated the following examples.

The first simple example is the problem of predicting the velocity of a falling ball when it hits the ground, given the height $H$ of its starting point. The correct answer for this problem is $\sqrt{19.6H}$. A typical ADATE's solution for this problem is shown in Code 7.4.

```
1 fun f H =
2   realAdd(
```

```
      realMultiply (
 4        H,
          tanh ( tanh ( sqrt ( tanh ( tanh ( tanh ( tanh ( tanh ( H ) ) ) ) ) ) ) )
 6        ),
      realAdd (
 8        tanh ( H ),
          sqrt (3)
10        )
        )
```

<div align="center">Code 7.4: Typical ADATE's solution for SPEED problem</div>

As you can see, the ADATE tried to approximate the correct function by a relatively complex program with some redundant operations. However, when comparing this program to a dense neural networks with only 10 hidden nodes, the solution in Code 7.4 is much simpler. A dense neural networks with only 1 input, 10 hidden nodes, and 1 output when represented in a "program" form needs 42 operations[4], while the ADATE's solution above contains only 13 operations (some operations are even redundant and can be discarded without affecting the function much)[5]. Therefore, a neural network has much more freedom in its parameter's space, which could explain why ADATE's solution for this SPEED problem can be generalized much better than a neural network. I hypothesize that the Code 7.4 can be represented by a very deep and sparse neural network, which is the current trends, and the best way in designing neural network nowadays. The Code 7.4 contains *realMultiply* and *sqrt* function, which cannot be directly translated into a neural network. However, I believe that if we only make linear operations (i.e. realAdd and realSubstract) and an activation function (e.g. tanh or sigmoid function) available as primitive functions, ADATE will generate a solution that can be represented exactly by a very deep and sparse neural network. Besides, multiplication, division, sqrt, or many other arithmetic operations can be efficiently represented by a small one-layer neural network, which is proven by [37].

Let check my hypothesis on a different problem, which contains more explanatory variables. There is a wine quality problem where we have to predict the quality of wine based on its eleven attributes: Fixedacidity, Volatileacidity, Citricacid, Residualsugar, Chlorides, Freesulfurdioxide, Totalsulfurdioxide, Density, Ph, Sulphates, and Alcohol. All of these explanatory variables are continuous. One typical ADATE's solution for this problem is shown in Code 7.5. You can see that the synthesized model look very like a deep sparse neural network. If assumed that the realMultiply and realDivide operations in that code can be approximated by a small shallow neural network, we can actually approximate that f function by a five-layers and very sparse neural networks.

```
 1  fun  f
          (
 3            Fixedacidity ,
             Volatileacidity ,
 5            Citricacid ,
             Residualsugar ,
 7            Chlorides ,
```

---

[4]$2 * 10$ linear operations and 10 activation functions for the first layer, and $11 * 1$ linear operations + 1 activation function for the output layer

[5]In fact, the formula
$H*\tanh(\tanh(\sqrt{(\tanh(\tanh(\tanh(\tanh(\tanh(H)))))))}))$ can be approximated by two simple linear equations: $y = 0$ if $H < 0$ and $y = 0.57143H$ if $H \geq 0$

```
                    Freesulfurdioxide ,
9                   Totalsulfurdioxide ,
                    Density ,
11                  Ph,
                    Sulphates ,
13                  Alcohol
                    ) =
15      realMultiply (
            Alcohol ,
17          realAdd (
                Volatileacidity ,
19              sigmoid (
                    realAdd (
21                      Totalsulfurdioxide ,
                        realDivide (
23                          realSubtract ( Volatileacidity ,
                                           Residualsugar  ) ,
25                          sigmoid ( Volatileacidity  )
                            )
27                      )
                    )
29              )
            )
```

Code 7.5: Typical ADATE's solution for Wine quality problem

Another very important discovery when I investigated many different ADATE's solutions for regression problems is that the ADATE system usually does not synthesize programs that split the input space into different sub-spaces, as in case of classification problems. Without these divide and conquer ability, the ADATE has lost a very important part of its fitting power. This lost, however, makes the ADATE system much harder to be overfitted. Generally, on regression problems, the ADATE has to build a single model for the whole input space, instead of building different simple model for different parts of the input domain. This may explain why the generality of ADATE's solutions for regression problem is better that its solutions for classification problems.

The evolution strategy used in ADATE can be a possible factor which causes this phenomenon. The evolution strategy searches for the smallest programs first. After that, it generates other bigger programs by applying different kinds of transformation and mutation. In regression problem, we need the function f to return a continuous value. Therefore, the smallest programs which return continuous values will be more likely to be picked for later transforming and mutating. If ADATE tries to add a "case" instruction into an existing program, that case instruction has to condition on a continuous variable, which can not serve as a branching statement. Of course, the ADATE can first apply a function that returns a boolean value like "realLess" function, and introduce the case instruction after that. However, because boolean values can not be used in an equation of real-value variables, the ADATE usually ignores these transformations. In other words, if the ADATE wants to add a branching statement into a current program, it must allow a two-step transformation, where it can add a boolean-return function and a case instruction at the same time. Because if it, almost any case instruction in ADATE's solutions for regression problems serves as an declaration instruction for temporary variables instead of a branching statement. Of course, the ADATE system can take advantages of categorical explanatory variables to generate branching statement if they exist. However, in typical regression problems, all of the explanatory variables are continuous.

### 7.2.3  ADATE on Classical algorithm

ADATE has been proven able to generate the correct algorithms for many different classical problems, such as: inserting and deleting a node in binary tree, generating all permutations of a list, or sorting a list. We would probably ask why it could generate the correct answers, instead of an approximate answer, for this task. I believe that the following reasons may answer the question:

- *Existence*: ADATE can discover the correct solutions for these problems may be just because they exist. In typical classification or regression problems, there is usually a probabilistic model behind the scene, instead of a deterministic model. Therefore, synthesizing an exact algorithm for these tasks is impossible.

- *Suitable Representation*: ADATE represents its answer in a "functional program" form, which is very suitable for expressing algorithms. Besides, these algorithms have correct answers that can be represented in a very short functional program, which is exactly what the ADATE evolution strategy prefers.

- *No Constant*: I hypothesize that ADATE is not very good at generating a constant number in its program. It may be because a constant number that is useful in current program will become completely irrelevant in its transformations. In other words, unlike other operations, a constant number depends very much on the program context, and when we change the program just a little bit, it becomes useless. However, you probably will not see any random constant number in classical algorithms, which may help the ADATE very much.

Code 7.6 illustrates one of the best ADATE's solution for sorting problem, where you have to sort a list of integer number into ascending order.

```
fun f (V1) =
  case V1 of
    nil => V1
  | (V1_1 :: V1') =>
  let
    fun g (V2) =
      case V2 of
        nil => (V1_1 :: nil)
      | (V2_1 :: V2') =>
      case (V1_1 < V2_1) of
        true => (V1_1 :: V2')
      | false =>
      (V2_1 :: g( V2' ))
  in
    g( f( V1' ) )
  end
```

Code 7.6: An ADATE's solution for sorting problem

This code is hard to understand at first sight. But if we look closer, we can see that the variable $V1\_1$ is included in the closure of the $g(.)$ function. Therefore, we can change the function call $g(f(V1'))$ into $g(V1\_1, f(V1'))$. Now, that tail recursion is trying to apply the g function on all components of V1. For example, if $V1 = A1 :: A2 :: A3 :: A4 :: nil$, that tail recursion is equivalent to this function call:

$$g(A1, g(A2, g(A3, g(A4, nil))))$$  (7.1)

Now if we change the name of function $g(x, V)$ into $insert(x, V)$, the equation 7.1 is now equivalent to

$$insert(A1, insert(A2, insert(A3, insert(A4, nil)))) \qquad (7.2)$$

We can see that if the $insert(x, V)$ function returns a sorted list by insert x into V, given that V is a sorted list, then the algorithm is correct. Look at the body of $insert(x, V)$ function in Code 7.6[6], you can see that it compare $x$ to the first element $y$ in $V$. If $x$ smaller than $y$, then it return the list $x :: V$. However, if $x$ is equal or greater than $y$, then $y$ will be the first element in the return list. The rest of return list will be sorted again by calling $insert(x, V')$ with $V'$ is the vector $V$ after removing its first element $y$. This recursion continues until $x$ smaller than $y$ or $V$ becomes nil. Therefore, this ADATE's solution for sorting problems is indeed a correct algorithm.

### 7.2.4   ADATE on Meta-learning

One special ability of ADATE that you cannot find in most of other machine learning tools is that it can do meta-learning or "learn how to learn". It means that we can use ADATE to improve other algorithms or machine learning methods. The evolution strategy that employed in ADATE is very flexible which allow it to work in very different ways.

Most of machine learning optimization methods need a well-defined continuous objective function, which represents a clear relationship between the current model's parameters configuration and its performance. Based on that function, they use some searching strategy to optimize the parameters, such as gradient-based methods. For example, the mean squared errors is a common objective function in regression, while negative log likelihood is used in classification problems. The need of continuous objective function limits the possible problems that these methods can be applied on.

On the other hand, the ADATE system uses *fitness function* in its searching strategy. Although the fitness function can be considered as a special type of objective function, it is very different from other types. Generally, it is not needed to be a continuous function, and the relationship between a model and its fitness measurement can be very vague. Therefore, we usually cannot get any gradient information from a fitness function.

One example that can demonstrate this ADATE flexibility clearly is using ADATE to improve decision tree pruning [13]. In this problem, the authors wanted to improve an error based pruning (EBP) algorithm used in C4.5 decision tree. Starting from the initial $f$ function that is a naive implementation of EBP, ADATE has discovered a better version of it. The original $f$ function is relatively long and complicated function. Therefore, I only choose a small part of it to demonstrate the ADATE flexibility. Code 7.7 shows the original and improved version of errorEstimate function used in EBP algorithm. This function takes two arguments: $n$ – the number of instances that reach a given tree node, and $c$ – the number of these instances that are correctly classified by the subtree corresponding to the node.

```
(* Original errorEstimate(.) function *)
fun errorEstimate ( ( c , n) : real * real ) : real =
let
  val e = (n - c)/n
  val z = 0.69
  val z2 = z * z
  val val1 = (e/n) - (( e*e )/n )) + (z2/(4 . 0*n*n)
```

---
[6]the $g(V2)$ function is the $insert(x, V)$ function with $x$ is $V1\_1$ and $V$ is $V2$

```
 8    val val2 = (e + z2 / (2 . 0 * n)) + z*(sqrt val1)
      val val3 = 1.0 + z2/n
10   in
      val2/val3
12   end
     (* Improved errorEstimate(.) function *)
14   fun errorEstimate ( ( c , n ) : real * real ) : real =
     let
16     val v1 = tanh( tanh( tanh( (n − c) / n)))
       val v2 = sqrt( tanh( tanh( sqrt( n))))
18     val v3 = tanh( sqrt( sqrt( c)))
     in
20    (v1 + v2) / v3
     end
```

Code 7.7: The original and synthesized errorEstimate function used in EBP algorithm

This can be considered as a regression problem, where we want to predict (i.e. estimate) the error value, based on the two explanatory variables $c$ and $n$. The ADATE's solution for this problem can also be represented by a neural network. However, we cannot directly use neural networks or other machine learning methods on this problem, because there is no clear relationship between the output of the function $errorEstimate(.)$ and its performance. In ADATE, that $errorEstimate(.)$ function is called in the decision tree code, which then is tested in several different training datasets to check its performance.

In general, the ADATE can be, and has been, successfully used to improve other machine learning methods, especially when they include heuristic functions. However, the current version of ADATE has some following limitations, which can possibly be fixed in future:

- *Slow Searching Strategy*: Evolution strategy is a very good but slow optimization method. Therefore, we usually need to design small artificial datasets to train the ADATE first, before generalizing its solution to the real-world problems. However, in some cases, its solutions for the small datasets cannot be generalized well to other bigger datasets. This maybe comes from a bad design of artificial datasets, or bad ADATE's configuration.

- *Synthesized program can't be called*: In current ADATE version, outside of the ADATE-ML part, the synthesized program cannot be called. This means that you have to implement the whole original algorithm in ADATE-ML if you want to call the synthesized program in your algorithm. This work is not trivial, because ADATE-ML is a simplified version of Standard-ML, which is a very small and limited language. However, this limitation is going to be fixed soon. In the next ADATE version, the synthesized program may even be called from $C$ or other external programs, which will make ADATE much easier to use and experiment. This change can also speed up the ADATE system, because programs written in C usually run much faster than their ML versions.

- *Hard to understand*: One reason that makes ADATE's improved version of other machine learning methods unpopular is that we usually cannot understand completely the synthesized program. It means that we cannot prove the correctness of that program in a mathematical way, which is usually desired by other researchers.

# Chapter 8

# ADATE Experiments

This chapter presents our experiments as a process, which can answer all of the research questions posed in Chapter 1. Based on deep learning knowledge that we summarized in previous chapters, we analyzed why we chose the initialization part of deep learning algorithm to experiment first. After that, we presented how we built tiny datasets and neural networks library for ADATE. After running the ADATE system to evolve the initial sparse initialization scheme, we got the sparse-3 program, which will be analyzed carefully in this chapter. A short analysis of overfitting problem of synthesized programs is also presented at the end of the chapter.

## 8.1 Selecting Target

The previous chapters about deep learning present many deep learning methods, which could possibly be improved by ADATE. The most valuable one if we could improve is the unsupervised pre-training method. However, this is a complex and time-consuming process, which is hard to work on for the first experiment. We wanted to select a simple but effective method, which does not have dependency relationship with other parts of the deep learning process. The most potential targets are:

- *Autoencoder models:* many different types of autoencoders have been introduced in chapter 3. By adding an extra term into the objective function, we can get a much better version of autoencoders. We can use ADATE to discover a new better term. However, we need to take the gradient of the new objective function automatically, which is relatively hard and error-prone.

- *Optimization methods:* chaper 4 shows that the neural networks performance can be improved significantly by changing the initialization scheme or the momentum schedule without using pre-training methods. While improving momentum schedule is currently not possible with current version of ADATE, the initialization scheme is a seperate part of the training process and can easily be improved by ADATE.

- *Regularization methods:* chapter 5 presents dropout and dropconnect, the two very new but effective regularization methods. However, we have to call the regularization method for each iteration of the training process, which is not possible in current ADATE version.

- *Activation function:* chapter 6 shows that by changing the activation function, we can improve the network's performance. However, because we have to take the gradient of the new activation function automatically, this is not an easy target.

Therefore, the initialization scheme was chosen for our first experiment with ADATE. Other parts of the deep learning process are also potential, but we have to wait until the next version of ADATE, when we can call the synthesized program from outside of the ADATE part.

### 8.1.1   Checking the Effectiveness of Initialization Schemes

Before starting to improve the initialization scheme, we had to check if changing it makes a significant difference in the network's performance. In this experiment, we trained a deep neural network with 3 hidden layers, their corresponding sizes are: 500, 500 and 2000, using steepest gradient descent with momentum suggested in [46]. L2 weight decay was also used and fixed at $10^{-5}$. The batch size is chosen at 200 as in [46]. Basically, all hyper-parameters in our experiment are fixed, except for the learning rate and initialization method as presented in table 8.1.

Table 8.1: Parameters used when training on MNIST dataset

| Parameters | Configuration |
|---|---|
| Dataset: | MNIST |
| Network structure: | [784 500 500 2000 10] |
| Cost function: | Negative log likelihood |
| Momentum schedule: | $\mu = min(1 - 2^{-1-log_2(\lfloor t/250 \rfloor + 1)}, \mu_{max})$ |
| Momentum max: | 0.999 |
| L2 weight decay: | $10^{-5}$ |
| Batch size: | 200 |
| Learning rates: | chosen from [0.05 0.01 0.005 0.001 0.0005 0.0001] |
| Initialization methods: | Normal; Normalized; Sparse |

In the first experiment, we trained the network for all possible combinations of learning rate and initialization method and stop the training process after the first 100 epochs. The result is expected to be noisy because of random initialization and 100 epochs are not enough to get the training process to be saturated, especially with small learning rate. However, we can still see the difference in performance of different initialization methods on different learning rate.

Table 8.2: Test error rates after first 100 iterations for different learning rates and initializations

| Initialization | Learning rates | | | | | |
|---|---|---|---|---|---|---|
| | 0.05 | 0.01 | 0.005 | 0.001 | 0.0005 | 0.0001 |
| Normal: | 4.33% | ***2.89%*** | 3.92% | 6.57% | 7.89% | 10.76% |
| Normalized: | 4.44% | 3.17% | 4.46% | 7.28% | 8.27% | 11.21% |
| Sparse: | 3.86% | 2.94% | 3.47% | 4.6% | 5.26% | 7.56% |
| $\Delta_{\text{Normal-Sparse}}$: | $1.45\% \pm 1.34\%$ | | | | | |
| $\Delta_{\text{Normalized-Sparse}}$: | $1.86\% \pm 1.43\%$ | | | | | |

In table 8.2, we can see that sparse initialization consistently outperforms other methods, regardless of what learning rate is using.

In the second experiment, we checked performance of different initialization methods at 0.01 learning rate, which is the best learning rate for all of them in previous experiment. For each initialization method, we trained the network 5 times in attempt to reduce noise produced by random initialization. The result in table 8.3 shows the same thing in the first experiment, but with lower noise. We again see that sparse initialization can easily outperform other initialization methods, at the same or different learning rate.

Table 8.3: Test error rates after first 100 iterations for different initializations at the same learning rates

| Initialization | Learning rates | | | | |
|---|---|---|---|---|---|
| | 0.01 | 0.01 | 0.01 | 0.01 | 0.01 |
| Normal: | 3.08% | 2.96% | 3.05% | 3.17% | 3.12% |
| Normalized: | 3.11% | 3.16% | 3.11% | 3.19% | 3.01% |
| Sparse: | 2.70% | 2.56% | 2.60% | 2.45% | 2.69% |
| $\Delta_{\text{Normal-Sparse}}$: | $0.48\% \pm 0.139\%$ | | | | |
| $\Delta_{\text{Normalized-Sparse}}$: | $0.516\% \pm 0.164\%$ | | | | |

These two above experiments proved that initialization scheme is an essential part of the training process. In addition, if we can improve it by ADATE, we can improve the network's performance significantly.

## 8.2 Building Tiny Datasets

ADATE usually needs to generate and evaluate at least hundreds of thousand or even millions of different programs before possibly discovering the best ones. Therefore, using huge dataset like MNIST directly is definitely impossible. Other smaller datasets such as Curves or USPS could not help also. What we need is a tiny dataset, on which training a DNN costs least than ten seconds. Besides, one crucial property that the tiny dataset has to possess is that an initialization scheme that performs well on it can be generalized and perform as well on other real and bigger datasets.

### 8.2.1 TinyDigits

In our first experiment, after searching for many possibilities, we finally ended up using the TinyDigits dataset. TinyDigits is a 10x10 digits images dataset, which is generated using the elastic deformation.

We chose to generate our own synthetic dataset consisting of 10x10 pixel images of digits that are distorted using elastic deformations. To generate this so-called TinyDigits dataset, we first hand-designed three different 8x8 pixel patterns for each digit. From these patterns, we extended their border to get 10x10 pixel images, and ran elastic deformation to auto-generate other training examples. We used the following parameters for the elastic deformation algorithm in TinyDigits:

- $\alpha$ and $\sigma$: parameters for elastic distortions; set to 3 and 7 respectively. (see Simard et al. 2013 [44] for more details).

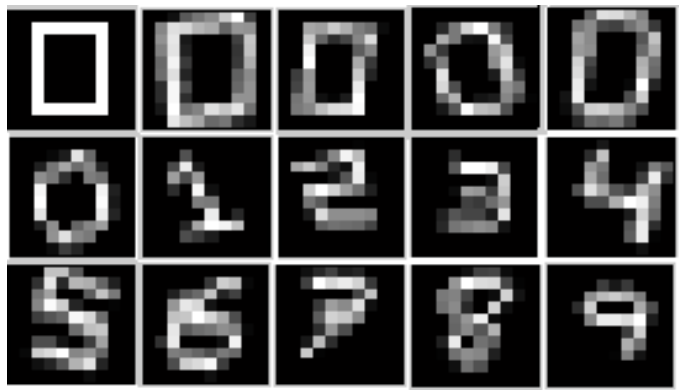

Figure 8.1: Elastic deformation and TinyDigits. **First line**: a hand-designed pattern for the digit "0" and its deformed versions. **Two bottom lines**: examples of generated digits from 0 to 9. Note that all of these deformed images are intelligible

- $\beta$: a random angle from $[-\beta, \beta]$ is used for rotation; set to $\frac{\pi}{12}$.

- $\gamma$: a random scaling from $[1 - \frac{\gamma}{100}, 1 + \frac{\gamma}{100}]$ is used for horizontal and vertical scaling; set to 15.

The elastic deformation and the TinyDigits dataset are illustrated in Figure 8.1.

## 8.2.2   TinyUSPS

After testing some ADATE-generated initialization schemes, we suspected that they are overfited to the TinyDigits task. To test this hypothesis, we need more different tiny datasets, because testing on the original MNIST is time-consuming[1] . TinyUSPS was our first and very simple solution.

USPS is the US Postal Service handwritten digits recognition corpus. It contains normalized grey scale images of size 16x16, divided into a training set of 7291 images and a test set of 2007 images. We took the center 10x10 pixels of these USPS images to create the TinyUSPS dataset. Of course, doing this way could discard some useful information for distinguishing different numbers, which makes the task harder. However, this dataset was indeed very useful and helped us recognize the overfitting problem clearly.

## 8.2.3   Compressed MNIST, Cifar-10, and SVHN

Three of the most famous datasets for image recognition task are MNIST, CIFAR-10 and SVHN:

- The MNIST is a handwritten digits dataset, which has a training set of 60,000 examples, and a test set of 10,000 examples. The digits have been size-normalized and centered in a fixed-size image.

- The CIFAR-10 dataset is an object recognition dataset which consists of 60000 32x32 colour images in 10 classes, with 6000 images per class. There are 50000 training images and 10000 test images.

---

[1] we need to re-optimize many hyper-parameters on the MNIST dataset first, and then train the neural network for at least 10 times to check if the differences are statistical significant

- The Street View House Numbers (i.e. SVHN) is a real-world image dataset which can be seen as similar in flavor to MNIST (e.g., the images are of small cropped digits), but comes from a significantly harder, unsolved, real world problem (recognizing digits and numbers in natural scene images). SVHN is obtained from house numbers in Google Street View images.

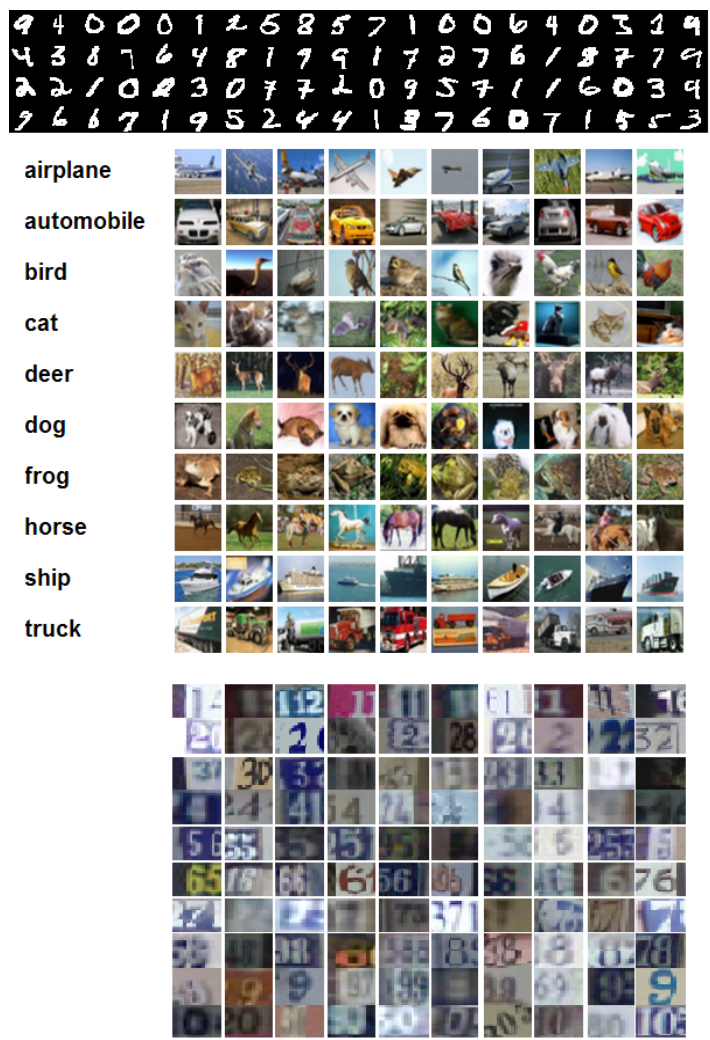

Figure 8.2: From top to bottom: MNIST, CIFAR-10 and SVHN datasets

MNIST, Cifar-10, and SVHN datasets contain much bigger images compared to the USPS (28x28, 32x32, and 32x32 respectively). Therefore, taking 10x10 center pixels of the images would make them unrecognizable, and change the task completely. One possible solution is that we can train a deep autoencoder to compress these datasets. For MNIST, we have used the network: [784-1000-500-100-500-1000-784] and [784-1000-500-30-500-1000-784] to create the compressedMNIST100 and compressedMNIST30 respectively. Other similar network architecture was also used for Cifar-10 and SVHN.

Currently, we use the Hinton implementation of deep autoencoder, which uses staked RBM for pre-training and Conjugate Gradient for fine-tuning. Of course, there are other newer and better ways to train a deep autoencoder, such as using staked denoising autoencoder or sparse autoencoder. However, the Hinton's implementation is easy to use and still a good one.

## 8.3   Design and Implementation

As mentioned earlier, a neural network library has to be implemented in SML before being able to take advantage of ADATE to improve some parts of it. Within this chapter, I will report how I had designed and developed the library. This is indeed a time-consuming and error-prone process, on which I did make several mistakes. Therefore, I want to share my experience on this process, which could be very helpful for other students or researchers. It took me months to overcome all of these mistakes.

Because SML does not support matrix operations and some useful random number generators, I had to build these libraries myself also. I also build a small unit-test library to test my implementation.

All the codes mentioned in this section are provided in Appendix A

### 8.3.1   Matrix library

Matrix operations are essential part of many machine learning algorithms, which base heavily on linear algebra. After searching for an efficient matrix library for SML and could not find any, we have decided to implement a new one. This library represents a matrix as a list of list, which supports common operators on matrix, including multiply, dot multiply, sum, . . . Besides, there is also print functions to help debugging process easier. Storing a matrix as list of list, instead of array of array, makes the implementation efficient and fit naturally in SML language. Many functions were inspired by the standard List library of SML. Besides, all matrix operators in this library have the equivalent complexity compared to what could be done using array of array implementation. We believe that with this library, we can re-implement many different machine learning algorithms easily.

However, we are trying to replace this matrix library by the standard BLAS library, which could provide much higher matrix operations' performance. Please check the Chapter 9 about future works for more details.

#### Matrix data structure

In this library, the matrix data type is defined as:

```
1   type 'a vector = 'a list
    datatype 'a matrix =
3           COLMATRIX of ('a vector list * int * int)
          | ROWMATRIX of ('a vector list * int * int)
```

There are two matrix types: COLMATRIX - matrix that stored as list of column vectors; and ROWMATRIX - matrix that stored as list of row vectors. This storing way could make many matrix operations become as efficient as using array of array. Changing between COLMATRIX and ROWMATRIX type has complexity $O(n * m)$ with $n$ and $m$ are the number of rows and columns of the matrix. However, we can easily transpose a matrix (and switch the matrix type at the same time) instantly at $O(1)$ complexity. A matrix also has its size in its data structure - the last two number in the tuple - to make checking size operations more efficient. We also provide a *checkValid* function to check if a matrix is valid or not. However, if we are using this library as a separated module, the matrix datatype is hided and can only be created using one of the provided initialization functions, which makes sure that all the matrices are valid at all time. These mentioned functions above have the following signatures:

```
val changeType:    'a matrix -> 'a matrix
val transpose:     'a matrix -> 'a matrix
val checkValid:    'a matrix -> bool
val size:          'a matrix -> int * int
```

Matrix in this library can be either an int matrix or a real matrix. However, because SML is a static type language, many matrix operations have an int and a real version. We provide six different ways to initialize either a column or a row matrix, which have the following signatures:

```
val zeroesRealCols: int * int -> real matrix
val zeroesRealRows: int * int -> real matrix
val zeroesIntCols:  int * int -> int matrix
val zeroesIntRows:  int * int -> int matrix
val fromList2Cols: 'a list * int * int -> 'a matrix
val fromList2Rows: 'a list * int * int -> 'a matrix
```

There are also two printing functions, one for real matrix and one for int matrix, to support debugging:

```
val printMatrixReal: real matrix -> unit
val printMatrixInt:  int matrix -> unit
```

This library uses two exceptions: *WrongMatrixType* and *UnmatchedDimension*. It raises the WrongMatrixType exception when the input matrix is not at the desired type, and raises the UnmatchedDimension when the dimension of a matrix itself is unmatched (not a valid matrix) or the two input matrices' dimensions are unmatched (as in matrix multiplication).

**Scalar operators**

Scalar operators are the operators that affect all matrix elements in the same way, such as add a number to the whole matrix. All scalar operators in this library are implemented based on the function *map f m*, which apply f to all elements in the matrix m. This function is very similar to the function List.map, and indeed implemented based on it:

```
fun mapVectors f vs =
    List.map (fn v => List.map f v) vs

fun map f (COLMATRIX(vs, rows, cols)) =
    COLMATRIX (mapVectors f vs, rows, cols)
  | map f (ROWMATRIX(vs, rows, cols)) =
    ROWMATRIX (mapVectors f vs, rows, cols)
```

Using this map function, we can easily implement any scalar operators, for example:

```
1  fun addScalarInt  (m, x:int) = map (fn a => a+x) m
   fun addScalarReal (m, x:real) = map (fn a => a+x) m
3  fun mulScalarInt  (m, x:int) = map (fn a => a*x) m
   fun mulScalarReal (m, x:real) = map (fn a => a*x) m
```

## Matrix Element-wise operators

Matrix element-wise operators are the operators that take two equal-size matrices and combine their elements at the same position by an arbitrary function to a new create a new matrix. All matrix element-wise operators in this library are based on the *merge* function, which take a combining function and two input matrices to produce a new one. It use function mergeVector to merge two vector, and mergeVectors to merge two list of vectors:

```
1  fun mergeVector f v1 v2 =
2    case (v1, v2) of
       ([], []) =>[]
4    | (hdv1::v1', hdv2::v2') =>
            f(hdv1, hdv2)::(mergeVector f v1' v2')
6    | _ => raise UnmatchedDimension

8  fun mergeVectors f vs1 vs2 =
     mergeVector (fn (v1, v2) => mergeVector f v1 v2)
10              vs1 vs2

12 fun merge f (m1, m2) =
     case (m1, m2) of
14   (COLMATRIX(vs1, r1, c1), COLMATRIX(vs2, r2, c2))=>
        COLMATRIX(mergeVectors f vs1 vs2, r1, c1)
16 | (ROWMATRIX(vs1, r1, c1), ROWMATRIX(vs2, r2, c2))=>
        ROWMATRIX(mergeVectors f vs1 vs2, r1, c1)
18 | _ => raise WrongMatrixType
```

Using this merge function, we can easily implement any other matrix element-wise operators, such as:

```
1  val dotMulMatrixInt  = merge (fn (a:int, b)=>a*b)
2  val dotMulMatrixReal = merge (fn (a:real, b)=>a*b)
   val addMatrixInt     = merge (fn (a:int, b)=>a+b)
4  val addMatrixReal     = merge (fn (a:real, b)=>a+b)
```

## Matrix multiplication

Matrix multiplication is the most time-consuming operation in many algorithms. Therefore, We tried to make this operation as fast as possible. This library *only* allows multiply a ROWMATRIX to a COLMATRIX, and raise the exception WrongMatrixType in all other cases. Because only in that case, we can multiply the two list of list right away to produce the desired matrix product. Of course, we can use *changeType()* function to change the matrix type to the desired one. However, that process costs $O(n * m)$ and we decided to force the users do it manually, to make sure that they are aware of that overhead computing cost. Besides, users can choose to produce a ROWMATRIX or a COLMATRIX as the matrix result. I believe that with this flexibility, you usually do not have to change the matrix type in most algorithms. Moreover, the transpose operation in this library literally cost nothing while costing $O(n * m)$ if using array of array implementation. Therefore, using transpose function can compensate for using changeType function.

Implementation of matrix multiplication is a little bit more complicated than the above functions. It uses three helper functions: *foldl2Vectors*, *mulVectorsVector* and *mulVectors*. The *foldl2Vectors* is very similar to the *List.foldl* function, but take two equal-size input vectors and fold them using an input function. The *mulVectorsVector* function multiplies a list of vectors to a vector, and the *mulVectors* multiplies two lists of vector. The signature of the matrix multiplication functions are:

```
val mulMatrixIntR: int matrix*int matrix->int matrix
val mulMatrixIntC: int matrix*int matrix->int matrix
val mulMatrixRealR: realmatrix*realmatrix->realmatrix
val mulMatrixRealC: realmatrix*realmatrix->realmatrix
```

### 8.3.2 RandomExt library

Because SML does not have any built-in function for Gaussian random number generator or random permutation operator, which is needed in neural network implementation, we had to implement them myself. For the Gaussian random generator, we used the *Box–Muller* method, which is very well-known and popular. For generating random permutation (i.e. shuffling), the *Fisher–Yates* method was used.

- *Box–Muller* method: is a pseudo-random number sampling method for generating pairs of independent, standard, normally distributed (zero expectation, unit variance) random numbers, given a source of uniformly distributed random numbers. There are two ways to implement this method: take samples from uniform distribution on the interval (0, 1] or [-1, +1]. In our implementation, we used the second one, which takes two samples from the uniform distribution on the interval [-1, +1] and maps them to two standard, normally distributed samples U(0, 1)

- *Fisher–Yates* method: also known as the Knuth shuffle (after Donald Knuth), is an algorithm for generating a random permutation of a finite set–in plain terms, for randomly shuffling the set. The Fisher–Yates shuffle is unbiased, so that every permutation is equally likely. Fisher–Yates shuffling is similar to randomly picking numbered tickets out of a hat without replacement until there are none left.

### 8.3.3 Smlunit library

Building Matrix, RandomExt, and Neural network library require implementing many different mathematical formula, which are very hard to debug if any mistake is made. Therefore, unit test is a must in this project. After searching and could not find any suitable unit-test library for SML, we had to build our own one, which is simple but functional. This library supports comparing different datatype, measuring running time, and setting time out (for avoiding endless loop).

The most important function in this library is assertFun(), which let us test one input-output pair (i.e. test case) when applying the input into a specific function. Because SML is static type language which does not support type inference, the assertFun() function also needs us to provide the isEqual() function, which helps it compare the desired output and the function result. Each test case also has a corresponding explanation, which will be printed out while testing to help debugging easier. The library also provide the assertFuns() function which let us test a list of test case (i.e. test suits) for a specific function. Using assertFun() and assertFuns() functions, we can easily create appropriate testing function for different output datatypes.

```
//------assertFun() function ------//
fun assertFun isEqual f (input, output, desc) =
  let
  fun runFun ()=
    let
      val timer = Timer.startRealTimer ()
      val result = f(input)
      val time = p_time (Timer.checkRealTimer timer)
    in
      (result, time)
    end
  val (result, time) = runFun();
  in
    assert isEqual (result, output,
              desc ^ " (" ^ time ^")")
  end

//------assertFuns() function ------//
fun assertFuns isEqual f desc testcases =
  let
    val assert = assertFun isEqual f;
    fun assertAll [] = ()
      | assertAll (testcase::testcases) =
        (assert testcase; assertAll (testcases))
  in
    (print ("-------- " ^ desc ^ " --------\n");
    assertAll (testcases))
  end
```

Code 8.1: implementation of assertFun() and assertFuns()

Every library that we have implemented for this thesis has its own unit-test file to check its correctness. Whenever any function is modified or added, we update the unit-test file to reflect the desired changes, and run these files to check that all existing functions are working well, and new modification does not interfere with them.

### 8.3.4   Neural Network Library

Neural network is in fact a general term to indicate a type of model, which consists many different types of techniques from initialization schemes, training algorithms, to regularization methods... To build a flexible and extendable library, we modularized the neural network training process into different parts, which different techniques can be chosen to apply on. We also designed a new appropriate data structure for neural network model, which helps developing process become easier.

**Datatype**

Neural network is a layered model, which are connected by adaptive weights. Based on these properties, we represent a neural network by a list of layer, where each layer consists of one layer of weight (not layer of node), and has the following datatype:

```
type nnlayer = {
    input: real matrix,  (* ROWMATRIX *)
    output: real matrix, (* ROWMATRIX *)
    GradW: real matrix,  (* COLMATRIX *)
    GradB: real vector,
    B: real vector,
    W: real matrix,        (* COLMATRIX *)
    actType: actType
}
```

Code 8.2: datatype for one layer neural network

As you can see, each layer consists of input and output vector, which represent activation activities of input and output nodes of that layer. Besides, W and B represent the weights and bias matrix. Because different layer can have different type of activation function, there is actType field to keep this information. We also include the gradW and gradB matrix, which are the gradients calculated by a particular training algorithm.

Moreover, because there are different techniques and training options that you can choose from before training, we created a params datatype, which contains all these information and can be set through function setParams(). Figure 8.3 shows the params datatype and all available training options so far.

```
type params = {
    batchsize: int,     (* number of training cases per batch *)
    nBatches: int,      (* number of batches *)
    testsize: int,      (* number of testing cases *)
    lambda: real,       (* momentum coefficient *)
    momentumSchedule: bool, (*use/no use Marten's momentum schedule*)
    maxLambda: real,    (* max momentum used in momentum schedule*)
    lr: real,           (* learning rate *)
    costType: costType, (* cost function type
        support: NLL|MSE|CE|PER*)
    initType: initType, (* initialization type
        support: SPARSE|NORMAL|NORMALISED*)
    actType: actType,   (* activation function type
        support: SIGM|TANH|LINEAR*)
    layerSizes: int list, (* structure of network *)
    initWs: real matrix option list, (* pre-initialized Ws matrices *)
    initBs: real vector option list, (* pre-initialized Bs matrices *)
    nItrs: int,    (* number of iterations/epoches *)
    wdType: wdType,     (* weight decay type
        support: L0|L1|L2 *)
    wdValue: real,      (* weight decay value *)
    verbose: bool (* print training information or not *)
}
```

Code 8.3: Params datatype which shows all available training options

**Modules**

The general neural network training process are modularized as shown in figure 8.3. Implementation of these process boxes are the following functions:

- *Pre-processing and batches creating:* implemented by function readData, which can read a dataset contained in a CSV files, and split it into a list of batches.

- *Network architecture*: is defined in params variable and set by function setParams().

- *Initialize Network*: each type of initialization scheme is implemented in different function. Each function takes responsibility of initializing one network layer, and will be called by function initLayers() to generate the whole initialized network. We can also predefine the initialized neural networks by using setParams(), which is very useful when using ADATE to improve the initialization scheme.

- *Forward propagation*: implemented by fprop1layer() and fprop() functions, which do forward propagating for one layer and the whole network respectively.

- *Compute cost*: implemented in computeCost() function, which currently supports: mean square errors (MSE), cross entropy (CE), and NLL (negative log likelihood). This function is also used to compute the error rates of the network.

- *Compute search direction and step size*: currently, this neural network library only supports the SGD training algorithm together with momentum. This training process is computed through back propagation, which propagates the gradient of cost function from the last to the first layer. This process is implemented by bprop1layer() and bprop() for back propagating one layer and for the whole network respectively.

- *Update weights*: implemented by update1layer() and update() functions, which can update one layer and the whole network respectively.

- *Stop criteria*: currently the training process is stopped after a specific number of epochs (stored in the *nItrs* field of params). However, there are two options for the return value: performance of the last trained network (by using trainNN() function) or the best network during the training process (by using trainBest() function).

### 8.3.5   Important Warning!

We highly recommend that you should use Mlton instead of Standard ML of New Jersey (i.e. SML/NJ) for compiling final SML code. Mlton is a whole-program optimizing compiler for SML, which can produce very fast executables file (compared to other SML compiler). In our experience, Mlton produce executable file that run at least five times faster than SML/NJ. More importantly, the floating-point operations in Mlton are equivalent to which of Matlab or C++. We usually implement new algorithm in Matlab first, because it is easier and run faster. After that, we re-implement it in SML. Therefore, we need these two implementations to produce exactly the same result. However, if the SML/NJ is used, it is almost impossible.

However, for writing code, we still recommend using SML/NJ, because it supports a nice REPL (Read–eval–print loop) - an interactive programming environment - which you cannot find in Mlton. Moreover, you should use Emacs as the SML text editor. It has SML mode that provides good syntax highlighting, indentation, and integration with the SML environment.

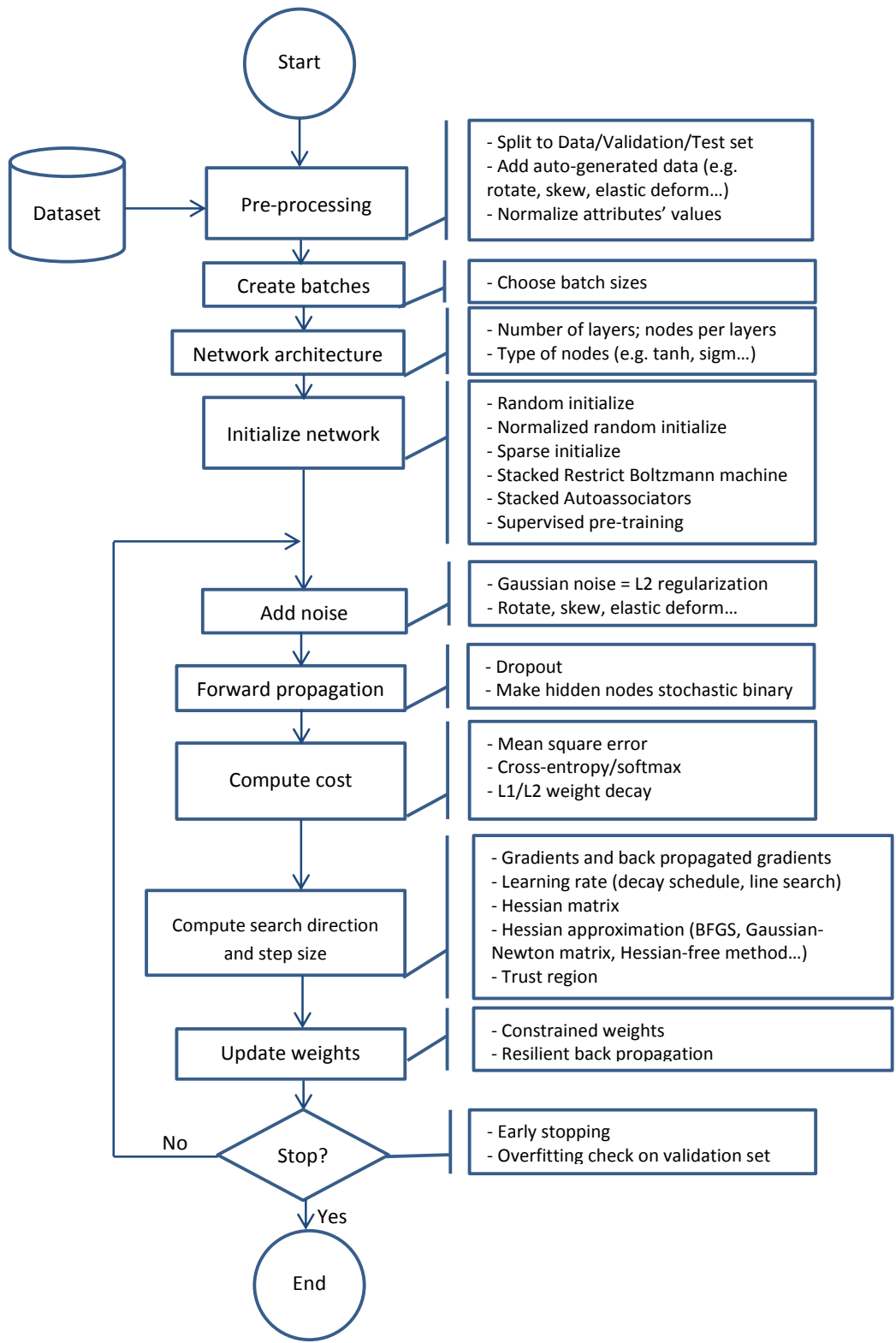

Figure 8.3: General flow chart of gradient-based training algorithms for Neural Networks

## 8.4   Writing Specification file

Thanks to the Neural Network library, writing specification file was pretty simple. The $f(.)$ function is the function that returns a list of initialized weights for each node in the neural network. The original $f(.)$ was the sparse initialization scheme. The only problem was that the $f(.)$ function is written in ADATE-ML part, which does not support the SML list datatype which is used in the neural network library. Therefore, we needed helper functions that transforms the ADATE-ML datatype to SML datatype.

The neural network in ADATE-ML defined as a weightMatrix_list datatype:

```
1  datatype real_list = rnil | consr of real * real_list

3  datatype real_list_list = rlnil | consrl of real_list * real_list_list

5  datatype weightMatrix = weightMatrix of real * real_list_list

7  datatype weightMatrix_list = wnil | consw of weightMatrix *
       weightMatrix_list
```

while in neural network library, the neural network is represented by a list of matrix, which in turn is a list of list also.

The following functions was used to transform the weightMatrix_list datatype into list of matrix datatype.

```
1  (* convert ADATE list type to ML list *)
   fun toRealListListList (Ws: weightMatrix_list):(real * real list list) list
        =
3      let
       fun toRealList (rs): real list =
5          case rs of
           rnil => []
7            |consr(r, rs') => r::toRealList(rs')
       fun toRealListList (rls: real_list_list): real list list =
9          case rls of
           rlnil => []
11           |consrl(rl, rls') => toRealList(rl)::toRealListList(rls')
       in
13       case Ws of
           wnil => []
15         | consw((nInputs, rls),Ws') =>
           (nInputs, toRealListList(rls)) :: toRealListListList(Ws')
17      end
   fun toWeightList (Ws: (real * real list list) list)
19      : real Matrix.matrix option list =
       let
21      fun initWeightMatrix(arg as (n, W):(real * real list list))
           : real Matrix.matrix option =
23          let
           val nInputs = Real.floor (n)
25          fun initSparseList ((ids, initializedWeights, nInputs, I):
```

```
                              (int list * real list * int * int))
27            : real list =
              case (ids, I <= nInputs) of
29            (_, false) => []
                |( [], true)  =>
31       0.0::initSparseList (ids, initializedWeights, nInputs, I+1)
                |( idx::ids', true) =>
33            if (idx = I) then
                 hd(initializedWeights)
35                ::initSparseList (ids',
                             tl(initializedWeights),
37                           nInputs, I+1)
              else
39                 0.0::initSparseList (ids,
                              initializedWeights,
41                            nInputs, I+1)
         fun initSparseVects (nInputs, vs) =
43            case vs of
              [] => []
45                |v::vs' => initSparseList (
                       (Randomext.rand_perm (length(v), nInputs)),
47                    v, nInputs, 1)::
                       (initSparseVects(nInputs, vs'))
49            in
              SOME (Matrix.fromVectors2Cols(initSparseVects(nInputs, W),
51                    (nInputs, length(W))))
            end
53    in
    case Ws of
55        [] => []
        |W::Ws' => initWeightMatrix(W) :: toWeightList(Ws')
57    end
```

For more detail about the specification file, please refer to Appendix B

## 8.5   Experiment Results

A set of 5 training and 5 validation TinyDigits datasets were generated and used for this experiment. Each training dataset consists of 900 training examples, while validation datasets consist of 600 examples. We were using a small DNN $(100 - 80 - 80 - 200 - 10)$, which takes only around 1 second for each epoch.

In our experiment, we were using tanh as the activation function and soft-max with negative log likelihood as the output unit and cost function. The network structure and the number of epochs are fixed. Marten's sparse initialization was used as the starting point for ADATE. After several days of learning, ADATE had synthesized a completely new initialization scheme, which we call sparse-3. The original sparse initialization and the sparse-3 initialization code generated by ADATE are shown in Code 8.4.

```
1 //--Original sparse initialization --//
  fun f( NInputs, NOutputs, LayerType ) : real_list =
3 let
    fun h( N : real ) : real_list =
5     case 0.0 < N of
        false => rnil
```

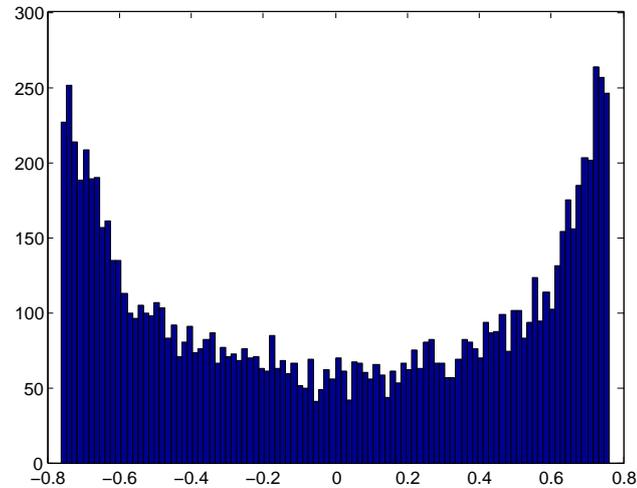

Figure 8.4: Histogram of 10000 instances drawn from the tanh(tanh(randn(.))) distribution

```
 7        | true => consr( randn( 0.0,  1.0 ),  h( N - 1.0 ) )
    in
 9      h  15.0
    end
11  //————————Sparse−3 function ————————//
    fun  f( NInputs,  NOutputs,  LayerType ) =
13    consr(
        0.456463462775,
15      consr(  ~1.43515478736,
          consr( tanh( tanh( rand_normal( 0.0,  1.0 ) ) ), rnil )
17        )
        )
```

Code 8.4: Original sparse and generated sparse-3 initialization functions. The f(.) function is used to generate a list of weights for a node. These weights are then randomly assigned to its incoming connections

### 8.5.1   Sparse-3 description

As shown in Code 8.4, ADATE has created a totally new initialization scheme where each node only has 3 non-zero incoming weights, Two of them are set to the two constants 0.456463462775 and $-1.43515478736$. The third value is drawn from a new distribution: tanh(tanh(randn(.))), in which randn(.) is the normal distribution with $\mu = 0, \sigma^2 = 1$. This distribution looks like an inverted bell curve, sometimes called "*the well curve*", which is bi-modal and usually appears in economic and social phenomena. A histogram of this new distribution is shown in figure 8.4.

At first sight, one could imagine that the sparse-3 method may be overfitted to the artificially generated TinyDigits datasets. Starting with very large weights at the beginning of neural network training could create configurations of weights, that might be useful on TinyDigits but unlikely to be useful on other datasets. Erhan et al., 2009 also suggested that sampling from a fat-tailed distribution in order to initialize a deep architecture could actually hurt the performance of a deep architecture[7].

However, as we shall see shortly in the next section, the performance of sparse-3 method on the MNIST dataset is statistically equivalent [2] to sparse and normalized initialization, but converges much faster. Moreover, while other intialization methods need L2 regularization to overcome overfitting, sparse-3 does not need it. Without weight decay, sparse and normalized methods are outperformed by sparse-3 [2].

### 8.5.2 Sparse-3 Testing

**Methodology**

We experimented on MNIST, a well-known 28 x 28 handwritten digits images dataset composed of 60000 training examples and 10000 test examples. The original training set was further split into a 50000-examples training set and a 10000-examples validation set in our experiments.

For sparse and normalized initialization, the learning rate and $l_2$ cost penalty hyper-parameters are optimized, chosen from [0.05, 0.02, 0.01, 0.005] and [[$10^{-4}, 10^{-5}, 10^{-6}$]] respectively. The sparse-3 initialization is tested without using $l_2$ regularization [3]. We used the momentum schedule suggested by Ilya Sutskever et al. 2012 [46]. All experiments on MNIST used "mini-batches" with a batch cardinality of 200 training examples.

All other hyper-parameters are shown in table 8.4. Each initialization method was experimentally evaluated on network structures with different depths using 10 different random initialization seeds. For the purpose of comparison, we also tested the performance of sparse and normalized initialization on a 4-layer network without $l_2$ regularization.

Table 8.4: Settings used for the experiments

| Hyper-parameters | Configuration |
|---|---:|
| Dataset: | MNIST |
| 4 layers network: | [500 500 2000] |
| 5 layers network: | [2000 1500 1000 500] |
| 6 layers network: | [1500 2000 1500 1000 500] |
| Activation function: | tanh(.) |
| Cost function: | Negative log likelihood |
| Momentum schedule: | $\mu = min(1 - 2^{-1-log_2(\lfloor t/250 \rfloor+1)}, \mu_{max})$ |
| Momentum max: | $\mu_{max} = 0.999$ |
| L2 weight decay: | [$10^{-4}, 10^{-5}, 10^{-6}$] |
| Batch cardinality: | 200 |
| Learning rates: | [0.05, 0.02, 0.01, 0.005] |

**The Convergence Speed Advantage of Sparse-3**

An obvious advantage when training sparse-3 initialized networks is that the networks converge much faster than for any other initialization method, at least twice as fast. The difference is even bigger as the networks become deeper. As you can see in Table 8.5, the validation error rates of sparse-3 at the beginning of the training process are much lower than for the other initializations. Until 50 epochs, sparse-3 has a clear advantage over other methods. However, as the training process comes to around 100 epochs, sparse-3 is losing its dominance and let other method (with help from $l_2$ regularization) catch up.

---

[2]under the null hypothesis test with $p = 0.005$

[3]$l_2$ regularization could hurt sparse-3 performance

Table 8.5: Validation error rates after specific training epochs for different network depth

| | | Epochs | | | | | | |
|---|---|---|---|---|---|---|---|---|
| | | 1 | 5 | 10 | 20 | 30 | 40 | 50 |
| 4-layers | Normalized | 8.90% | 6.44% | 4.70% | 3.30% | 2.62% | 2.27% | 2.22% |
| | Normalized* | 8.92% | 6.38% | 4.49% | 3.32% | 2.53% | 2.30% | 2.22% |
| | Sparse | 8.64% | 4.59% | 3.38% | 2.88% | 2.35% | 2.22% | 2.11% |
| | Sparse* | 8.66% | 4.68% | 3.43% | 2.86% | 2.31% | 2.24% | 2.20% |
| | Sparse-3 | **6.62%** | 3.51% | 2.79% | 2.47% | 2.20% | 2.14% | 1.97% |
| 5-layers | Normalized | 8.12% | 4.86% | 3.45% | 2.56% | 2.13% | 2.06% | 2.03% |
| | Sparse | 8.24% | 3.86% | 3.05% | 2.63% | 2.03% | 2.07% | 2.05% |
| | Sparse-3 | **5.76%** | 2.90% | 2.48% | 2.10% | 2.00% | 1.88% | 1.83% |
| 6-layers | Normalized | 7.89% | 4.45% | 3.19% | 2.53% | 2.24% | 2.17% | 2.04% |
| | Sparse | 8.46% | 3.59% | 3.28% | 2.61% | 2.04% | 2.03% | 2.02% |
| | Sparse-3 | **5.35%** | 2.78% | 2.29% | 1.85% | 1.78% | 1.81% | 1.76% |

This can be explained as follows. After a few dozen epochs, the $l_2$ regularization starts to make effect on the network generalization, which leads to better validation error rates. The $l_2$, however, is not a suitable regularizer for sparse-3 and could hurt its performance.

### Sparse-3 Advantage Regarding Optimization and Generalization

The experimental results in Table 8.5 show the performance of sparse-3, sparse, and normalized initialization for different neural network architectures. It is obvious from the table that sparse-3 converges much faster. For example, sparse-3 reaches a validation error rate of 1.85% after only 20 epochs for a 6-layer architecture whereas the closest state-of-the-art competitor, sparse, only reaches 2.02% after 50 epochs.

Moreover, without help from $l_2$ regularization, sparse-3 significantly outperforms sparse and normalized initialization.

To better understand its advantage, we also compared the learning curves of these methods without using $l_2$ regularization and at a fixed 0.05 learning rate. As we all know, the error surface of deep architectures is very non-convex and hard to optimize with many local minima. As can be seen in Figure 8.5, the sparse-3 learning curves for training as well as validation error are much smoother and converge faster than those for sparse* and normalization*. This suggests that sparse-3 initialization puts us in a region of parameter space where optimization is easier.

Note that adding a $l_2$ regularization term to the training cost will make the optimization process become harder (i.e. longer) in return for better generalization. Therefore, training with sparse-3 initialized networks is indeed much faster and easier. Moreover, this advantage is also magnified when the network gets deeper. This property could be very useful in training very deep neural networks like autoencoders, where underfitting is a big trouble. In fact, the original sparse initialization method was invented by Martens (2010) to overcome exactly that problem.

Table 8.6: Average test error (10 initialization seeds) for best validation for different network structures

| Network | Standard | Normalized | Normalized* | Sparse | Sparse* | Sparse-3 |
|---|---|---|---|---|---|---|
| [500 − 500 − 2000] | 2.17%[1] | **1.74%** | 1.87 | **1.73%** | 1.84% | **1.73%** |
| [2000 − 1500 − 1000 − 500] | -[2] | 1.64%[3] | - | 1.66% | - | 1.63% |
| [1500 − 2000 − 1500 − 1000 − 500] | - | 1.68% | - | 1.62% | - | 1.60% |

* : $l_2$ regularization was not used.
- : results are not relevant.
[1]: produced by Erhan et.al. 2009 [7]. Network structure was also optimized.
[2]: Erhan et.al. 2009 stated that they were unable to effectively train 5-layer models using standard initialization. While X. Glorot and Y. Bengio (2010) produced 1.76% error rates.
[3]: produced by X. Glorot and Y. Bengio (2010) [9]. The network structure is unknown. Our experiments showed slightly worse results.

Figure 8.6 also shows that without $l_2$ regularization, sparse-3 yields better generalization (validation errors) than other methods. Table 8.5 also confirms this point, at least during the first 50 epochs. Combining Figure 8.5 and Figure 8.6, we can see that at the same training cost level, sparse-3 yields a lower test cost. In this sense, sparse-3 appears to bring an effect to that of a regularizer, probably due to its sparsity. This could also explain why sparse-3 does not need $l_2$ regularization.

**Sparse-3 Opimization**

When looking at the sparse-3 initialization, one would probably ask if the two constants in sparse-3 already are optimal. After conducting a small gridsearch on the MNIST dataset with a 4-layer neural network, we concluded that these constants already are relatively optimal, despite being optimized for the TinyDigits dataset. Thus, the constants do not need to be changed when going from TinyDigits to MNIST even if the neural networks for the latter are about two orders of magnitude bigger.

Table 8.7: Gridsearch for the two constants in sparse-3

| Variations | $-1.43515478736$ | | | | |
|---|---|---|---|---|---|
| 0.456463462775 | -0.2 | -0.1 | 0.0 | 0.1 | 0.2 |
| -0.06 | 1.88% | 1.85% | 1.87% | **1.64%** | 1.69% |
| -0.03 | 1.98% | 1.75% | 1.90% | 1.82% | 1.75% |
| 0.00 | 1.89% | 1.76% | **1.68%** | 1.72% | 1.82% |
| 0.03 | 1.78% | 1.98% | 1.77% | 1.84% | 1.76% |
| 0.06 | 1.88% | 2.01% | 1.81% | 1.88% | 1.72% |

## 8.6   Overfitting Problem

Sparse-3 initialization was invented by ADATE after a relatively short run. After sparse-3, we have tried to run ADATE for much longer time. Although new generated programs perform very well on TinyDigits dataset, they are almost useless when applied on MNIST. We attempted to solve this problem using different solutions. However, we did not know exactly what the main cause of overfitting was. We had tried to increase the number of tinyDigits training datasets, make the validation sets bigger, and increase the number of epochs... However, they turned out to be a very wrong way.

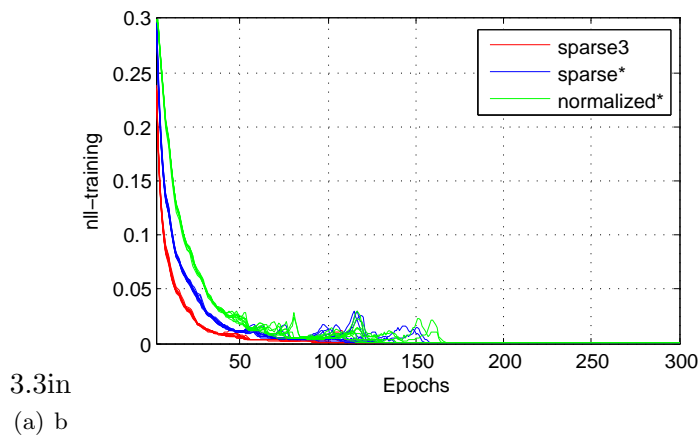

(a) b

Figure 8.5: Learning curves for negative log likelihood training cost

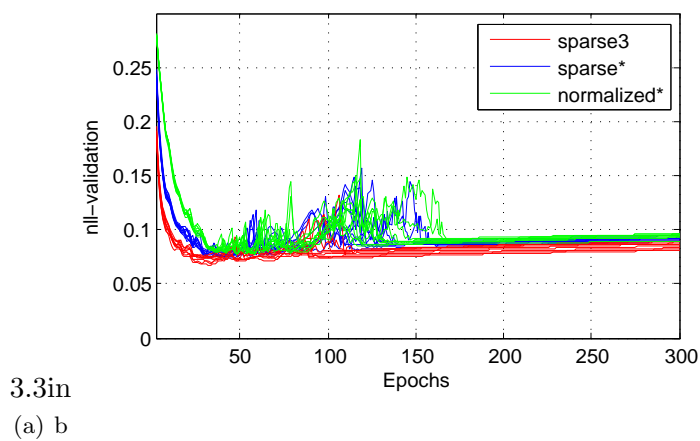

(a) b

Figure 8.6: Learning curves on negative log likelihood validation cost

Figure 8.7: Learning curves of sparse3, sparse*, and normalized* tested on 4-layer neural networks using a 0.05 learning rate. Each method was applied for 10 different initialization seeds.

After that, we decided to make a more systematic experiment on checking what the real overfitting source is. The followings were suspected: architecture, batch size, learning rate, weight decay, momentum, and datasets. In our first experiment, the first five suspected factors were tested. We tested the best ADATE-generated program at that time on the TinyDigits and compare the result to the sparse initialization. We changed these factors' value at each run to see if the ADATE-generated program ovefits to the settings that used during its training process.

As shown in figure 8.8, the ADATE-generated program does not overfit to any of the above factors. Therefore, we conducted a new experiment to check if it overfits to the tinyDigits datasets. Note that we can check this using the original MNIST dataset directly. However, training neural network on MNIST takes a lot of time, including optimizing many hyper-parameters, waiting for many epochs, and training network for at least five times to assure the result. Besides, MNIST contains 28x28 images, which is different from tinyDigits. Therefore, we have to change the network structure and some other hyper-parameters to be able to train on MNIST. This makes it impossible to know where the overfitting comes from: the difference in datasets or the difference in hyper-parameters. Therefore, we used the tinyUSPS and compressedMNIST dataset in this experiment.

The figure 8.9 shows that the datasets are obviously the main source of overfitting, where the ADATE-generated program outperform the sparse initialization on tinyDigits dataset, but being much worse on the other tiny datasets. This suggests that we should mix the TinyDigits dataset with the TinyUSPS or compressedMNIST100 to make ADATE harder to overfit to the training dataset.

| Architecture | | | Batch size | | |
|---|---|---|---|---|---|
| Options | ADATE | Sparse | Options | ADATE | Sparse |
| 80 80 200 | 3.20% | 4.27% | 10 | 3.20% | 4.33% |
| 80 80 80 | 2.60% | 4.27% | 5 | 2.53% | 3.40% |
| 100 100 100 | 3% | 3.67% | 15 | 3.27% | 4.27% |
| 100 100 200 | 4.47% | 4.27% | 20 | 3.33% | 4% |
| 100 200 200 | 2.87% | 3.27% | 25 | 3.53% | 4.13% |
| 200 200 200 | 2.67% | 4% | 30 | 3.67% | 4.53% |
| **Learning Rate** | | | **Momentum** | | |
| Options | ADATE | Sparse | Options | ADATE | Sparse |
| 0.05 | 3.20% | 4.20% | 0.99 | 3.20% | 4.53% |
| 0.025 | 3.33% | 3.73% | 0.95 | 3.20% | 4.60% |
| 0.01 | 4.20% | 4.73% | 0.9 | 3.20% | 3.73% |
| 0.075 | 2.80% | 4.07% | 0 | 3.53% | 3.80% |
| 0.1 | 2.53% | 3.53% | | | |
| **Weight Decay** | | | | | |
| Options | ADATE | Sparse | | | |
| 10^-5 | 3.20% | 4.07% | | | |
| 10^-4 | 3.20% | 3.73% | | | |
| 0 | 3.20% | 4.53% | | | |

Figure 8.8: Experiment on testing overfitting sources: Architecture, Batchsize, Learning Rate, Momentum and Weight Decay. The first value for each factor was the value that used during the ADATE training process

| | TinyDigits | | | TinyUSPS | | | CompressedMnist100 | | |
|---|---|---|---|---|---|---|---|---|---|
| Random Seeds | ADATE | Sparse3 | Sparse | ADATE | Sparse3 | Sparse | ADATE | Sparse3 | Sparse |
| 1 | 3.20% | 3.00% | 3.67% | 4.78% | 4.55% | 4.62% | 3.12% | 2.87% | 2.77% |
| 2 | 2.67% | 3.80% | 4.00% | 4.55% | 4.16% | 4.16% | 2.75% | 2.58% | 2.71% |
| 3 | 3.27% | 4.20% | 3.93% | 4.93% | 4.31% | 4.24% | 2.98% | 2.91% | 2.75% |
| 4 | 3.80% | 3.33% | 3.60% | 4.08% | 4.70% | 3.85% | 3.21% | 2.66% | 2.61% |
| 5 | 2.40% | 4.13% | 4.07% | 4.78% | 4.55% | 4.24% | 3.03% | 2.84% | 2.37% |
| Mean | 3.07% | 3.69% | 3.85% | 4.62% | 4.45% | 4.22% | 3.02% | 2.77% | 2.64% |
| Architecture | 80 - 80 - 200 | | | 80 - 80 - 200 | | | 80 - 80 - 200 | | |
| Batchsize | 10 | | | 10 | | | 200 | | |
| MaxEpochs | 100 | | | 100 | | | 100 | | |

Figure 8.9: Experiment on testing overfitting on datasets

# Chapter 9

# Conclusion and Future Works

This thesis aims at two main purposes: introducing deep learning and its state-of-the-art algorithms, and conducting ADATE experiments to improve deep learning.

Chapter 3 to 6 fulfil the first purpose by giving a short introduction to deep learning and summarizing many recent important discoveries which lead to the deep learning's flourishing such as: unsupervised pre-training strategy or different types of new optimization, regularization methods, and activation functions designed particularly for deep learning. This knowledge is extremely useful for conducting ADATE experiments, especially for deciding which part of deep learning we could improve.

Chapter 8 fulfils the second purpose and answers the research question posed at the beginning of the thesis: "How can the ADATE system improve the performance of deep learning?". The question is answered in a process where we have succesfully designed several different tiny datasets for ADATE, implemented a neural networks library in SML language, and synthesized a brand new sparse-3 initialization scheme. Despite its simplicity, our experiment results have proved the advantage of sparse-3 on classification task over other existing initialization methods. The sparse-3 can double the convergence speed of deep learning. This might suggest that deep learning is still a new subject with many aspects we still do not deeply understand. Therefore, automatic programming can help us overcome our subjective judgments, break our belief, and come up with strange but effective algorithms. This is only our first try on using ADATE to improve deep learning algorithms, and there are many other potential possibilities, such as activation function, objective function, regularization term, learning rate schedule, or momentum formula...

After the discovery of sparse-3, we also conducted other experiments but the results were not as good. We did an analysis for this and recognized the dataset overfitting problem when using the tinyDigits dataset. Other tiny datasets have been built to assess the overfitting. However, because of the time limitation, we could not complete further experiments with these new datasets for this thesis. Besides, the limitation of the ADATE current version that we cannot call the synthesized program from outside of the ADATE part makes it hard to experiment with other parts of deep learning. We expect that based on deep learning knowledge and the neural networks library provided in this thesis, other researchers who find it interesting can conduct future experiments easily, especially when the next version of ADATE is available.

## 9.1   Future Works

For future experiments, we suggest these following development directions

- *Improve Neural Network library performance:* If we can make the SML neural network library run faster, we could open up new possibilities. We can train ADATE with bigger datasets or deeper neural networks in a shorter time. We could even train ADATE on real-world dataset, which can help it overcome the overfitting problem on generated datasets. One improvement that we can do first is instead of using the SML matrix library, we could use the BLAS (Basic Linear Algebra Subprograms) library. BLAS is a high-performance low-level linear algebra library which supports basic linear algebra operations such as copying, vector dot products, linear combinations, and matrix multiplication. There are several BLAS implementations that optimized for specific architectures (e.g. Intel, AMD, ARM...). It is used as a building-block in almost any high-performance scientific computing languages such as MATLAB, R, or Numpy. Using BLAS implementation could speed up matrix multiplication operation, which is used heavily in neural network, 10x-100x faster than using normal loop implementation (depends on CPU architectures and size of the matrices). Currently, we intend to use OpenBLAS, a BLAS implementation that is optimized for different Intel and AMD architectures. To achieve best performance, we suggest compiling the OpenBLAS library to dynamic library (e.g. *.dll or *.so files) for each computer in the cluster to get it optimized for different architectures.

- *Overfitting problem:* The most serious problem that we have to deal with when using ADATE to improve deep learning is the overfitting. We suggest two main approaches, which could possibly help overcome this problem. First, we can improve the quality of the training datasets. By improving performance of neural network library, we can train with bigger datasets, or even with real-world datasets, which could reduce the overfitting. We could also use some kinds of autoencoders to create compressed version of high-dimensionality input. The second approach for this problem is that we can make the overfitting happen as what we want. We can prove that by using ADATE, we can tune a part of deep learning to be optimized for a specific task in an acceptable time.

- *State of the art Algorithms:* Despite being a very new research area, the deep learning literature is developing at extreme speed and dominating all other methods in machine learning, especially for high-level abstraction tasks. Supported by an expanding and very active research community, there are new deep learning algorithms and technologies introduced every year. Therefore, to maximize the ability of ADATE on improving deep learning, we need to keep up with new technologies in deep learning. The best way to do this is to use a standard deep learning library, which is updated frequently with all new central methods. We suggest Pylearn2 for this purpose. It is a machine learning library written in Python and developed by the well-known LISA lab (University of Montreal). It consists of many state-of-the-art deep learning algorithms, and supported by one of the most active deep learning research lab.

# Appendix A

# Neural Network Libraries

## A.1   Matrix Library

### A.1.1   Matrix Signature

```
(*
   File: matrix.sml
   Content: matrix library using List of List type
   Author: Dang Ha The Hien, Hiof, hdthe@hiof.no
   Convention for variable names:
     matrix:          m, m1, m2 ...
     vector:          v, v1, v2 ...
     list of vectors: vs, v1s, v2s ...

*)
signature MATRIX =
sig
   type 'a vector = 'a list
   type 'a matrix
   exception WrongMatrixType
   exception UnmatchedDimension
   (* basic functions for datatype matrix *)
   val changeType:    'a matrix -> 'a matrix
   val checkValid:    'a matrix -> bool
   val transpose:     'a matrix -> 'a matrix
   val transpose_changeType: 'a matrix -> 'a matrix
   val toVector:      'a matrix -> 'a list
   val size:          'a matrix -> int * int
   (*  functions to create new matrix *)
   val initCols:      'a * (int * int) -> 'a matrix
   val initRows:      'a * (int * int) -> 'a matrix
   val uniRandRows: Random.rand  ->  real * (int * int) -> real matrix
   val uniRandCols:   Random.rand  ->  real * (int * int) -> real matrix
   val fromVector2Cols: 'a list * (int * int) -> 'a matrix
   val fromVector2Rows: 'a list * (int * int) -> 'a matrix
   val fromVectors2Cols: 'a list list * (int * int) -> 'a matrix
   val fromVectors2Rows: 'a list list * (int * int) -> 'a matrix
   (*  functions for debugging *)
   val printMatrixReal: real matrix -> unit
   val printMatrixInt:  int matrix -> unit
   (*  functions for output *)
   val printfMatrixReal: string * real matrix -> unit
```

```
38    val printfMatrixInt: string * int matrix -> unit
      (* supported operators on vector *)
40    val mergeVector: (('a*'a)->'b) -> ('a vector * 'a vector) -> 'b vector
      val mergeVect2Matrix: (('a * 'a) -> 'b) -> ('a vector * 'a matrix)
42                                                -> 'b matrix
      val mergeVect'2Matrix: (('a * 'a) -> 'b) -> ('a vector * 'a matrix)
44                                                -> 'b matrix
      val addVect2MatrixReal: real list * real matrix -> real matrix
46    val addVect2MatrixInt: int list * int matrix -> int matrix
      (* supported operators on matrix *)
48    (* scalar operator *)
      val map:              ('a -> 'b) -> 'a matrix -> 'b matrix
50    val addScalarInt:     int matrix * int -> int matrix
      val mulScalarInt:     int matrix * int -> int matrix
52    val addScalarReal:    real matrix * real -> real matrix
      val mulScalarReal:    real matrix * real -> real matrix
54    (* accumulate operator *)
      val foldl:            (('a*'b) -> 'b) -> 'b -> ('a matrix) -> 'b vector
56    val sumInt:           (int matrix) -> int vector
      val sumReal:          (real matrix) -> real vector
58    (* matrix operator *)
      val merge:            (('a*'b)->'c) -> ('a matrix * 'b matrix)
60                                          -> 'c matrix
      val dotMulMatrixInt:  int matrix * int matrix -> int matrix
62    val dotMulMatrixReal: real matrix * real matrix -> real matrix
      val addMatrixInt:     int matrix * int matrix -> int matrix
64    val addMatrixReal:    real matrix * real matrix -> real matrix
      (* real multiply operation *)
66    val mulMatrixIntR:    int matrix * int matrix -> int matrix
      val mulMatrixIntC:    int matrix * int matrix -> int matrix
68    val mulMatrixRealR:   real matrix * real matrix -> real matrix
      val mulMatrixRealC:   real matrix * real matrix -> real matrix
70  end
```

## A.1.2  Matrix Structure

```
   structure Matrix :> MATRIX =
2  struct
     type 'a vector = 'a list
4    datatype 'a matrix = COLMATRIX of ('a vector list * int * int)
                        | ROWMATRIX of ('a vector list * int * int)
6    exception WrongMatrixType
     exception UnmatchedDimension
8     fun changeVectors vs =
     let
10      fun getHeads vs =
       case vs of
12         []      => ([], [])
         | v::vs' => case v of
14              []     => ([], [])
             | hdv::tlv =>
16              case getHeads(vs') of
                   (hdvs, tlvs) => (hdv::hdvs, tlv::tlvs)
18   in
       case getHeads(vs) of
20       ([], _)    => []
       |  (hdvs, tlvs) => hdvs::changeVectors(tlvs)
```

```sml
22    end
      fun changeType (COLMATRIX(vs, rows, cols)) =
24           ROWMATRIX(changeVectors(vs), rows, cols)
         | changeType (ROWMATRIX(vs, rows, cols))  =
26           COLMATRIX(changeVectors(vs), rows, cols)
      fun transpose m =
28    case m of
          COLMATRIX (vs, rows, cols) => changeType(ROWMATRIX (vs, cols, rows))
30        | ROWMATRIX (vs, rows, cols) => changeType(COLMATRIX (vs, cols, rows))

32    fun transpose_changeType m =
      case m of
34        COLMATRIX (vs, rows, cols) => ROWMATRIX (vs, cols, rows)
          | ROWMATRIX (vs, rows, cols) => COLMATRIX (vs, cols, rows)
36        fun fromVectors2List [] = []
            | fromVectors2List (v::vs) = v @ fromVectors2List(vs)
38
      fun toVector m =
40    case m of
          ROWMATRIX (vs, rows, cols) => fromVectors2List (vs)
42        |COLMATRIX (vs, rows, cols) => fromVectors2List (changeVectors(vs))

44    fun sizeVectors vs =
      let
46        val nElems = List.foldl (fn (v, lengthv) =>
                        case (lengthv, (List.length(v) = lengthv)) of
48                          (0, _)     => List.length(v)
                            | (_, true) => List.length(v)
50                          | (_, false) => ~1)
                        0 vs
52        val nVectors = List.length(vs)
      in
54        (nElems, nVectors)
      end
56
      fun checkValid (COLMATRIX(vs, rows, cols)) =
58    let val (nElems, nVectors) = sizeVectors(vs)
      in
60        if (nElems = rows) andalso (nVectors = cols) then true else false
      end
62    | checkValid (ROWMATRIX(vs, rows, cols))  =
      let val (nElems, nVectors) = sizeVectors(vs)
64    in if (nElems = cols) andalso (nVectors = rows)then true else false
      end
66
      fun initVector (value, n, acc) =
68    case n of
        0 => acc
70    | n => initVector(value, n−1, value::acc)
      fun initVectors (value, rows, cols) =
72    let
        fun reduce(cols, acc) =
74        case cols of
              0 => acc
76          | cols => reduce(cols−1,
                      initVector(value, rows, [])::acc)
78    in
          reduce(cols, [])
```

```sml
80    end
      fun initCols (value, (rows, cols)) = COLMATRIX (initVectors(value, rows,
        cols),
82                              rows, cols)
      fun initRows (value, (rows, cols)) = transpose_changeType (initCols(
        value,
84                              (cols, rows)))

86    fun uniRandVectors RandState (nVectors, length, max) =
      let
88      fun uniRandList(n) =
        if n = 0 then []
90      else (max*(2.0*(Random.randReal RandState)-1.0))::uniRandList(n-1)
        fun uniRandVectors(nVectors) =
92      if nVectors = 0 then []
        else uniRandList(length)::uniRandVectors(nVectors-1)
94    in
        uniRandVectors(nVectors)
96    end

98    fun uniRandRows RandState (max, (rows, cols)) =
        ROWMATRIX(uniRandVectors RandState (rows, cols, max), rows, cols)
100   fun uniRandCols RandState (max, (rows, cols)) =
        COLMATRIX(uniRandVectors RandState (cols, rows, max), rows, cols)
102
      fun printInfo m =
104   case m of
          COLMATRIX (_, rows, cols) =>
106         print("Collum matrix "^Int.toString(rows)^" * "
              ^Int.toString(cols)^"\n")
108       | ROWMATRIX (_, rows, cols) =>
            print("Row matrix "^Int.toString(rows)^" * "
110             ^Int.toString(cols)^"\n")
      fun printMatrix printElem printEnd m =
112   let
        fun printVector v =
114     case v of
            [] => ()
116     | last::[] => printEnd last
        | head::v'  => (printElem head; printVector v')
118   in
        case m of
120       ROWMATRIX (vs, rows, cols)=>
          (printInfo m; List.app printVector vs)
122         | COLMATRIX (vs, rows, cols)=>
          (printInfo m; List.app printVector (changeVectors vs))
124   end

126   fun real2str x =
        if x >= 0.0 then Real.toString(x)
128     else "-" ^ Real.toString(~x)
      fun int2str x =
130     if x >= 0 then Int.toString(x)
        else "-" ^ Int.toString(~x)
132
      val printMatrixReal = printMatrix (fn a => print (real2str(a)^", "))
134                             (fn a => print (real2str(a) ^"\n"))
      val printMatrixInt  = printMatrix (fn a => print (int2str(a)^ ", "))
```

```
136                                          (fn a => print (int2str(a) ^ "\n"))

138    fun printfMatrixReal (fname, m) =
          let val fout = TextIO.openOut (fname)
140       in
              ((printMatrix (fn a => TextIO.output (fout, (real2str(a)^", ")))
142                          (fn a => TextIO.output (fout, (real2str(a)^"\n")))
                             m);
144              TextIO.flushOut(fout);
                 TextIO.closeOut(fout))
146       end

148    fun printfMatrixInt (fname, m) =
          let val fout = TextIO.openOut (fname)
150       in
              ((printMatrix (fn a => TextIO.output (fout, (int2str(a)^", ")))
152                         (fn a => TextIO.output (fout, (int2str(a)^"\n")))
                            m);
154              TextIO.flushOut (fout);
                 TextIO.closeOut (fout))
156       end

158    fun size (COLMATRIX(_, rows, cols)) = (rows, cols)
         | size (ROWMATRIX(_, rows, cols)) = (rows, cols)
160
       fun fromList2Vectors (a, nElems, nVectors) =
162    let
           fun fromList2Vector (a, nElems) =
164       case (a, nElems) of
               (_, 0) => (a, [])
166           | ([], nElems) => raise UnmatchedDimension
               | (x::a', nElems) => case fromList2Vector(a', nElems - 1) of
168                         (a'', acc) => (a'', x::acc)
       in
170        case nVectors of
           0 => []
172        | nVectors => case fromList2Vector(a, nElems) of
                  (a', v) => v::fromList2Vectors(a', nElems, nVectors -1)
174    end

176    fun fromVector2Rows (a, (rows, cols)) =
          if List.length(a) <> rows*cols
178       then raise UnmatchedDimension
          else ROWMATRIX(fromList2Vectors(a, cols, rows), rows, cols)
180    fun fromVector2Cols (a, (rows, cols)) = transpose_changeType
                                               (fromVector2Rows(a, (cols, rows)))
182
       fun fromVectors2Cols (vs, (rows, cols)) =
184    if checkValid(COLMATRIX(vs, rows, cols)) then COLMATRIX(vs, rows, cols)
          else raise UnmatchedDimension
186    fun fromVectors2Rows (vs, (rows, cols)) =
          if checkValid(ROWMATRIX(vs, rows, cols)) then ROWMATRIX(vs, rows, cols)
188       else raise UnmatchedDimension

190    fun mapVectors f vs =
          List.map (fn v => List.map f v) vs
192    fun map f (COLMATRIX(vs, rows, cols)) = COLMATRIX (mapVectors f vs, rows,
          cols)
```

```sml
          | map f (ROWMATRIX(vs, rows, cols)) = ROWMATRIX (mapVectors f vs, rows,
          cols)

    fun addScalarInt  (m, x:int) = map (fn a => a+x) m
    fun addScalarReal (m, x:real) = map (fn a => a+x) m
    fun mulScalarInt  (m, x:int) = map (fn a => a*x) m
    fun mulScalarReal (m, x:real) = map (fn a => a*x) m

    fun mergeVector f (v1, v2) =
      case (v1, v2) of
          ([], []) =>[]
          | (hdv1::v1', hdv2::v2') => f(hdv1, hdv2)::(mergeVector f (v1', v2'))
          | _ => raise UnmatchedDimension

    fun mergeVectors f vs1 vs2 = mergeVector (fn (v1, v2) =>
                                    mergeVector f (v1, v2)) (vs1, vs2)

    fun merge f (m1, m2) =
      case (m1, m2) of
          (COLMATRIX (vs1, rows1, cols1), COLMATRIX(vs2, rows2, cols2)) =>
          COLMATRIX(mergeVectors f vs1 vs2, rows1, cols1)
          | (ROWMATRIX(vs1, rows1, cols1), ROWMATRIX(vs2, rows2, cols2)) =>
          ROWMATRIX(mergeVectors f vs1 vs2, rows1, cols1)
          | _ => raise WrongMatrixType

    val dotMulMatrixInt  = merge (fn (a:int, b)=>a*b)
    val dotMulMatrixReal = merge (fn (a:real, b)=>a*b)
    val addMatrixInt     = merge (fn (a:int, b)=>a+b)
    val addMatrixReal    = merge (fn (a:real, b)=>a+b)

    fun mergeVect2Vects f (v1, v2s) =
    case v2s of
        [] => []
      | v2::v2s' => (mergeVector f (v1, v2))::(mergeVect2Vects f (v1, v2s'))

    fun mergeVect2Matrix f (vx, m) =
    case m of
        COLMATRIX(vs, rows, cols) => COLMATRIX(mergeVect2Vects f (vx, vs),
        rows, cols)
        | ROWMATRIX(vs, rows, cols) => ROWMATRIX(mergeVect2Vects f (vx, vs),
        rows, cols)
    val addVect2MatrixReal  = mergeVect2Matrix (fn (v1:real, v2)=> v1+v2)
    val addVect2MatrixInt   = mergeVect2Matrix (fn (v1:int, v2)=> v1+v2)

    fun mergeVect'2Vects f (v1, v2s) =
    let
        fun curf a b = f (a, b)
    in
        case (v1, v2s) of
            ([], []) => []
          | (x::v1', v2::v2s') => (List.map (curf x) v2) :: (mergeVect'2Vects
          f (v1', v2s'))
          | _ => raise UnmatchedDimension
    end

    fun mergeVect'2Matrix f (vx, m) =
    case m of
```

```sml
          COLMATRIX(vs, rows, cols) => COLMATRIX(mergeVect'2Vects f (vx, vs),
        rows, cols)
248     | ROWMATRIX(vs, rows, cols) => ROWMATRIX(mergeVect'2Vects f (vx, vs),
        rows, cols)

250   fun foldl f acc m =
      let fun foldlVectors(vs) =
252       case vs of
             []       => []
254        | v::vs' => (List.foldl f acc v)::foldlVectors(vs')
      in
256       case m of
            COLMATRIX(vs, rows, cols) => foldlVectors(vs)
258        | ROWMATRIX(vs, rows, cols) => foldlVectors(vs)
      end
260   val sumInt = foldl (fn (x, acc:int) => acc+x) 0
      val sumReal = foldl (fn (x, acc:real) => acc+x) 0.0
262
      fun foldl2Vectors f acc (v1, v2) =
264   let fun foldl(v1, v2, acc) =
          case (v1, v2) of
266         (hdv1::v1', hdv2::v2') => foldl(v1', v2', f(hdv1, hdv2, acc))
           | ([], [])             => acc
268        | _                    => raise UnmatchedDimension
      in
270       foldl(v1, v2, acc)
      end
272
      fun mulVectors mul2VectorsFun (v1s, v2s) =
274   let
          fun mulVectorsVector (v1s, v2) =
276       case v1s of
             []        => []
278        | v1::v1s' => (mul2VectorsFun(v1, v2))::
                         (mulVectorsVector (v1s',v2))
280   in
          case v2s of
282         []        => []
           | v2::v2s' => (mulVectorsVector (v1s, v2))::(mulVectors
        mul2VectorsFun (v1s, v2s'))
284   end
      val mul2VectorsInt   = foldl2Vectors (fn (x, y, acc:int) => acc+x*y) 0
286   val mul2VectorsReal  = foldl2Vectors (fn (x, y, acc:real) => acc+x*y) 0.0
      val mulVectorsInt = mulVectors mul2VectorsInt
288   val mulVectorsReal = mulVectors mul2VectorsReal

290   fun mulMatrix mulVectorsFun returnRow ((ROWMATRIX (vs1, rows1, cols1)),
                                             (COLMATRIX (vs2, rows2, cols2))) =
292   if cols1 <> rows2
      then raise UnmatchedDimension
294   else if returnRow then
          (ROWMATRIX (mulVectorsFun(vs2, vs1), rows1, cols2))
296   else
          (COLMATRIX (mulVectorsFun(vs1, vs2), rows1, cols2))
298     | mulMatrix mulVectorsFun returnRow (_, _) = raise WrongMatrixType
      val mulMatrixIntC = mulMatrix mulVectorsInt false
300   val mulMatrixIntR = mulMatrix mulVectorsInt true
      val mulMatrixRealC = mulMatrix mulVectorsReal false
```

```
302    val mulMatrixRealR = mulMatrix mulVectorsReal true
     end
```

## A.2   Random library

### A.2.1   Random Signature

```
1  signature RANDOMEXT =
   sig
3      val rand_normal: Random.rand -> real * real -> real
       val rand_perm:   Random.rand -> int * int -> int list
5  end
```

### A.2.2   Random Structure

```
1    val cached_rand_normal = ref 0.0;
     val rand_use_last = ref false;
3    fun rand_normal RandState (mean, std) =
       let
5        fun box_muller () =
         let
7          val x1 = 2.0 * (Random.randReal RandState) - 1.0
           val x2 = 2.0 * (Random.randReal RandState) - 1.0
9          val w = x1 * x1 + x2 * x2
         in
11         if (w < 1.0) then
             let
13             val v = Math.sqrt((~2.0 * Math.ln(w)) / w)
             in
15               (rand_use_last := true;
                 cached_rand_normal := x2 * v;
17               mean + x1 * v * std)
             end
19         else
               box_muller()
21         end
       in
23       case !rand_use_last of
           true => (rand_use_last := false; mean + !cached_rand_normal*std)
25       | false => box_muller()
         end
27
     fun rand_perm  RandState (m, n) =
29     let
         fun fisher_yates (a, i, n, m) =
31         if (i = n orelse i = m) then a
           else
33         let val randj = i + Real.floor((Random.randReal RandState)
                             *Real.fromInt(n-i))
35           val j = if (randj = n) then n-1 else randj
             val tmp = Array.sub(a, j)
37         in
             (Array.update(a, j, Array.sub(a, i));
39
```

```
                    Array.update(a, i, tmp);
                    fisher_yates (a, i+1, n, m))
41          end
        fun seqArray(a, i, n) =
            if i = n then a
45          else (Array.update(a, i, i+1);
            seqArray(a, i+1, n))
47          fun arrayToList (a, i, l) =
                if ((i = Array.length(a)) orelse (l = 0)) then []
49              else Array.sub(a, i)::arrayToList(a, i+1, l-1)
        val array_perm = fisher_yates(seqArray(Array.array(n, 0), 0, n)
51                                      , 0, n, m)
        val list_perm = arrayToList(array_perm, 0, m)
53      in
        ListMergeSort.sort (fn (a, b) => a>b) list_perm
55      end
```

## A.3   Neural Network Library

### A.3.1   Neural Network Signature

```
1  signature NN =
   sig
3      type 'a matrix = 'a Matrix.matrix
        type 'a vector = 'a Matrix.vector
5      datatype costType = NLL|MSE|CE|PER
        datatype initType = SPARSE|NORMAL|NORMALISED
7      datatype actType = SIGM|TANH|LINEAR
        datatype wdType = L0|L1|L2
9      type nnlayer = {input: real Matrix.matrix, (* ROWMATRIX *)
            output: real Matrix.matrix, (* ROWMATRIX *)
11          GradW: real Matrix.matrix, (* COLMATRIX *)
            GradB: real Matrix.vector,
13          B: real Matrix.vector,
            W: real Matrix.matrix,        (* COLMATRIX *)
15          actType:actType
                    }
17      type params = {batchsize: int, (* number of training cases per batch *)
                nBatches: int,     (* number of batches *)
19              testsize: int,     (* number of testing cases *)
                lambda: real,      (* momentum coefficient *)
21              momentumSchedule: bool,
                maxLambda: real,
23              lr: real,          (* learning rate *)
                costType: costType,  (* cost function type *)
25              initType: initType,  (* initialization type *)
                actType: actType,    (* activation function type *)
27              layerSizes: int list, (* structure of network *)
                initWs: real Matrix.matrix option list,
29                              (* preinitialized Ws matrices *)
                initBs: real Matrix.vector option list,
31                              (* preinitialized Bs matrices *)
                nItrs: int,        (* number of iterations/epoches *)
33              wdType: wdType,
                wdValue: real,
35              verbose: bool
```

```
              }
37    type fileNames = { data_train: string ,
              labels_train: string ,
39            data_test: string ,
              labels_test: string
41        }

43    exception InputError
      exception NotSupported
45
      val setParams:    params -> unit
47    val run:   Random.rand ->        params * fileNames -> nnlayer list * real
      val readData:      string * int * int * int -> real matrix list
49    val trainNN:       nnlayer list * real matrix list
                  * real matrix list -> nnlayer list
51    val trainBest:     nnlayer list * real matrix list * real matrix list
                  * real matrix list * real matrix list -> real * nnlayer list
53    val update:        nnlayer list -> nnlayer list
      val update1layer: nnlayer -> nnlayer
55    val computeCost:   real matrix * real matrix * costType
                  -> real * real matrix
57    val bprop:         nnlayer list * real matrix
                  -> nnlayer list * real matrix
59    val bprop1layer:  nnlayer * real matrix -> nnlayer * real matrix

61    val fprop:         nnlayer list * real matrix
                  -> nnlayer list * real matrix
63    val fprop1layer:  nnlayer * real matrix -> nnlayer * real matrix

65    val initLayers: Random.rand ->   int list * real matrix option list
                  * real vector option list -> nnlayer list
67    val initLayer:  Random.rand ->   int * int * actType
                  * real matrix option * real vector option-> nnlayer
69 end
```

## A.3.2   Neural Network Structure

```
1 structure NN :> NN =
  struct
3 type 'a matrix = 'a Matrix.matrix
  type 'a vector = 'a Matrix.vector
5 exception InputError
  exception NotSupported
7 (* nnlayer: neural network layer type
             a neural network is a list of nnlayer variables.
9            each layer contains neccesary information
             for training and predicting process
11 *)
  (* costType: supported cost function type
13           - Negative log likelihood: only use when you're using oneHot labels.
           - Mean square error
15         - Cross entropy
           - Error percentage - for benchmark only
17 *)
  datatype costType = NLL|MSE|CE|PER
19 (* initType: supported initialization methods
       - Sparse
```

```
21        - Normal
          - Normalized
23 *)
   datatype initType = SPARSE|NORMAL|NORMALISED
25 (* actType: supported activation function
          - Sigmoid
27        - Tanh
   *)
29 datatype actType = SIGM|TANH|LINEAR
   (* wdType: supported weight decay type
31      - L0: no weight decay
        - L1: L1 weight decay
33      - L2: L2 weight decay
   *)
35 datatype wdType = L0|L1|L2
   type nnlayer = {input: real matrix, (* ROWMATRIX *)
37          output: real matrix, (* ROWMATRIX *)
           GradW: real matrix,  (* COLMATRIX *)
39         GradB: real vector,
           B: real vector,
41         W: real matrix,        (* COLMATRIX *)
           actType:actType
43                }
   type params = {batchsize: int,   (* number of training cases per batch *)
45          nBatches: int,     (* number of batches *)
           testsize: int,     (* number of testing cases *)
47         lambda: real,    (* momentum coefficient *)
           momentumSchedule: bool,
49         maxLambda: real,
           lr: real,          (* learning rate *)
51         costType: costType,   (* cost function type *)
           initType: initType, (* initialization type *)
53         actType: actType,    (* activation function type *)
           layerSizes: int list , (* structure of network *)
55         initWs: real matrix option list , (* preinitialized Ws matrices *)
           initBs: real vector option list , (* preinitialized Bs matrices *)
57         nItrs: int,          (* number of iterations/epoches *)
           wdType: wdType,
59         wdValue: real ,
           verbose: bool
61                }
   val params:params ref =
63      ref{batchsize = 10,    (* number of training cases per batch *)
           nBatches = 12,     (* number of batches *)
65         testsize = 30,     (* number of testing cases *)
           lambda = 0.0,    (* momentum coefficient *)
67         momentumSchedule = false ,
           maxLambda = 0.0,
69         lr = 0.005,          (* learning rate *)
           costType = MSE,    (* cost function type *)
71         initType = NORMAL, (* initialization type *)
           actType = SIGM,     (* activation function type *)
73         layerSizes = [4, 8, 3], (* structure of network *)
           nItrs = 40,          (* number of iterations/epoches *)
75         initWs = [] ,
           initBs = [] ,
77         wdType = L0,
           wdValue = 0.0,
```

```sml
79        verbose = false
          };
81 type fileNames = { data_train: string,
                labels_train: string,
83              data_test: string,
              labels_test: string
85          }
   (* all input file names *)
87 val fileNames:fileNames =
       { data_train   = "data_train.csv",
89      labels_train = "labels_train.csv",
        data_test    = "data_test.csv",
91      labels_test  = "labels_test.csv"
       }

95 (* -------------------- Functions for reading inputs --------------*)
   (* parseLine: parse line of string
97    input
       - line: string of numbers seperated by commas
99    output
       - list of real numbers
101 *)
   fun setParams(p: params) =
103     params:=p
   fun parseLine(line) =
105 let
       fun getFirstNum nil = nil
107     |   getFirstNum (x::xs) = if x = #"," then [] else x::getFirstNum(xs);
   in
109     case line of
       [] => []
111     | #","::line' => parseLine(line')
       | #" "::line' => parseLine(line')
113     | _ =>
       let
115         val numStr = implode(getFirstNum(line))
            val num = valOf(Real.fromString(numStr))
117     in num::parseLine(List.drop(line, size(numStr)))
       end
119 end
   (* readData: read input data
121    input
       - nAtts: Each line has nAtts numbers seperated by commas
123     - batchsize: number of lines to create a matrix
        - nBatches: total number of batches
125    output
       - list of batches (each batch is a ROWMATRIX)
127  *)
   fun readData (fileName, nAtts, batchsize, nBatches) =
129     let
       fun readLines (fh, nlines) =
131         case (TextIO.endOfStream fh, nlines = 0) of
            ( _ , true) => []
133         | (false, false) =>
            (parseLine(explode(valOf(TextIO.inputLine fh))))
135         ::readLines (fh, nlines-1)
            | (true, false) => raise InputError
```

```
137   fun readBatches (fh, nBatches) =
          if nBatches > 0
139       then Matrix.fromVectors2Rows(readLines(fh, batchsize),
                              (batchsize, nAtts))
141           ::readBatches(fh, nBatches-1)
          else (TextIO.closeIn fh; [])
143   in
      readBatches(TextIO.openIn fileName, nBatches)
145   end
  (* -------------------- Functions for training process --------------*)
147
  fun initSparseW RandState (nInputs, nOutputs) =
149   let
      val nconn = 4;
151   fun initaNode (nInputs) =
          let
153       fun initSparseList (ids, i, n) =
                  case (ids, i <= n) of
155               (_, false) => []
                  |([], true)  => 0.0::initSparseList (ids, i+1, n)
157               |( idx::ids', true) =>
                  if (idx = i) then
159                   Randomext.rand_normal RandState (0.0, 1.0)
                      ::initSparseList (ids', i+1, n)
161               else
                      0.0::initSparseList (ids, i+1, n)
163       in
          initSparseList (Randomext.rand_perm  RandState (nconn, nInputs)
165                   , 1, nInputs)
          end
167   fun initSparseVects (nInputs, nOutputs) =
          if (nOutputs > 0) then
169       initaNode (nInputs)::initSparseVects (nInputs, nOutputs-1)
          else []
171   in
      Matrix.fromVectors2Cols(initSparseVects (nInputs, nOutputs),
173               (nInputs, nOutputs))
      end
175
  (* initLayer: initialize a neurral network layer
177   input:
          - nInputs: number of incomming nodes
179       - nOutputs: number of outgoing nodes
      output:
181       - a initialized layer
  *)
183 fun initLayer RandState (nInputs, nOutputs, actType, initW, initB):nnlayer
      =
      let
185   val input = Matrix.initRows (0.0, (1, 1))
      val output = Matrix.initRows (0.0, (1, 1))
187   val GradW = Matrix.initCols (0.0, (nInputs, nOutputs))
      val GradB = Matrix.toVector(Matrix.initRows(0.0, (nOutputs, 1)))
189   val B = case initB of
              NONE => Matrix.toVector(Matrix.initRows(0.0, (nOutputs, 1)))
191         | SOME v => v
      val W = case initW of
193         NONE => (case #initType (!params) of
```

```
                        SPARSE => initSparseW RandState (nInputs, nOutputs)
195                        | NORMAL =>
                        Matrix.uniRandCols RandState (
197                            1.0/Math.sqrt(Real.fromInt(nInputs)),
                            (nInputs, nOutputs))
199                        |_ => raise NotSupported)
                | SOME m => m
201        in
          {input = input, output = output, GradW = GradW,
203         GradB = GradB, B = B, W = W, actType = actType}
        end

205
(* initLayers: init the whole neural network
207    input
        - layerSizes: structure of neural network
209    output
        - a list of initialized layer (e.g. a initlaized network)
211    NOTE:
        - if cost function is NLL or CE,
213          last layer should not use any activation function
*)
215 fun initLayers RandState (layerSizes, initWs, initBs) =
    let
217     val (initW, initWs') = case initWs of
                        [] => (NONE, [])
219                    | W::initWs' => (W, initWs')
        val (initB, initBs') = case initBs of
221                        [] => (NONE, [])
                        | B::initBs' => (B, initBs')
223    in
    case layerSizes of
225        [] => []
        | last::[] => []
227        | nInputs::nOutputs::[] =>
            if (#costType(!params) = NLL) orelse (#costType(!params) = CE) then
229            initLayer RandState (nInputs, nOutputs, LINEAR, initW, initB) :: []
            else
231            initLayer RandState (nInputs, nOutputs, #actType(!params), initW,
        initB)
            :: []
233        | nInputs::nOutputs::layerSizes' =>
            initLayer RandState (nInputs, nOutputs, #actType(!params), initW,
        initB)
235            :: initLayers RandState (nOutputs::layerSizes', initWs', initBs')
    end

237
fun sigm x =
239        if x > 13.0 then 1.0
        else if x < ~13.0 then 0.0
241        else 1.0/(1.0+Math.exp(~x))
(* fprop1layer: forward propagate 1 layer
243    input:
        - layer: layer to be propagated
245        - input: input data to be propagated
        output:
247        - (propagated layer, propagated input)
*)
249 fun fprop1layer (layer:nnlayer, input) =
```

```sml
        let
251     val z = Matrix.addVect2MatrixReal(
                #B(layer), Matrix.mulMatrixRealR(input, #W(layer)))
253     val output =
            case #actType(layer) of
255         SIGM => Matrix.map sigm z
            | TANH => Matrix.map Math.tanh z
257         | LINEAR => z
        in
259     ({input = input, output = output, GradW = #GradW(layer),
          GradB = #GradB(layer), B = #B(layer), W = #W(layer),
261       actType = #actType(layer)}, output)
        end
263 (* fprop: fpropagate the whole network
    input:
265     - layers: neural network to be forward propagated
        - input:  input to the first layer (training data)
267   output:
        - (propagated network, final propagated input)
269   ** Note that returing network has the inversed order of layers
         This will be inversed again when using bprop function
271 *)
    fun fprop (layers, input) =
273     List.foldl(fn(layer, (layers', input)) =>
            case fprop1layer(layer, input) of
275             (fpropedLayer, output) => (fpropedLayer::layers', output))
            ([], input) layers

277
    (* bprop1layer: backward propagate 1 layer
279   input:
        - layer: layer to be propagated
281     - gradInput: input gradient to be propagated
      output:
283     - (propagated layer, propagated gradient)
    *)
285 fun bprop1layer (layer:nnlayer, gradInput) =
        let
287     val lambda = #lambda(!params)
        val wdValue = #wdValue(!params)
289     fun dervSigm x = x * (1.0 - x)
        fun dervTanh x = 1.0-x*x
291     fun momentumUpdate (a, b) =  lambda*a + ((1.0 - lambda)*b)
        fun sign x = if x > 0.0 then 1.0 else ~1.0

293
        val gradFunc = case #actType(layer) of
295             SIGM => Matrix.map dervSigm (#output(layer))
                | TANH => Matrix.map dervTanh (#output(layer))
297             | LINEAR =>  Matrix.initRows(1.0,
                            Matrix.size(#output(layer)))
299     val gradOutput = Matrix.dotMulMatrixReal(gradInput, gradFunc)
        val gradOutputC = Matrix.changeType(gradOutput)
301     val oldGradW = #GradW(layer)
        val oldGradB = #GradB(layer)
303     val GradW_noWd = Matrix.mulMatrixRealC(Matrix.transpose(#input(layer)),
                        gradOutputC)
305     val GradW = case #wdType(!params) of
                L0 => GradW_noWd
307             | L1 => Matrix.merge (fn(a, b) => a + wdValue * sign b)
```

```
                              (GradW_noWd, #W(layer))
309               | L2 => Matrix.merge (fn(a, b) => a + wdValue * b)
                              (GradW_noWd, #W(layer))
311     val GradB = Matrix.sumReal(gradOutputC)
        val newGradW = Matrix.merge momentumUpdate (oldGradW, GradW)
313     val newGradB = Matrix.mergeVector momentumUpdate (oldGradB, GradB)
        val propedGrad  = Matrix.mulMatrixRealR(
315         gradOutput, Matrix.transpose(#W(layer)))
        in
317     ({input = #input(layer), output = #output(layer), GradW = newGradW,
          GradB = newGradB, B = #B(layer), W = #W(layer),
319         actType = #actType(layer)}, propedGrad)
        end
321
(* bprop: back propagate the whole network
323     input:
            - layers: neural network with inversed order of layers created by
325     fprop
        - gradInput:  gradient to the last layer
        output:
327         - (propagated network, final propagated gradient)
        ** Note that the input layers has to be in inversed layer order
329         Which is the order in the network returned from fprop function
*)
331 fun bprop (layers, gradInput) =
        List.foldl (fn (layer, (layers', gradInput)) =>
333             case bprop1layer(layer, gradInput) of
                    (bpropedLayer, gradOutput) =>
335                     (bpropedLayer::layers', gradOutput))
                ([], gradInput) layers
337 (* computeCost: compute training/validating cost
        input:
339         - output: predicted result from the neural network
            - target: the desired target
341         - costType: type of cost - MSE/NLL/PER
        output:
343         - (errors, gradient)
*)
345 val i:int ref = ref 0;
    fun computeCost(output, target, costType) =
347     let
        val nSamples = Real.fromInt(#1(Matrix.size(output)))
349     val max = Matrix.foldl (fn (x, acc) => if x>acc  then x else acc)
                    ~10000.0
351     fun mean v = (List.foldl (fn (x, acc) => acc+x) 0.0 v) /
                    Real.fromInt(List.length(v))
353     fun calMSE() =
            let
355         val diff = Matrix.merge (fn (a, b) => a - b) (output, target)
            val errors = Matrix.sumReal(Matrix.map (fn a => 0.5*a*a) diff)
357         val gradient = Matrix.map (fn a => a/nSamples)
                            diff
359         in
            (mean errors, gradient)
361         end
        fun softmax(input) =
363         let
            val maxInput =  max input
```

```
365        val expInput = Matrix.mergeVect '2 Matrix
                          (fn (a, b) => Math.exp(b - a))
367                       (maxInput, input)
           val sumExp = Matrix.sumReal expInput
369        in
           Matrix.mergeVect '2 Matrix (fn (a, b) => b/a)
371                   (sumExp, expInput)
           end
373    fun calNLL() =
           let
375        val _ = i:= (!i+1)
           val softmaxOutput = softmax(output)
377        val errors = Matrix.sumReal(
               Matrix.merge (fn(a,b) => ~(Math.ln(a)*b))
379                   (softmaxOutput, target))
           val gradient = Matrix.merge (fn(a, b) => (~a + b)/nSamples)
381                       (target, softmaxOutput)
           in
383        (mean errors, gradient)
           end
385    fun calCE() =
           let
387        val errors = Matrix.sumReal(
               Matrix.merge
389               (fn(a,b) => Math.ln(1.0 + Math.exp(a)) - a*b)
                  (output, target))
391        val gradient = Matrix.merge
                          (fn (a, b) => (sigm(a) - b)/nSamples)
393                       (output, target)
           in
395        (mean errors, gradient)
           end
397    fun equalReal(a, b) = Real.abs(a-b)<0.0000000001
       fun calPER() =
399        let
           val max_nonzeros = Matrix.foldl
401           (fn (x, acc) => if (x>acc andalso
                              not (equalReal(x, 0.0)))
403               then x else acc)
                   ~20000.0
405        val nSamples = Real.fromInt(#1(Matrix.size(output)))
           val maxOutput = max output
407        val maxTarget = max_nonzeros (Matrix.merge (fn (a, b) => a*b)
                          (output, target))
409        val answers = Matrix.mergeVector (fn (x, y) => equalReal(x, y))
                          (maxOutput, maxTarget)
411        val nRights = List.foldl (fn (x, acc) => if x then acc + 1
                              else acc)
413               0 answers
           in
415        (1.0 - (Real.fromInt(nRights) / nSamples),
            Matrix.initRows(0.0, (1, 1)))
417        end

419    in
       case costType of
421        MSE => calMSE ()
         | NLL => calNLL ()
```

```sml
423          | CE  => calCE ()
             | PER => calPER ()
425        end
    (* update1layer: update 1 layer
427       input
            - layer: a layer to be updated
429       output
            - updated layer
431 *)
    fun update1layer (layer:nnlayer) =
433       let
          val newW = Matrix.merge (fn (a,b) => a - (#lr (!params))*b)
435                     (#W(layer), #GradW(layer))
          val newB = Matrix.mergeVector (fn (a,b) => a - (#lr (!params))*b)
437                     (#B(layer), #GradB(layer))
          in
439       {input = #input(layer), output = #output(layer), GradW = #GradW(layer),
          GradB = #GradB(layer), actType = #actType(layer), W = newW, B = newB}
441       end

443 (* update: update the whole network
    *)
445 fun update (layers) = List.foldr (fn (a, acc) => update1layer(a)::acc)
                          [] layers
447
    (*
449    Support method for training neural network
    *)
451 val i:int ref = ref 0;
    fun train1Batch (layers, input, target) =
453       let
          val (fpropedLayers, output) = fprop (layers, input)
455       val _ = i:= (!i+1)
          val (errors, grad) = computeCost(output, target,
457                        #costType(!params))
          val (bpropedLayers, _) = bprop (fpropedLayers, grad)
459       val _ = if #verbose(!params) then print (Real.toString(errors) ^ "\n")
                                      else ()
461       in
          update(bpropedLayers)
463       end
    fun trainBatches (layers, inputs, targets) =
465       case (inputs, targets) of
          ([], []) => layers
467        | (input::inputs', target::targets') =>
          trainBatches(train1Batch(layers, input, target),
469                inputs', targets')
          | _ => raise Matrix.UnmatchedDimension
471 (* trainNN: train neural network
       input:
473        - data_train: list of batches of training cases
           - target_train: list of batches of desired target
475     output:
            - trained neural network
477 *)
    fun trainNN(startLayers, data_train, target_train) =
479       let
              fun trainEpoches (layers, nItrs) =
```

```sml
481              let val _ = if #verbose (!params) then print
           ("****** epochs: "^Int.toString(nItrs) ^ " ******\n") else ()
483              in
                 if nItrs = 0 then layers
485              else
                     trainEpoches(
487                  trainBatches (layers , data_train , target_train)
                         ,nItrs − 1)
489              end
     in
491          trainEpoches(startLayers , #nItrs (!params))
     end

493
(*
495    TrainNN on training set and pick the best result on validation set
*)
497 fun trainBest(startLayers , data_train , target_train ,
                  data_validation , target_validation) =
499   let
        fun trainEpoches (layers , nItrs , bestErr) =
501       let val _ = if #verbose (!params) then print
          ("****** epochs: "^Int.toString(nItrs) ^ " ******\n") else ()
503          val (_, output) = fprop(layers , hd(data_validation))
             val (errors , _) = computeCost(output , hd(target_validation), PER)
505          val bestErr = if bestErr < errors then bestErr else errors
             val _ = if #verbose (!params) then print
507       ("Best Validation Err = "^ Real.toString(bestErr) ^ "\n") else ()
          in
509          if nItrs = 0 then (bestErr , layers)
          else
511             trainEpoches(
                    trainBatches (layers , data_train , target_train)
513             ,nItrs − 1, bestErr)
          end
515   in
          trainEpoches(trainBatches (startLayers , data_train , target_train),
517                  #nItrs (!params), 1.0)
      end
519 (* run: read input data and train a neural network
*)
521 fun run RandState (p:params, fs:fileNames) =
        let
523       val _ = params := p
          val _ = if #verbose (!params) then print
525               ("********* Reading data ********\n") else ()
          val data_train = readData(#data_train (fs),
527                  hd(#layerSizes (!params)),
                     #batchsize (!params), #nBatches (!params))
529       val labels_train = readData(#labels_train (fs),
                        List.last(#layerSizes (!params)),
531                  #batchsize (!params), #nBatches (!params))
          val data_test = readData(#data_test (fs),
533                  hd(#layerSizes (!params)),
                     #testsize (!params), 1)
535       val labels_test = readData(#labels_test (fs),
                        List.last(#layerSizes (!params)),
537                  #testsize (!params), 1)
          val _ = if #verbose (!params) then print
```

```
539         ("****** Done reading data, start training *****\n") else ()
     val startLayers = initLayers RandState (#layerSizes (!params),
541                        #initWs (!params), #initBs (!params))

543     val trainedLayers = trainNN(startLayers, data_train, labels_train)
     val (_, output) = fprop(trainedLayers, hd(data_test))
545     val (errors, _) = computeCost(output, hd(labels_test),
                          PER)
547     in
          (trainedLayers, errors)
549     end
 end
```

# Appendix B

# ADATE Specification for Initialization Experiment

```
datatype aUnit = aUnit

datatype real_list = rnil | consr of real * real_list

datatype real_list_list = rlnil | consrl of real_list * real_list_list

datatype weightMatrix = weightMatrix of real * real_list_list

datatype weightMatrix_list = wnil | consw of weightMatrix *
      weightMatrix_list
datatype layerType = visHid | hidHid1 | hidHid2 | hidOut

fun rconstLess( ( X, C ) : real * rconst ) : bool =
  case C of rconst( Compl, StepSize , Current ) => realLess( X, Current )

fun rand_normal( ( Mean, Std) : real * real ) : real =
    let
    fun box_muller( Dummy : aUnit ) : real  =
        case realSubtract(realMultiply(2.0,
        (aRand 0)), 1.0) of X1
        => case realSubtract(realMultiply(2.0,
        (aRand 0)), 1.0 ) of X2
        => case realAdd(realMultiply(X1, X1), realMultiply( X2, X2)) of W
        => case realLess(W, 1.0) of
        true  => realAdd(Mean,
              realMultiply(Std,
              realMultiply(X1,
              sqrt(realDivide(realMultiply(~2.0 , ln(W)), W)))))
        | false => box_muller aUnit
      in
    box_muller aUnit
    end

fun rand_2 () =
    2.0*(aRand 0) - 1.0
(* Normalized initialization *)
fun f( NInputs , NOutputs , LayerType ) =
```

```
    let
40      fun h( N: real ) : real_list =
            case  0.0 < N of
42              false => rnil
            |   true =>
44                  consr ( realDevide ( realMultiply(Math.sqrt(6.0), rand_2 ())
        ,
                                    Math.sqrt(realAdd(NInputs, NOutputs)) )
        ,
46                  h(realSubtract( N, 1.0 )))
    in
48      h NInputs
    end



52  fun initW ( (NInputs, NOutputs, LayerType) : real * real * layerType )
                                                    : weightMatrix =
54  let
      fun initSparseVects( I : real ) : real_list_list =
56          case realLess (I, NOutputs) of
              false => rlnil
58          | true =>
                consrl( f( NInputs, NOutputs, LayerType ),
60                  initSparseVects( I + 1.0 ) )
    in
62      weightMatrix( NInputs, initSparseVects 0.0 )
    end



    fun main ((Layer1size, Layer2size, Layer3size, Layer4size, Layer5size) :
66      real * real * real * real * real ) : weightMatrix_list =
        consw(initW(Layer1size, Layer2size, visHid),
68        consw(initW(Layer2size, Layer3size, hidHid1),
        consw(initW(Layer3size, Layer4size, hidHid2),
70        consw(initW(Layer4size, Layer5size, hidOut),
        wnil))))

72
    %%



    val NumInputs =
76    case getCommandOption "--numInputs" of SOME S =>
      case Int.fromString S of SOME N => N



80  val NumIterations =
      case getCommandOption "--numIterations" of SOME S =>
82    case Int.fromString S of SOME N => N

84  val TrainParams : NN.params =
        {batchsize = 10,   (* number of training cases per batch *)
86         nBatches = 90,   (* number of batches *)
           testsize = 4500,  (* number of testing cases *)
88         lambda = 0.99,      (* momentum coefficient *)
           momentumSchedule = false,  (* momentum schedule *)
90         maxLambda = 0.0,    (* max momentum *)
           lr = 0.05,         (* learning rate *)
92         costType = NN.NLL,   (* cost function type *)
           initType = NN.SPARSE,(* initialization type *)
94         actType = NN.TANH,    (* activation function type *)
```

```sml
        layerSizes = [100, 80, 80, 200, 10], (* structure of network *)
96      nItrs = NumIterations,            (* number of iterations/epoches *)
        initWs = [],            (* init weight matrices *)
98      initBs = [],
        wdType = NN.L2,       (* weight decay type *)
100     wdValue = 0.00001,
        verbose = false
102     };

104 val ValidationParams : NN.params =
        {batchsize = 10,    (* number of training cases per batch *)
106     nBatches = 90,     (* number of batches *)
        testsize = 4500,    (* number of testing cases *)
108     lambda = 0.99,          (* momentum coefficient *)
        momentumSchedule = false, (* momentum schedule *)
110     maxLambda = 0.0,    (* max momentum *)
        lr = 0.05,          (* learning rate *)
112     costType = NN.NLL,   (* cost function type *)
        initType = NN.SPARSE, (* initialization type *)
114     actType = NN.TANH,   (* activation function type *)
        layerSizes = [100, 80, 80, 200, 10], (* structure of network *)
116     nItrs = 80,             (* number of iterations/epoches *)
        initWs = [],            (* init weight matrices *)
118     initBs = [],
        wdType = NN.L2,       (* weight decay type *)
120     wdValue = 0.00001,
        verbose = false
122     };

124 (* all training and validation use the same Network structure *)

126 val Inputs =
            [(100.0, 80.0, 80.0, 200.0, 10.0),
128         (100.0, 80.0, 80.0, 200.0, 10.0),
            (100.0, 80.0, 80.0, 200.0, 10.0),
130         (100.0, 80.0, 80.0, 200.0, 10.0),
            (100.0, 80.0, 80.0, 200.0, 10.0),
132         (100.0, 80.0, 80.0, 200.0, 10.0),
            (100.0, 80.0, 80.0, 200.0, 10.0),
134         (100.0, 80.0, 80.0, 200.0, 10.0),
            (100.0, 80.0, 80.0, 200.0, 10.0),
136         (100.0, 80.0, 80.0, 200.0, 10.0),
            (100.0, 80.0, 80.0, 200.0, 10.0),
138         (100.0, 80.0, 80.0, 200.0, 10.0),
            (100.0, 80.0, 80.0, 200.0, 10.0),
140         (100.0, 80.0, 80.0, 200.0, 10.0),
            (100.0, 80.0, 80.0, 200.0, 10.0),
142         (100.0, 80.0, 80.0, 200.0, 10.0),
            (100.0, 80.0, 80.0, 200.0, 10.0),
144         (100.0, 80.0, 80.0, 200.0, 10.0),
            (100.0, 80.0, 80.0, 200.0, 10.0),
146         (100.0, 80.0, 80.0, 200.0, 10.0)]

148 val Test_inputs =
        [ (100.0, 80.0, 80.0, 200.0, 10.0),
150         (100.0, 80.0, 80.0, 200.0, 10.0),
            (100.0, 80.0, 80.0, 200.0, 10.0),
152         (100.0, 80.0, 80.0, 200.0, 10.0),
```

```
              ( 100.0 , 80.0 , 80.0 , 200.0 , 10.0 ),
154           ( 100.0 , 80.0 , 80.0 , 200.0 , 10.0 ),
              ( 100.0 , 80.0 , 80.0 , 200.0 , 10.0 ),
156           ( 100.0 , 80.0 , 80.0 , 200.0 , 10.0 ),
              ( 100.0 , 80.0 , 80.0 , 200.0 , 10.0 ),
158           ( 100.0 , 80.0 , 80.0 , 200.0 , 10.0 ),
              ( 100.0 , 80.0 , 80.0 , 200.0 , 10.0 ),
160           ( 100.0 , 80.0 , 80.0 , 200.0 , 10.0 ),
              ( 100.0 , 80.0 , 80.0 , 200.0 , 10.0 ),
162           ( 100.0 , 80.0 , 80.0 , 200.0 , 10.0 ),
              ( 100.0 , 80.0 , 80.0 , 200.0 , 10.0 ),
164           ( 100.0 , 80.0 , 80.0 , 200.0 , 10.0 ),
              ( 100.0 , 80.0 , 80.0 , 200.0 , 10.0 ),
166           ( 100.0 , 80.0 , 80.0 , 200.0 , 10.0 ),
              ( 100.0 , 80.0 , 80.0 , 200.0 , 10.0 ),
168           ( 100.0 , 80.0 , 80.0 , 200.0 , 10.0 )
              ]
170           )

172 val input_files = [("/local/data1.csv", "/local/labels1.csv",
              "/local/datavalid1.csv", "/local/labelsvalid1.csv"),
174           ("/local/data2.csv", "/local/labels2.csv",
              "/local/datavalid2.csv", "/local/labelsvalid2.csv"),
176           ("/local/data3.csv", "/local/labels3.csv",
              "/local/datavalid3.csv", "/local/labelsvalid3.csv"),
178           ("/local/data4.csv", "/local/labels4.csv",
              "/local/datavalid4.csv", "/local/labelsvalid4.csv"),
180           ("/local/data5.csv", "/local/labels5.csv",
              "/local/datavalid5.csv", "/local/labelsvalid5.csv"),
182           ("/local/data6.csv", "/local/labels6.csv",
              "/local/datavalid6.csv", "/local/labelsvalid6.csv"),
184           ("/local/data7.csv", "/local/labels7.csv",
              "/local/datavalid7.csv", "/local/labelsvalid7.csv"),
186           ("/local/data8.csv", "/local/labels8.csv",
              "/local/datavalid8.csv", "/local/labelsvalid8.csv"),
188           ("/local/data9.csv", "/local/labels9.csv",
              "/local/datavalid9.csv", "/local/labelsvalid9.csv"),
190           ("/local/data10.csv", "/local/labels10.csv",
              "/local/datavalid10.csv", "/local/labelsvalid10.csv"),
192
              ("/local/data11.csv", "/local/labels11.csv",
194           "/local/datavalid11.csv", "/local/labelsvalid11.csv"),
              ("/local/data12.csv", "/local/labels12.csv",
196           "/local/datavalid12.csv", "/local/labelsvalid12.csv"),
              ("/local/data13.csv", "/local/labels13.csv",
198           "/local/datavalid13.csv", "/local/labelsvalid13.csv"),
              ("/local/data14.csv", "/local/labels14.csv",
200           "/local/datavalid14.csv", "/local/labelsvalid14.csv"),
              ("/local/data15.csv", "/local/labels15.csv",
202           "/local/datavalid15.csv", "/local/labelsvalid15.csv"),
              ("/local/data16.csv", "/local/labels16.csv",
204           "/local/datavalid16.csv", "/local/labelsvalid16.csv"),
              ("/local/data17.csv", "/local/labels17.csv",
206           "/local/datavalid17.csv", "/local/labelsvalid17.csv"),
              ("/local/data18.csv", "/local/labels18.csv",
208           "/local/datavalid18.csv", "/local/labelsvalid18.csv"),
              ("/local/data19.csv", "/local/labels19.csv",
210           "/local/datavalid19.csv", "/local/labelsvalid19.csv"),
```

```sml
            ("/local/data20.csv", "/local/labels20.csv",
212             "/local/datavalid20.csv", "/local/labelsvalid20.csv"),
             ];

214
fun readData (data, target, data_valid, target_valid) =
216     (NN.readData(data,
          hd(#layerSizes(TrainParams)),
218      #batchsize(TrainParams), #nBatches(TrainParams)),
         NN.readData(target,
220      List.last(#layerSizes(TrainParams)),
         #batchsize(TrainParams), #nBatches(TrainParams)),
222      NN.readData(data_valid,
         hd(#layerSizes(TrainParams)),
224      #testsize(TrainParams), 1),
         NN.readData(target_valid,
226      List.last(#layerSizes(TrainParams)),
         #testsize(TrainParams), 1))
228 (* real input data for training process *)
val input_data = Array.fromList (List.map readData input_files)

230
(* convert ADATE list type to ML list *)
232 fun toRealListListList (Ws: weightMatrix_list):(real * real list list) list
       =
       let
234   fun toRealList (rs): real list =
          case rs of
236       rnil => []
             |consr(r, rs') => r::toRealList(rs')
238   fun toRealListList (rls: real_list_list): real list list =
          case rls of
240       rlnil => []
             |consrl(rl, rls') => toRealList(rl)::toRealListList(rls')
242      in
       case Ws of
244        wnil => []
       | consw( weightMatrix(nInputs, rls),Ws') =>
246        (nInputs, toRealListList(rls)) :: toRealListListList(Ws')
         end
248 fun toWeightList RandState (Ws: (real * real list list) list)
         : real Matrix.matrix option list =
250       let
       fun initWeightMatrix(arg as (n, W):(real * real list list)
252          : real Matrix.matrix option =
            let
254        val nInputs = Real.floor (n)
          fun initSparseList ((ids, initializedWeights, nInputs, I):
256             (Int32.int list * real list * Int32.int * Int32.int))
             : real list =
258          case (ids, I <= nInputs) of
          (_, false) => []
260          |([], true)  =>
        0.0::initSparseList (ids, initializedWeights, nInputs, I+1)
262          |( idx::ids', true) =>
          if (idx = I) then
264            hd(initializedWeights)
             ::initSparseList (ids',
266                tl(initializedWeights),
                nInputs, I+1)
```

```
          else
              0.0 :: initSparseList ( ids ,
                  initializedWeights ,
                  nInputs , I+1)
      fun initSparseVects ( nInputs , vs ) =
          case vs of
          [] => []
              | v :: vs ' => initSparseList (
                  ( rand_perm RandState ( length(v) , nInputs )),
                  v , nInputs , 1) ::
              ( initSparseVects( nInputs , vs '))
          in
        SOME ( Matrix.fromVectors2Cols( initSparseVects( nInputs , W) ,
                  ( nInputs , length(W))))
      end
    in
  case Ws of
      [] => []
      | W :: Ws ' => initWeightMatrix(W) :: toWeightList RandState ( Ws ')
      end

val Abstract_types = []
val Reject_funs = []
fun restore_transform D = D
fun compile_transform D = D
val print_synted_program = Print.print_dec '

val Funs_to_use = [
  "false", "true",
  "realLess", "realAdd", "realSubtract", "realMultiply",
  "tanh",
  "tor", "rconstLess",
  "rand_normal",
  "0",
  "aRand",
  "rnil", "consr"
  ]

fun to( G : real ) : LargeInt.int =
    Real.toLargeInt IEEEReal.TO_NEAREST ( G * 1.0e14 )

structure Grade : GRADE =
struct
type grade = LargeInt.int
val NONE = LargeInt.maxInt
val zero = LargeInt.fromInt 0
val op+ = LargeInt.+
val comparisons = [ LargeInt.compare ]
val N = LargeInt.fromInt 1000000 * LargeInt.fromInt 1000000
val significantComparisons = [ fn( E1, E2 )
                              => LargeInt.compare( E1 div N , E2 div N )  ]

fun toString( G : grade ) : string =
    Real.toString( Real.fromLargeInt G / 1.0E14 )

val pack = LargeInt.toString
```

```
326
    fun unpack( S : string ) : grade =
328   case LargeInt.fromString S of SOME G => G

330 val post_process = fn X => X

332 val toRealOpt = NONE

334 end

336 val Inputs = take( NumInputs, Inputs )

338 fun output_eval_fun( exactlyOne( I : int, _ : (real * real * real * real *
        real),
            WeightList : weightMatrix_list ) ) = [
340 let
      val _ = NN.setParams
342         ( if I < Int64.fromInt NumInputs
          then TrainParams
344       else ValidationParams )
      val  RandState = Random.rand( 10, Int64.toInt I )
346   val ( data_train, labels_train, data_test, labels_test) =
        Array.sub(input_data, Int64.toInt I)
348   val Ws = toWeightList RandState (toRealListListList(WeightList))
      val startLayers = NN.initLayers RandState (#layerSizes(TrainParams), Ws,
350     [])
      val trainedLayers = NN.trainNN(startLayers, data_train, labels_train)
      val (_, output) = NN.fprop(trainedLayers, hd(data_test))
352   val (errors, _) = NN.computeCost(output, hd(labels_test), NN.PER)

354   val () = (
        p"\noutput_eval_fun: I = "; print_int64 I;
356     p"  errors = "; print_real errors;
        p"\n"
358     )

360 in
      if errors > 1.0E30 orelse not ( Real.isFinite errors ) then
362       { numCorrect = 0 : int, numWrong = 1 : int, grade = to 1.0E30 }
      else
364       { numCorrect = 1, numWrong = 0, grade = to errors }
end
366     ]

368 exception MaxSyntComplExn
    val MaxSyntCompl = (
370   case getCommandOption "--maxSyntacticComplexity" of
        NONE => 150.0
372   | SOME S => case Real.fromString S of SOME N => N
      ) handle Ex => raise MaxSyntComplExn

376 fun rlEq( rnil,  rnil ) = true
      | rlEq( rnil,  consr( _, _ ) ) = false
378   | rlEq( consr( _, _ ), _ ) = false
      | rlEq( consr( X1, Xs1 ), consr( Y1, Ys1 ) ) =
380         real_eq( X1, Y1 ) andalso rlEq( Xs1, Ys1 )
```

```
382
     fun rllEq( rlnil,   rlnil ) = true
384    | rllEq( rlnil,   consrl( _, _ ) ) = false
       | rllEq( consrl( _, _ ), _ ) = false
386    | rllEq( consrl( X1, Xs1 ), consrl( Y1, Ys1 ) ) =
           rlEq( X1, Y1 ) andalso rllEq( Xs1, Ys1 )
388
     fun wmEq( weightMatrix( X, Xss ), weightMatrix( Y, Yss ) ) =
390    real_eq( X, Y ) andalso rllEq( Xss, Yss )

392  fun wlEq( wnil,   wnil ) = true
       | wlEq( wnil,   consw( _, _ ) ) = false
394    | wlEq( consw( _, _ ), _ ) = false
       | wlEq( consw( X1, Xs1 ), consw( Y1, Ys1 ) ) =
396        wmEq( X1, Y1 ) andalso wlEq( Xs1, Ys1 )

398  val AllAtOnce = false
     val OnlyCountCalls = false
400  val TimeLimit : Int.int = 10000000
     val max_time_limit = fn () => Word64.fromInt TimeLimit : Word64.word
402  val max_test_time_limit = fn () => Word64.fromInt TimeLimit : Word64.word
     val time_limit_base = fn () => real TimeLimit
404
     fun max_syntactic_complexity() = MaxSyntCompl
406  fun min_syntactic_complexity() = 0.0
     val Use_test_data_for_max_syntactic_complexity = false
408
     val main_range_eq = wlEq
410  val File_name_extension =
       "numIterations" ^ Int.toString NumIterations ^
412    "numInputs" ^ Int.toString NumInputs

414  val Resolution = NONE
     val StochasticMode = false
416
     val Number_of_output_attributes : Int64.int = 4
418
     fun terminate( Nc, G ) = false
```